\documentclass[runningheads]{llncs}
\usepackage{graphicx}
\usepackage{comment}
\usepackage{amsmath,amssymb} %
\usepackage{color}

\usepackage{caption}
\usepackage{subfig}
\usepackage{booktabs}
\usepackage{multirow}

\begin{document}
\pagestyle{headings}
\mainmatter
\def\ECCVSubNumber{3125}  %

\title{Weakly-supervised 3D Shape Completion in the Wild} %

\titlerunning{Weakly-supervised 3D Shape Completion in the Wild}

\author{
    Jiayuan Gu\inst{1,2} \and
    Wei-Chiu Ma\inst{1,3} \and
    Sivabalan Manivasagam\inst{1,4} \and
    Wenyuan Zeng\inst{1,4} \and
    Zihao Wang\inst{1} \and
    Yuwen Xiong\inst{1,4} \and
    Hao Su\inst{2} \and
    Raquel Urtasun\inst{1,4}
}

\authorrunning{J. Gu et al.}
\institute{
    Uber Advanced Technologies Group \and
    University of California, San Diego \\ \email{\{jigu, haosu\}@eng.ucsd.edu} \and
    Massachusetts Institute of Techonology \and
    University of Toronto \\ \email{\{wichiu, manivasagam, wenyuan, yuwen, urtasun\}@uber.com}
}

\maketitle

\newcommand{\eg}{\textit{e}.\textit{g}.}
\newcommand{\ie}{\textit{i}.\textit{e}.}
\newcommand{\myparagraph}[1]{\noindent \textbf{#1}}

\begin{abstract}
3D shape completion for real data is important but challenging, since partial point clouds acquired by real-world sensors are usually sparse, noisy and unaligned.
Different from previous methods, we address the problem of learning 3D complete shape from unaligned and real-world partial point clouds.
To this end, we propose a weakly-supervised method to estimate both 3D canonical shape and 6-DoF pose for alignment, given multiple partial observations associated with the same instance.
The network jointly optimizes canonical shapes and poses with multi-view geometry constraints during training, and can infer the complete shape given a single partial point cloud.
Moreover, learned pose estimation can facilitate partial point cloud registration.
Experiments on both synthetic and real data show that it is feasible and promising to learn 3D shape completion through large-scale data without shape and pose supervision. 
\end{abstract}

\section{Introduction}
We are interested in the problem of 3D shape completion, which estimates the complete geometry of objects from partial observations. This task is a prerequisite for many important real-world applications. For example, complete shapes can facilitate automated vehicles to track objects~\cite{giancola2019leveraging} and robots to figure out the best pose to grasp objects~\cite{varley2017shape}.
Previous works~\cite{dai2017shape,insafutdinov2018unsupervised,yuan2018pcn} have successfully applied deep learning methods to learn shape priors from large-scale synthetic data, which results in improvement of the 3D shape completion task.
However, most these prior works have two major limitations: 1) they require the ground-truth shape for learning, and 2) they assume the input partial point clouds are aligned and normalized to the canonical frame, in which the object faces forward and are centered at the origin. 
In addition, models trained on synthetic data do not transfer well to the real world due to the domain gap.

We aim to use real data for the 3D shape completion task. However, since there is a lack of real 3D data that comes with sufficient high-quality ground-truth 3D shapes, we cannot directly adopt these supervised learning methods developed in the synthetic domain. Although there are a few datasets containing real scans, such as KITTI~\cite{Geiger2013IJRR} and ScanNet~\cite{dai2017scannet}, no efforts are made to explore the possibility of learning 3D shape completion in a weakly-supervised fashion.
There are three challenges to work on real 3D data, unique from the synthetic ones:
1) No or few ground-truth complete shapes are available for full supervision.
Note that annotating 3D shapes are more difficult and expensive than annotating 2D images;
2) Partial point clouds acquired by real-world sensors like RGB-D cameras or LiDAR are sparse and noisy;
3) Poses and sizes of objects are more diverse, and partial observations may be occluded by other objects.

In this paper, we address the problem of learning 3D shape completion from real, unaligned partial point clouds without shape and pose supervision (Sec~\ref{sec:problem}).
The proposed method is weakly-supervised by multi-view consistency of instances (Sec~\ref{sec:method}).
The key contributions of our work are as follows:
\begin{enumerate}
    \item We propose a weakly-supervised\footnote{We use the term ``weakly-supervised'' instead of ``unsupervised learning of shape and pose''~\cite{insafutdinov2018unsupervised} to avoid confusion, which are in fact equivalent.} approach to learn 3D shape completion from unaligned point clouds. Our promising results show that it is feasible to learn 3D shape completion from large-scale 
    3D data without shape and pose supervision.
    \item We showcase the extension of our method to tackle the challenging partial point cloud registration problem.
\end{enumerate}
\section{Related work}
\myparagraph{3D reconstruction from single images}
3D shape completion is highly related to 3D reconstruction from single images, since a partial point cloud can be obtained from a RGB-D image.
Since the problem is ill-posed by nature, many learning-based approaches are developed to learn shape priors from large-scale data.
\cite{choy20163d} uses a recurrent 3D CNN to predict a 3D occupancy grid given one or more images of an object.
\cite{tulsiani2017multi} proposes a differentiable `view consistency' loss and a probabilistic occupancy grid.
\cite{fan2017point} pioneers the representation of point clouds as output.
However, they all require full supervision from synthetic images rendered from ShapeNet~\cite{chang2015shapenet}.
Performance on real datasets like Pascal 3D+~\cite{xiang2014beyond} suffer from unrealistic ground truth shapes from aligned CAD models.

Thus, \cite{yan2016perspective,zhu2017rethinking,tulsiani2018multi,insafutdinov2018unsupervised} focus on reconstructing 3D shapes with weak supervision.
Especially, \cite{tulsiani2018multi} enforces geometric consistency between the independently predicted shape and pose from two views of the same instance. 
Differential point clouds (DPC) \cite{insafutdinov2018unsupervised} uses a similar strategy to reconstruct point clouds and devises differentiable projection of point clouds.
However, it is non-trivial to extend these methods to real-world data, which will be discussed in Sec~\ref{sec:dpc}.

\myparagraph{3D reconstruction from multiple frames}
By leveraging consecutive frames, 3D
shapes can be reconstructed from RGB images~\cite{triggs1999bundle,agarwal2010bundle,eigen2014depth} or depth images~\cite{newcombe2011kinectfusion}.
The problem is also known as Structure-from-Motion (SfM).
\cite{ummenhofer2017demon,zhou2017unsupervised,tang2018ba} are proposed to tackle it with deep learning.
Although poses are estimated in both SfM and our 3D shape completion, the main difference is that unseen 3D points are hallucinated in our 3D shape completion while depths are estimated in the SfM.

KinectFusion~\cite{newcombe2011kinectfusion} fuses all the depth data streamed from a Kinect sensor into a single global implicit surface model of the observed scene in real-time.
It demonstrates the advantages of maintaining a full surface model compared to frame-to-frame tracking.
Our method benefits from the similar idea, but differs from it in 2 aspects:
1) The proposed approach is a learning-based framework based on 3D point clouds only.
2) The trained model can predict the complete shape from a single point
cloud and the relative pose between two distant views during inference, which is demonstrated in our experiments.

\myparagraph{3D shape completion}
3D shape completion is usually performed on partial scans of individual objects.
With the success of deep learning, learning-based approaches show more flexibility and better performance compared with geometry-based and alignment-based methods.
\cite{dai2017shape} combines a data-driven shape predictor and analytic 3D shape synthesis.
\cite{yuan2018pcn} proposes a variant of PointNet~\cite{qi2017pointnet} to directly process point clouds and generate high-resolution outputs.
\cite{tchapmi2019topnet} devises a tree-style neural network to generate structured point clouds.

3D shape completion without full supervision is of increasing interest to the community.
\cite{stutz2018learning} finetunes the encoder on the target dataset, like KITTI~\cite{Geiger2013IJRR}, with a fixed generator pretrained on the ground truth SDF representation of synthetic data, like ShapeNet~\cite{chang2015shapenet}.
\cite{han2019multi} generates half-to-half sequence pairs from the ground truth complete point clouds of ShapeNet, and learns features by half-to-half prediction and self-reconstruction.
\cite{chen2019unpaired} trains autoencoders to learn embedding features of shape on clean, complete synthetic data and noisy, partial target data.
An adaption network is learned to transform the embedding space of noisy point clouds to that of clean point clouds with a GAN setup.
However, none of those works deals with unaligned point clouds and relies on complete synthetic data to pretrain.

\myparagraph{Deep learning for point clouds}
PointNet~\cite{qi2017pointnet} is the pioneer to directly process point clouds with a deep neural network, followed by many variants~\cite{qi2017pointnet++,dgcnn,li2018pointcnn}.
It extracts features for each point with a shared multi-layer perceptron (MLP), and outputs with an aggregation function invariant to permutation.
Any point-cloud-based neural network can work as the encoder of our method.
\section{Problem}
\label{sec:problem}
The goal of 3D shape completion is to predict a complete shape $Y$ given a partial observation $X$.
In this work, we represent the partial observation and the complete shape as point clouds: $X \in \mathcal{R}^{n \times 3}$ and $Y \in \mathcal{R}^{m \times 3}$, where $n$ and $m$ are the number of partial and complete points respectively.

Previous approaches~\cite{yuan2018pcn,stutz2018learning} have assumed that partial observations are normalized according to ground-truth bounding boxes and transformed into a predefined canonical frame, \eg, the forward-facing object centered at the origin.  
Past works may also assume the ground-truth shape $Y^{gt} \in \mathcal{R}^{m_{gt} \times 3}$ is available and train a model in a supervised setting via a permutation-invariant loss function $L(Y, Y^{gt}; X)$, such as Chamfer Distance (CD) or Earth Mover Distance (EMD) \cite{fan2017point} to evaluate reconstruction quality.
While these ground truth information may be available on synthetic data, they may not be available in the real-world setting.
Thus, we build on past works and propose a more general and challenging setting: 
\begin{itemize}
    \item We do not assume knowledge of the ground-truth canonical frame for normalizing and aligning the partial observations. Instead, we maintain the partial observations in the \emph{sensor} coordinate system.
    \item We do not assume knowledge of the ground-truth shape.
    \item The point cloud observation is captured by a sensor (\eg,~a LiDAR) from the real world and can therefore be sparse and also noisy.
    \item Instead of ground truth, we have access to a set of unaligned partial observations of the instance, captured at different viewpoints by the sensor.
    
\end{itemize}

We call this more realistic setting ``weakly-supervised shape completion in the wild''. 
This setting is especially applicable to the real-world setting such as in autonomous driving or indoor scene navigation, where the robot may observe other moving agents from multiple viewpoints and needs to reason about the shape and pose to perform shape completion.
In the next section, we propose our method for tackling weakly-supervised shape completion in the wild.

\section{Method}
\label{sec:method}
\begin{figure}[h!]
    \centering
    \includegraphics[width=0.95\linewidth]{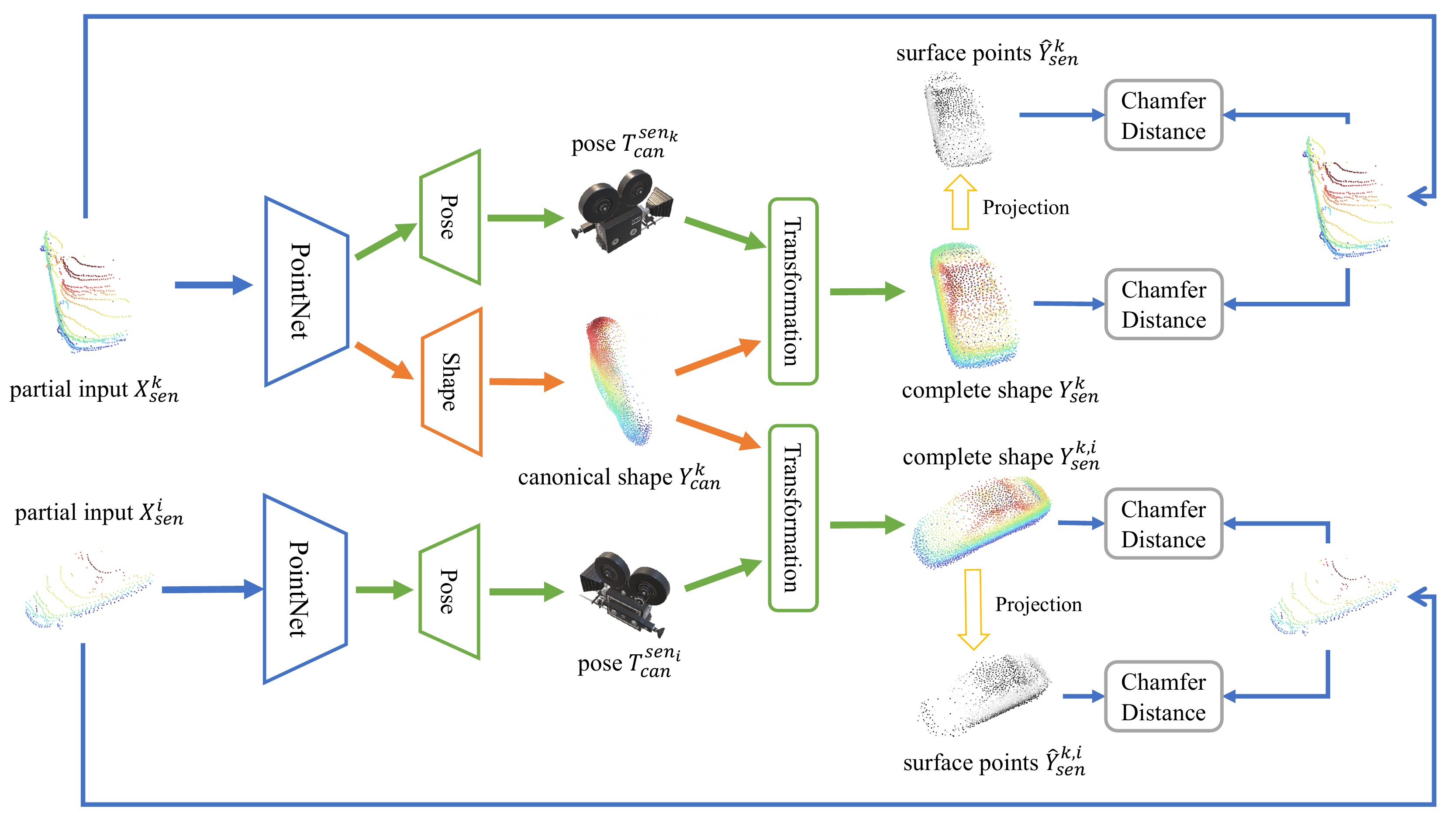}
    \caption{Overview of our weakly-supervised shape completion pipeline at training time. 
    This illustration is for a pair of partial inputs and can be extended to partial input of a shape from multiple views, by optimizing the averaged loss among selected pairs of inputs.}
    \label{fig:overview}
\end{figure}

\subsection{Overview}
We tackle weakly-supervised shape completion in the wild by jointly learning the canonical shape and pose of the object.
The underlying idea is that predicting a complete shape $Y_{sen}$ in the \emph{sensor} coordinate system is equivalent to predicting a complete shape $Y_{can}$ in the \emph{canonical} coordinate system\footnote{The \emph{canonical} frame in our method is not predefined, but emerges during training.} and then transforming it according to a 6-DoF pose $T_{can}^{sen}$, where $Y_{sen} = T_{can}^{sen}Y_{can}$.
But a key question remains: how do we learn the complete shape and pose when we do not have access to the ground-truth for either? 
We leverage the fact that, during training, we have access to multiple observations of the object from different viewpoints. We know that these observations, while noisy, accurately represent different views of the GT shape. By enforcing that predicted shapes and poses are consistent with recorded observations, we can train the network in a weakly-supervised fashion to estimate both the shape and pose from a single observation.

Our training approach is as follows:
Given a set of sensor observations of the object of interest, $\{X_{sen}^1, X_{sen}^2, \cdots, X_{sen}^M\}$, we apply a deep autoencoder network to each observation $X_{sen}^i$ and predict a canonical shape $Y_{can}$ and pose $T_{can}^{sen_i}$. 
We then apply two loss terms based on these outputs to guide the network to learn the correct shape and pose: (1) the partial observation points should match the completed shape transformed by the estimated pose (observation-matching-shape), and (2) the surface points of the completed shape as viewed by the estimated sensor pose should match the observation points with self-occlusion taken into consideration (shape-projection-matching-observation). 
Because we have access to multiple observations and multiple pose predictions, we can encourage the network to generate a completed shape estimate that minimizes these two loss terms with respect to all observations and poses.
We call this \emph{multi-view consistency}.

During inference, the trained network takes as input a single partial point cloud, and outputs the estimated pose and completed shape.
Our pipeline is illustrated in Fig~\ref{fig:overview}. 
The following sections describe in detail our approach.
Sec~\ref{sec:shape-pose} describes the network architecture to predict 3D canonical shapes and 6-DoF poses.
Sec~\ref{sec:match} and Sec~\ref{sec:projection} present the loss terms (observation-matching-shape and shape-silhouette-matching-observation). 
Sec~\ref{sec:multi-view-loss} describes how we extend the two loss terms to work on multiple observations.

\subsection{Predict canonical shape and pose}
\label{sec:shape-pose}
Given a partial observation $X_{sen}$, we employ a deep encoder-decoder network to predict both the 3D canonical shape and 6-DoF pose.
The encoder encodes the input $X_{sen}$ to a latent code $z$.
We use the same encoder architecture as PCN's~\cite{yuan2018pcn}, which is a variant of PointNet~\cite{qi2017pointnet}.
The shape decoder is a 3-layer MLP, which decodes $z$ to a fixed number of 3D coordinates $Y_{can} \in R^{m \times 3}$.
The pose decoder, also a 3-layer MLP, outputs a rotation $\hat{R}$ and a translation $t$.
Following~\cite{zhou2019continuity},the rotation is represented as a 6D vector, and the translation a 3D vector.
The inferred rotation matrix and translation form $T_{can}^{sen}$.
Thus, the predicted complete shape in the \emph{sensor} coordinate system is calculated as:

\begin{equation}
    Y_{sen} = T_{can}^{sen}Y_{can} = f_{T}(z)f_{shape}(z)
\end{equation}

To alleviate the issue of local minima and overcome bad initialization, we follow prior art and have multiple pose decoder branches in our network and train them with the hindsight loss introduced in DPC~\cite{insafutdinov2018unsupervised}.
In brief, hindsight loss is where, for each batch, gradients are only backpropagated to the branch with the lowest loss.

\subsection{Match partial observation with canonical shape}
\label{sec:match}
We now describe the observation-matching-shape loss, which we implement as an asymmetric Chamfer-Distance (CD) between the observation point cloud and the completed shape point cloud in the sensor-coordinate space.
The asymmetric CD (Eq~\ref{eq:cd-i2o}) between the input observation $X_{sen}$ and the output shape $Y_{sen}$ is
\begin{equation}
    CD(X_{sen} \mapsto Y_{sen}) = \frac{1}{|X_{sen}|} \sum_{x \in X_{sen}}{\min_{y \in Y_{sen}}{||x - y||_2}}
    \label{eq:cd-i2o}
\end{equation}
This forces the output canonical shape to completely cover the input observation.
However, it does not guarantee that $Y_{sen}$ is close to $X_{sen}$ --- even a point cloud that fills the whole 3D space would minimize Eq~\ref{eq:cd-i2o}, which is not desired.
Thus, we need to develop a more sophisticated loss term to enforce how the sensor acquires the point cloud and compute the distance between the input observation and the projection of the output, which is described next.

\subsection{Project canonical shape to partial observation}
\label{sec:projection}
We now describe the shape-projection-matching-observation term.
Using our knowledge of how the sensor acquires observations, we can ``simulate'' which points on the surface are observed based on the estimated complete shape point cloud and the estimated pose.
We can then force the "generated" point cloud to match the true observation.
We tailor this loss term based on knowledge of how the LiDAR sensor works.

Given a subset point cloud $\hat{Y}_{sen}$ of the predicted point cloud $Y_{sen}$, which are on the surface as viewed from the sensor (the ``simulated'' observation), another asymmetric CD (Eq~\ref{eq:cd-o2i}) between the input $X_{sen}$ and those \emph{surface} points is optimized.
\begin{equation}
    CD(\hat{Y}_{sen} \mapsto X_{sen}) = \frac{1}{|\hat{Y}_{sen}|} \sum_{\hat{y} \in \hat{Y}_{sen}}{\min_{x \in X_{sen}}{||\hat{y} - x||_2}}
    \label{eq:cd-o2i}
\end{equation}

We introduce a simple, flexible and efficient way to infer the \emph{surface} points.
The point cloud acquired by the LiDAR sensor can be projected to a range image, which is essentially the polar-coordinate system of the LiDAR sensor. 
The polar coordinate $(\phi, \theta, r)$ of a cartesian sensor observation point $(x, y, z)$ 
is calculated as Eq~\ref{eq:polar-transform}:
\begin{equation}
    r = \sqrt{x^2+y^2+z^2}, \phi = tan^{-1}\left(\frac{x}{y}\right), \theta = sin^{-1}\left(\frac{z}{r}\right)
    \label{eq:polar-transform}
\end{equation}
where $r$ is radial distance to the sensor and $\phi, \theta$ are the azimuth and pitch angles, respectively, of the ray shot from the LiDAR sensor.
According to the resolution of LiDAR $(d_{\phi}, d_{\theta})$, we can discretize $(\phi, \theta)$ to $(\lfloor \frac{\phi}{d_{\phi}} \rfloor, \lfloor \frac{\theta}{d_{\theta}} \rfloor)$, which forms several bins.
For each bin, the point with the smallest distance is considered to be on the \emph{surface}.
This is the depth buffer approach widely used for rasterization in the computer graphics literature.

This ``projection'' approach is differentiable and simple to implement, since we can just count the occupied bins and find the smallest distance in each.
No voxelization or normalization of the points is 
needed like in DPC~\cite{insafutdinov2018unsupervised}, which makes our approach more flexible, especially for real data without a normalized scale.
Additionally, we can flexibly adjust the projection resolution to be coarser than the real resolution, which helps with noisy and occluded real-world data.
Furthermore, this method is also efficient as the computation complexity is $O(m)$, where $m$ is the number of predicted points.

\subsection{Multi-view consistency}
\label{sec:multi-view-loss}
Both loss terms, observation-matching-shape and shape-projection-matching-observation, work not only for the input observation $X_{sen}^i$, but also works for all other observations of the instance in the set.
Inspired by \cite{tulsiani2018multi,insafutdinov2018unsupervised}, we leverage the consistency among multiple views associated with the same instance to supervise 3D shape prediction and 6-DoF pose estimation.
During training, we sample $M$ observations $\{X_{sen}^1, X_{sen}^2, \cdots, X_{sen}^M\}$ of one instance within a batch.
One view is selected as the \emph{reference}, denoted by index $k$.
Intuitively, 
all observations share the same complete rigid shape in the \emph{canonical} coordinate system. 
In other words, for any view $i$, $Y_{can}^i$ should be close to $Y_{can}^k$. 
Therefore, we can replace $Y_{sen}^i$ with $Y_{sen}^{k, i}= Y_{can}^{k} \hat{R}_{i}^T + t_{i}$, which forces the network to learn a complete canonical shape matching all the partial views.

The full loss for a given training example $\{X_{sen}^i, i \in 1...N\}$ with reference view $k$ is calculated as Eq. \ref{eq:loss}:
\begin{equation}
    \begin{aligned}
    L(\{X_{sen}^i\}) = & \sum_{i=1}^{M} CD(X_{sen}^i \mapsto Y_{sen}^{k, i}) 
    + \beta CD(\hat{Y}_{sen}^{k, i} \mapsto X_{sen}^i)
    \end{aligned}
    \label{eq:loss}
\end{equation}
where $\beta$ is a hyper-parameter, which can be adjusted according to the quality of data.
While we could apply multi-view consistency between all possible pairs (\ie,~make each index in the observation set the reference index and sum all terms), we choose one randomly to reduce training complexity.

\section{Experiments}
\label{sec:exp}
In this section, we demonstrate our method and several baselines on this new setting of weakly-supervised shape completion in the wild.
We first evaluate our method on the standard synthetic data benchmark, ShapeNet.
We then showcase the performance of our method on two real-world self-driving datasets for which we construct ground-truth complete shapes.
Furthermore, we demonstrate our method also works on the task of point cloud registration.
Finally, we compare our approach against a fully-supervised oracle.

\subsection{Datasets}
\myparagraph{ShapeNet~\cite{chang2015shapenet}}
ShapeNet is a richly-annotated, large-scale dataset of 3D synthetic shapes.
We focus on 3 categories: chairs, cars, and airplanes.
We use the same data and split provided by DPC~\cite{insafutdinov2018unsupervised},
where the data available for each training example is 5 random RGB-D views of the model.
We note that this data only has random viewpoint/orientation, and the translation component of the view is fixed.
To acquire partial point clouds in the camera coordinate system, we backproject depth maps according to the intrinsic matrix.
The average number of points of the partial point clouds for chairs, cars and airplanes is 3018, 2956, 756 respectively.
For evaluation, 8192 ground-truth points are sampled from the surface of the CAD models.

\myparagraph{3D vehicle dataset~\cite{manivasagam2020lidarsim}}
We build a collection of complete vehicle object point clouds from a large-scale LiDAR dataset for self-driving that contain bounding box instance annotations for over 1.2 million frames.
We generate the ground-truth complete shape as follows:
for each static object, we accumulate the LiDAR points inside the bounding box and determine the object relative coordinates for the LiDAR points based on the bounding box center.
Since cars are usually symmetric, we postprocess data by mirroring the aggregate point cloud along the vehicle's heading axis, followed by Gaussian statistical outlier removal, to acquire complete shapes for annotated objects.
Visualizations of the ground-truth shape can be seen in Fig~\ref{fig:real-data}.
There are 13700 annotated objects in total, splitted into 10000/700/3000 for train/val/test.
On average, each object is associated with 80 scans, and each scan contains 1163 points.
We filter observations to include at least 100 points to avoid overly sparse observations.

\myparagraph{SemanticKITTI~\cite{behley2019iccv}}
Instance and semantic annotations for the LiDAR point clouds are provided for all sequences of the Odometry Benchmark.
We use SemanticKITTI's odometry localization to aggregate partial point clouds of the same parked vehicle instance (with at least 512 points on average) into a single vehicle frame and apply radius outlier removal.
Following \cite{behley2019iccv}, we train our network on instances generated from sequences 00 to 10, except for sequence 08 instances which are used as test set.
There are 659/229 instances and 51186/16299 observations for training/test.
On average, each object is associated with 95 scans, and each scan contains 1377 points.

\subsection{Tasks and metrics}
\myparagraph{3D shape completion}
For shape completion, the algorithm is required to predict shapes in the sensor coordinate system.
Given the ground-truth canonical shape and pose, we can compute the ground-truth shape in the sensor coordinate system.
Then, we adopt the standard metrics used in the literature~\cite{fan2017point,yuan2018pcn,insafutdinov2018unsupervised,tulsiani2018multi}.
The main metric to evaluate shape completion against ground truth point cloud $Y_{sen}^{gt}$ is the Chamfer Distance (Eq~\ref{eq:cd}).
The first term is called the \emph{Precision} and the second term is called the \emph{Coverage}.
\begin{equation}
    \begin{split}
        CD(\hat{Y}_{sen} \leftrightarrow Y_{sen}^{gt}) = \frac{1}{|\hat{Y}_{sen}|} \sum_{\hat{y} \in \hat{Y}_{sen}}{\min_{y \in Y_{sen}}{||\hat{y} - y^{gt}||_2}}\\[0.1\baselineskip]  + \frac{1}{|Y_{sen}^{gt}|} \sum_{y^{gt} \in {Y}_{sen}^{gt}}{\min_{\hat{y} \in \hat{Y}_{sen}}{||y^{gt} - \hat{y}||_2}}
        \label{eq:cd}
    \end{split}
\end{equation}

\myparagraph{Partial point cloud registration}
Given two partial observations, the algorithm is required to predict the relative pose from one to the other.
This task is more challenging than common point cloud registration.
The algorithms are evaluated by calculating the quaternion distance $\theta$, or \emph{angle difference}, between the estimated pose $q_{pred}$ and the GT pose $q_{gt}$ for all instances in the dataset:
$\theta = 2 \arccos \langle q_{pred}, q_{gt} \rangle$.
Following DPC~\cite{insafutdinov2018unsupervised}, we report the median of angle difference and accuracy (the percentage of samples for which $\theta \leq 30^{\circ}$).
In addition, if the translation is predicted, we also report the median of mean-square-error between the prediction and the ground truth.

\subsection{Baselines}
To our knowledge, there are currently no weakly-supervised methods for shape completion that take as input a single unaligned partial point cloud.
We instead compare our method against the state-of-the-art single-image 3D reconstruction method, and standard point cloud alignment methods.

\myparagraph{DPC~\cite{insafutdinov2018unsupervised}}
\label{sec:dpc}
DPC is a weakly-supervised method, which is trained on image pairs to reconstruct 3D point clouds.
We compare to their reported results of shape reconstruction on ShapeNet, and adapt their method to range images.
We argue that it is non-trivial to adapt DPC to the "wild" setting, where both pose rotation and translation are unknown and the shape size is not bounded.
We list their drawbacks as follows:
1) It is assumed that canonical shapes are normalized into a unit cube;
2) A fixed camera distance to the object is provided and only rotations are considered in the original paper;
3) The projection loss only is sensitive to density and occlusion.
Despite these drawbacks, we modify DPC by replacing perspective transformation with polar transformation and scaling the normalized output by a fixed factor to match the real scale, denoted by DPC-LIDAR.
We refer readers to the supplementary for more details.

\myparagraph{ICP}
Since our ground-truth complete shapes are acquired by accumulating partial point clouds given ground-truth transformations, we introduce a baseline based on Iterative Closest Point (ICP).
For two consecutive frames, we calculate the rigid transformation between two partial point clouds by ICP.
Thus, given a pair of partial point clouds, the transformation can be calculated by accumulating results from ICP.
All the partial point clouds can be transformed into a certain frame and fused.
To reduce the accumulated error, we choose the middle frame as the reference.
For 3D shape completion, the fused point cloud in the reference frame is transformed according to the ground-truth pose to compare with the ground-truth complete shape.
For point cloud registration, we compare the transformation from one frame to the reference one with the ground-truth transformation.
Local ICP~\cite{besl1992method} and Global ICP~\cite{rusu2009fast} are the two ICP algorithms we compare against.
We use the implementation of Open3D~\cite{Zhou2018} and search the best hyper-parameters on the validation set (0.175 for the distance threshold in Local ICP and 0.125 for the voxel size in Global ICP).

\subsection{Implementation details}
As our synthetic and real datasets are of different sizes and come very different distributions, we slightly modify our data input and our implementation of the model for each setting:

\myparagraph{Input}
To ease the learning requirements for our model, we preprocess input partial point clouds.
Given a partial point cloud in the sensor coordinate system, we shift the point cloud to be centered at the origin without knowledge of the ground-truth shape or pose:
For ShapeNet, an axis-aligned bounding box in the camera coordinate system is calculated, and its center is shifted to the origin;
for real data, we first extract a bounding frustum of the input partial point cloud, and then centralize the frustum\footnote{The resulting coordinate system is similar to \emph{3D mask coordinate} introduced in \cite{qi2018frustum}.}.
After converting to this origin-shifted coordinate system, the input point cloud is resampled with replacement to a fixed size, as done in the original PointNet~\cite{qi2017pointnet}.
For ShapeNet and for real LiDAR datasets, we uniformly resample 3096 and 1024 points, respectively, from the input partial point cloud.

\myparagraph{Training}
During training, 4 observations per instance are sampled in a batch.
Adam is used as the optimizer.
For synthetic data, models are trained with an initial learning rate of $1e^{-4}$ for 300k iterations and a batch size of 32.
The learning rate is decayed by 0.5 every 100K iterations.
For real data, models are trained with an initial learning rate of $1e^{-4}$ for 400k iterations and a batch size of 32.
The learning rate is decayed by 0.7 every 100K iterations.
Especially, all the observations of one instance in a batch are within a window of 20 frames.
It takes less than 16 hours to train our model with a GTX 1080Ti.
The loss weight $\beta$ is set to 0.25 and 0.05 for synthetic and real data respectively.

\subsection{Results of 3D shape completion}
\label{sec:results-shape}

\myparagraph{ShapeNet}
We first demonstrate shape completion results on ShapeNet.
The chamfer distance, precision and coverage are reported on the test set.
We also include the chamfer distance of DPC reported by \cite{insafutdinov2018unsupervised}.
Besides, we compare our approach with DPC$^\dagger$, which predicts both rotation and translation of the pose.
Note that \cite{insafutdinov2018unsupervised} evaluates \emph{shape prediction}, rather than \emph{shape completion}, by aligning the predicted canonical shape according to the ground truth of the validation set.
We argue that this evaluation protocol assumes that all the objects share the same canonical space and it does not really disentangle shape and pose.

\begin{figure}[t]
    \centering
    \includegraphics{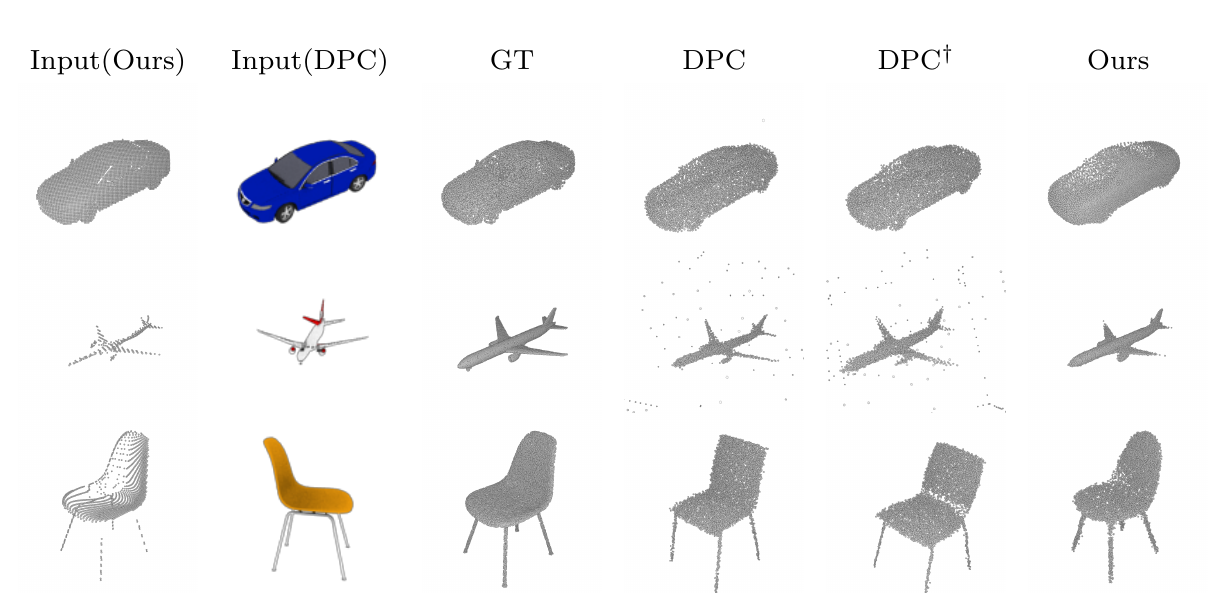}
    \caption{
        Qualitative results of 3D shape completion on the test set of ShapeNet.
        All the (input and predicted) point clouds are transformed to the ground-truth canonical frame and visualized at a fixed viewpoint.
        \emph{Note that the input of our method is the point cloud lifted from the depth map input of DPC}.
    }
    \label{fig:shapenet}
\end{figure}

Table~\ref{tab:shapenet} shows the quantitative comparison between our method and the DPC variants.
Despite not having access to the ground-truth translation or size of the object, our model is able to predict a more accurate complete shape.
Fig~\ref{fig:shapenet} shows the qualitative results.
Since planes are usually flat and result in sparse observations, DPC fails to learn a clean shape while our approach is more robust.
We refer readers to the supplementary for ablation studies and more details.

\begin{table}[t]
    \small
    \setlength{\tabcolsep}{4pt}
    \centering
    \begin{tabular}{l|ccc}
        \toprule
        & CD                      & Precision               & Coverage           \\
        \midrule
        DPC                                                          & 6.26/3.54/4.85          & 4.19/1.95/2.59          & 2.07/1.59/2.26          \\
        DPC$^\dagger$                                                   & 13.17/4.73/7.32         & 8.17/2.52/3.81          & 5.00/2.20/3.52          \\
        DPC (pre-aligned~\cite{insafutdinov2018unsupervised})           & 3.91/3.47/4.30          & -                       & -                       \\
        DPC$^\dagger$ (pre-aligned~\cite{insafutdinov2018unsupervised}) & 5.07/4.09/5.86          & -                       & -                       \\
        \midrule
        Ours                                                            & \textbf{1.95/2.68/3.33} & \textbf{0.91/1.27/1.69} & \textbf{1.05/1.41/1.64} \\
        \bottomrule
    \end{tabular}
    \caption{Quantitative results of 3D shape completion on the test set of ShapeNet (airplane/car/chair). All the values are multiplied by 100.
        We also report the original numbers from \cite{insafutdinov2018unsupervised}.
        Note that they align predicted shapes according to the validation set before evaluating on the test set.
    }
    \label{tab:shapenet}
\end{table}

\begin{figure}[t]
    \centering
    \includegraphics{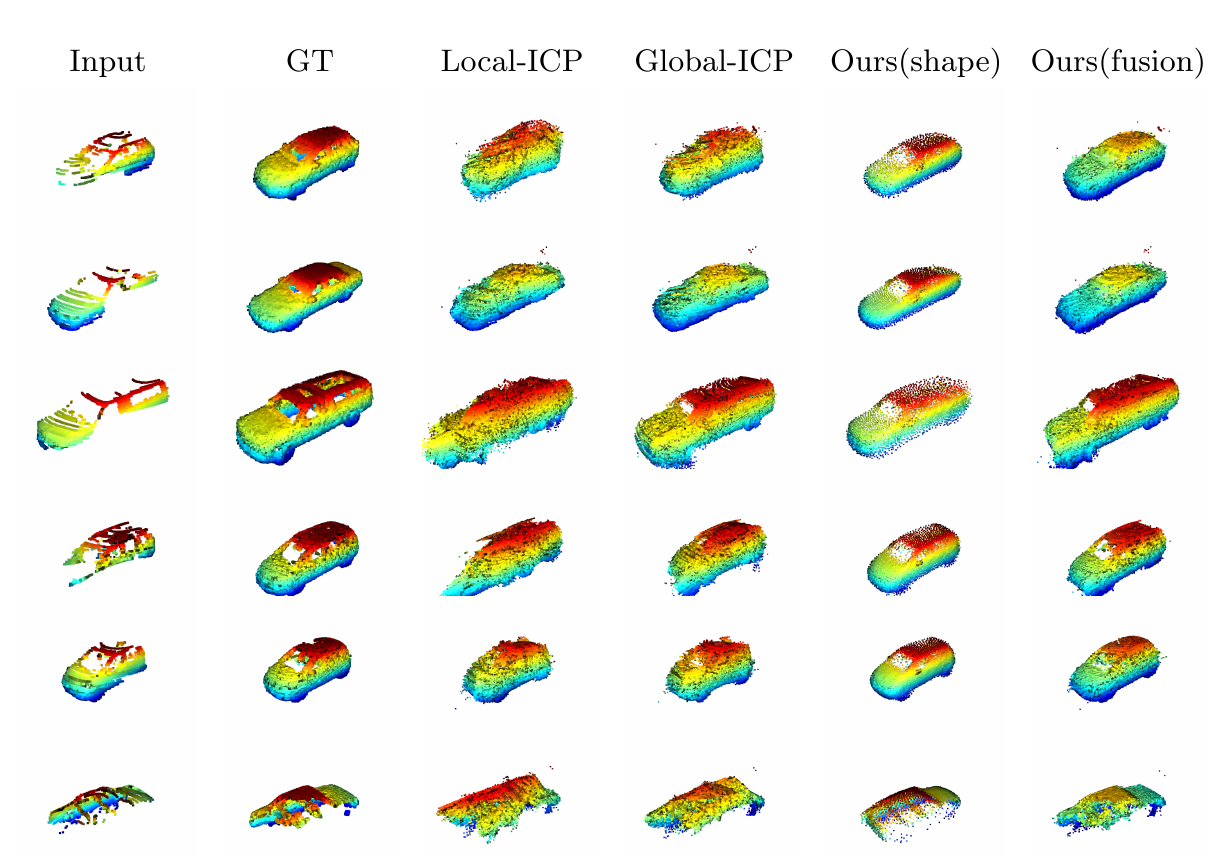}
    \caption{
        Qualitative results of our method compared against ground-truth and ICP on the real datasets (row 1-3: 3D vehicle dataset; row 4-6: SemanticKITTI).
        All the point clouds are transformed to the ground-truth canonical frame and visualized at a fixed viewpoint.
        We denote our approach for 3D shape completion and point cloud registration by \emph{Ours(shape)} and \emph{Ours(fusion)}.
    }
    \label{fig:real-data}
\end{figure}

\myparagraph{Real LiDAR datasets}
We now apply our method to real-world partial LiDAR scans of vehicles.
Table~\ref{tab:tor4d-shape} and \ref{tab:skitti-shape} show the comparison between our method and ICP baselines on real LiDAR datasets.
Fig~\ref{fig:real-data} showcases the qualitative results.
DPC-LIDAR does not converge and performs much worser than our approach, which implies it is better to process point clouds directly rather than project them into 2D planes and rely on existing 2D methods.
Moreover, compared to strong ICP baselines, our method shows higher precision and comparable coverage.
More results and analysis are provided in the supplementary.

\begin{table}[h]
    \small
    \centering
    \setlength{\tabcolsep}{2pt}
    \renewcommand{\arraystretch}{1.1}
    \subfloat[3D vehicle dataset]{
        \begin{tabular}{l|cccc}
            \toprule
                       & CD             & Precision      & Coverage       \\
            \midrule
            DPC-LIDAR  & 0.928          & 0.489          & 0.439          \\
            Local-ICP  & 0.315          & 0.170          & 0.145          \\
            Global-ICP & 0.309          & 0.174          & \textbf{0.135} \\
            Ours       & \textbf{0.255} & \textbf{0.083} & 0.172          \\
            \bottomrule
        \end{tabular}
        \label{tab:tor4d-shape}
    }
    \hfill
    \subfloat[SemanticKITTI]{
        \begin{tabular}{l|cccc}
            \toprule
                       & CD             & Precision      & Coverage       \\
            \midrule
            Local-ICP  & 0.246          & 0.152          & 0.094          \\
            Global-ICP & 0.213          & 0.138          & \textbf{0.075} \\
            Ours       & \textbf{0.194} & \textbf{0.087} & 0.107          \\
            \bottomrule
        \end{tabular}
        \label{tab:skitti-shape}
    }
    \caption{3D shape completion results on the test sets of real LiDAR datasets.}
\end{table}

\subsection{Partial point cloud registration}
In this section, we showcase that our method can be applied to point cloud registration.
It is challenging to align real-world partial point clouds for traditional methods like ICP, due to incompleteness, noise, and sparsity.
Furthermore, even if the transformation between two consecutive frames is accurate, the error may accumulate across frames.
However, our approach alleviates these issues since it implicitly encodes a complete canonical shape.
To evaluate the performance of point cloud registration, we select the middle frame in a sequence as the target, and align other frames to the reference according to estimated poses.
Given a source and a target point cloud, our method predicts $T^{src}_{can}$ and $T^{tgt}_{can}$.
Thus, the transformation from the source to the target is calculated as $T^{tgt}_{src} = T^{tgt}_{can} (T^{src}_{can})^{-1}$.
We report the accuracy, median angle difference, and median MSE between the groundtruth vs. our method, Local-ICP, Global-ICP in Table~\ref{tab:tor4d-pose} and \ref{tab:skitti-pose}.
Fig~\ref{fig:real-data} demonstrates fused point clouds according to estimated poses.
Our method outperforms standard alignment methods.

\begin{table}[!h]
    \small
    \centering
    \setlength{\tabcolsep}{3pt}
    \renewcommand{\arraystretch}{1.1}
    \subfloat[3D vehicle dataset]{
        \begin{tabular}{l|cccc}
            \toprule
                       & Acc            & Rot $\Delta \theta$ & Trans $\Delta t$ \\
            \midrule
            Local-ICP  & 84.09          & 11.33               & 0.30             \\
            Global-ICP & 83.83          & 10.69               & 0.26             \\
            Ours       & \textbf{97.68} & \textbf{2.37}       & \textbf{0.13}    \\
            \bottomrule
        \end{tabular}
        \label{tab:tor4d-pose}
    }
    \hfill
    \subfloat[SemanticKITTI]{
        \begin{tabular}{l|cccc}
            \toprule
                       & Acc            & Rot $\Delta \theta$ & Trans $\Delta t$ \\
            \midrule
            Local-ICP  & 85.29          & 13.04               & 0.31             \\
            Global-ICP & 85.28          & 10.59               & 0.23             \\
            Ours       & \textbf{89.37} & \textbf{2.86}       & \textbf{0.17}    \\
            \bottomrule
        \end{tabular}
        \label{tab:skitti-pose}
    }
    \caption{
        Point cloud registration results on the test sets of real LiDAR datasets.
        We report the median of angle difference and accuracy (the percentage of samples for which $\Delta \theta \leq 30^{\circ}$, as well as the median of translation error $\Delta t$.)
    }
\end{table}

\subsection{Comparison with fully-supervised counterparts}
\label{sec:full-sup}
Our method does not rely on any ground-truth shape and pose, or any prior knowledge of where the object is located, \eg, the bounding box.
Yet, it can still reconstruct the 3D shape reliably and accurately.
In order to understand the upper bound of our method for shape completion, we include an \emph{oracle} baseline, where our model is trained with ground-truth complete shapes.
Concretely, during training, given a partial point cloud in the \emph{sensor} coordinate system as input, we employ the same network to encode the input and decode the canonical complete shape without estimating the pose.
The Chamfer Distance is calculated between the output canonical shape and the ground-truth canonical complete shape.
Note that our method with full supervision is identical to \emph{PCN-FC}~\cite{yuan2018pcn}, except for unaligned point clouds as input.

Table~\ref{tab:oracle} shows that there exists a gap between our weakly-supervised approach and its fully-supervised counterpart.
However, the gap is even smaller than that between ours and DPC~\cite{insafutdinov2018unsupervised}.
The Chamfer Distance of the fully-supervised oracle is almost half of that of our weakly-supervised approach.
This may be due to the fact that the LiDAR sensor will only see half of the car by which it passes, and therefore the partial LiDAR observations alone are insufficient to see the other side of the car.
To solve this issue, prior knowledge of the category may help, which we leave for future work.

\begin{table}[t]
    \small
    \centering
    \setlength{\tabcolsep}{3pt}
    \begin{tabular}{l|c|ccc}
        \toprule
        Category & Method          & CD    & Precision & Coverage \\
        \midrule
        \multirow{2}{*}{Airplane}
                 & Ours            & 1.95  & 0.91      & 1.05     \\
                 & Ours(Full-Sup)  & 1.65  & 0.77      & 0.88     \\
        \midrule
        \multirow{2}{*}{Car}
                 & Ours            & 2.68  & 1.27      & 1.41     \\
                 & Ours(Full-Sup)  & 2.04  & 1.01      & 1.03     \\
        \midrule
        \multirow{2}{*}{Chair}
                 & Ours            & 3.33  & 1.69      & 1.64     \\
                 & Ours(Full-Sup)  & 2.82  & 1.47      & 1.35     \\
        \midrule
        \multirow{2}{*}{Real vehicle}
                 & Ours            & 0.255 & 0.083     & 0.172    \\
                 & Ours (Full-Sup) & 0.140 & 0.064     & 0.077    \\
        \bottomrule
    \end{tabular}
    \caption{
        Shape completion results on the test sets of ShapeNet and our 3D vehicle dataset.
        All the values for ShapeNet categories are multiplied by 100.
        The full-supervision oracle is denoted by \emph{Ours(Full-Sup)}.
    }
    \label{tab:oracle}
\end{table}

\section{Discussion and Future Work}
We have proposed a new setting, weakly-supervised 3D shape completion in the wild, which better captures the realistic scenario of being able to infer unknown shape from real world scans of objects.
We demonstrate that this challenging problem can be tackled by jointly learning both shape and pose with multi-view consistency.
However, there remains much space to improve and explore.
From visualization, we observe that the model tends to generate coarse shapes and miss details, due to the noise of pose estimation.
It is also observed in training that the loss calculated on point clouds is more sensitive to the density compared to that using 2D projection.
More efforts can be made to improve visual quality and narrow the gap with fully-supervised methods.
Besides, we use PointNet as the backbone for simplicity and efficiency in this work.
Differently designed networks can be applied to predict shapes and poses separately.
Furthermore, our approach currently requires knowing multiple views of a single rigid object.
Such ``annotations'' can be acquired by a 3D detector and tracker.
Thus, one future direction is to study self-supervised or weakly-supervised 3D detection and tracking.

\clearpage
\bibliographystyle{splncs04}
\bibliography{egbib}
\end{document}


\pagestyle{headings}
\mainmatter
\def\ECCVSubNumber{3125}  %

\title{Supplementary Material -- Weakly-supervised 3D Shape Completion in the Wild} %

\titlerunning{Weakly-supervised 3D Shape Completion in the Wild}

\author{
    Jiayuan Gu\inst{1,2} \and
    Wei-Chiu Ma\inst{1,3} \and
    Sivabalan Manivasagam\inst{1,4} \and
    Wenyuan Zeng\inst{1,4} \and
    Zihao Wang\inst{1} \and
    Yuwen Xiong\inst{1,4} \and
    Hao Su\inst{2} \and
    Raquel Urtasun\inst{1,4}
}

\authorrunning{J. Gu et al.}
\institute{
    Uber Advanced Technologies Group \and
    University of California, San Diego \\ \email{\{jigu, haosu\}@eng.ucsd.edu} \and
    Massachusetts Institute of Techonology \and
    University of Toronto \\ \email{\{wichiu, manivasagam, wenyuan, yuwen, urtasun\}@uber.com}
}
\maketitle

\section{Overview}
This supplementary material provides more detailed and thorough analysis of our weakly-supervised approach for 3D shape completion.
We hope readers can gain more insights into our approach.
Sec~\ref{sec:ablation} presents ablation studies to analyze our design.
We report the results of partial point cloud registration on ShapeNet in Sec~\ref{sec:shapenet-pose}, to show more quantitative comparison.
Moreover, we showcase an experiment where the model is fine-tuned on another category in the wild during inference in Sec~\ref{sec:skitti-truck}.
Sec~\ref{sec:more-qualitative} shows more visual comparison on both synthetic and real LiDAR datasets.
Last but not least, the sensitivity to initialization is investigated in Sec~\ref{sec:sensitivity}.

\section{Ablation studies}
\label{sec:ablation}
For ablation studies, we investigate several factors:
1) the shape-projection-matching-observation term,
2) the hindsight loss.
Table~\ref{tab:shapenet-ablation} shows the quantitative results on ShapeNet.
It is observed that:
1) Without the shape-projection-matching-observation term, the chamfer distance and precision increase while the coverage decreases.
It shows the effectiveness of our proposed projection approach, and verifies that the observation-matching-shape term only is not enough as it can not force the generated shape to be close to the observation.
On our 3D vehicle dataset, the shape-projection-matching-observation term decreases the precision but increases the coverage, which results in the chamfer distance similar to that without it.
However, the loss term can improve visual results.
2) Without the hindsight loss, the network is vulnerable to local minima, and performs worse.

\begin{table}[ht]
    \small
    \centering
    \setlength{\tabcolsep}{2pt}
    \renewcommand{\arraystretch}{1.1}
    \begin{tabular}{l|c|ccc}
        \toprule
        Category & Method         & CD   & Precision & Coverage \\
        \midrule
        \multirow{2}{*}{Airplane}
                 & w/o projection & 2.80 & 1.98      & 0.82     \\
                 & w/o hindsight  & 2.32 & 1.18      & 1.14     \\
                 & full           & 1.65 & 0.77      & 0.88     \\
        \midrule
        \multirow{2}{*}{Car}
                 & w/o projection & 2.96 & 1.67      & 1.29     \\
                 & w/o hindsight  & 2.82 & 1.46      & 1.36     \\
                 & full           & 2.68 & 1.27      & 1.41     \\
        \midrule
        \multirow{2}{*}{Chair}
                 & w/o projection & 3.94 & 2.50      & 1.44     \\
                 & w/o hindsight  & 3.80 & 2.09      & 1.71     \\
                 & full           & 3.33 & 1.69      & 1.64     \\
        \bottomrule
    \end{tabular}
    \caption{
        Ablation studies on ShapeNet.
        We report shape completion results on the test set.
        All the values are multiplied by 100.
    }
    \label{tab:shapenet-ablation}
\end{table}

In addition, we investigate the relation between the performance and the number of views during training.
Table~\ref{tab:number-of-views} shows results on our 3D vehicle dataset, w.r.t numbers of views.
With the same number of instances in a batch, the more the number of views, the better the performance is.
We select 4 views per instance as a trade-off between the performance and the computation.

\begin{table}[ht]
    \small
    \centering
    \setlength{\tabcolsep}{2pt}
    \renewcommand{\arraystretch}{1.1}
    \begin{tabular}{cc|ccc}
        \toprule
        \#views & \#inst & CD    & Rot $\Delta \theta$ & Trans $\Delta t$ \\
        \midrule
        2       & 8      & 0.307 & 6.185        & 0.213      \\
        4       & 8      & 0.261 & 4.208        & 0.160      \\
        8       & 8      & 0.242 & 3.995        & 0.142      \\
        \bottomrule
    \end{tabular}
    \caption{
        Ablation studies on our 3D vehicle dataset w.r.t different numbers of views.
        Note that we report an average of 5 trials instead of the best trial here.
    }
    \label{tab:number-of-views}
\end{table}

\section{Point cloud registration on ShapeNet}
\label{sec:shapenet-pose}

\begin{figure}[ht]
    \centering
    \begin{subfigure}[t]{0.18\textwidth}
        \caption*{Input(Ours)}
        \vspace{-0.5em}
        \includegraphics[width=\textwidth]{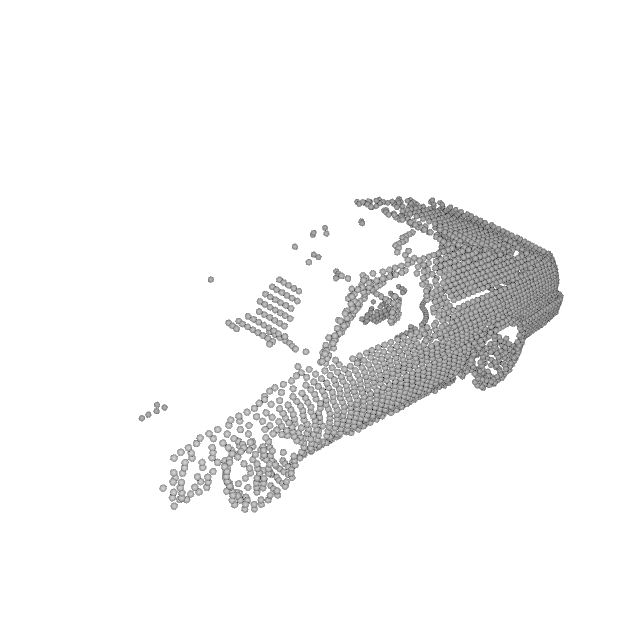}
    \end{subfigure}
    \begin{subfigure}[t]{0.18\textwidth}
        \caption*{GT}
        \vspace{-0.5em}
        \includegraphics[width=\textwidth]{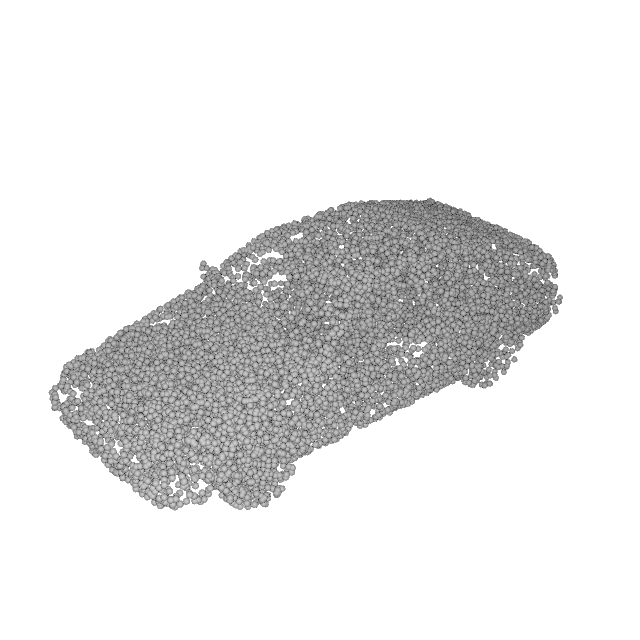}
    \end{subfigure}
    \begin{subfigure}[t]{0.18\textwidth}
        \caption*{DPC}
        \vspace{-0.5em}
        \includegraphics[width=\textwidth]{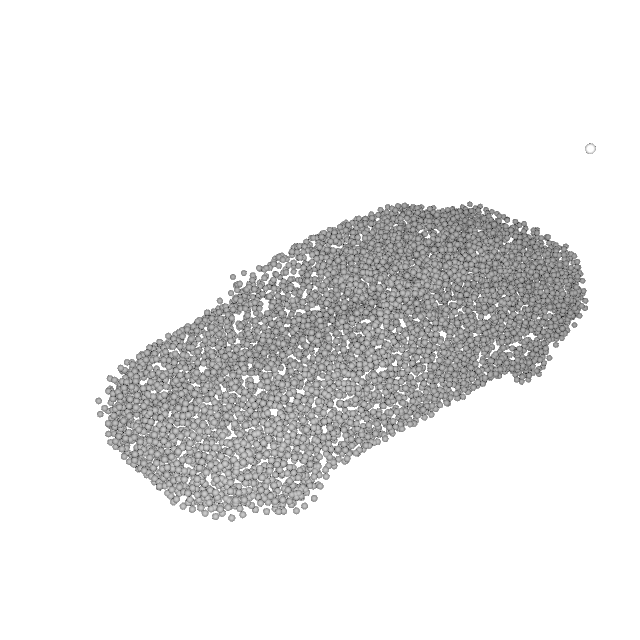}
    \end{subfigure}
    \begin{subfigure}[t]{0.18\textwidth}
        \caption*{Ours}
        \vspace{-0.5em}
        \includegraphics[width=\textwidth]{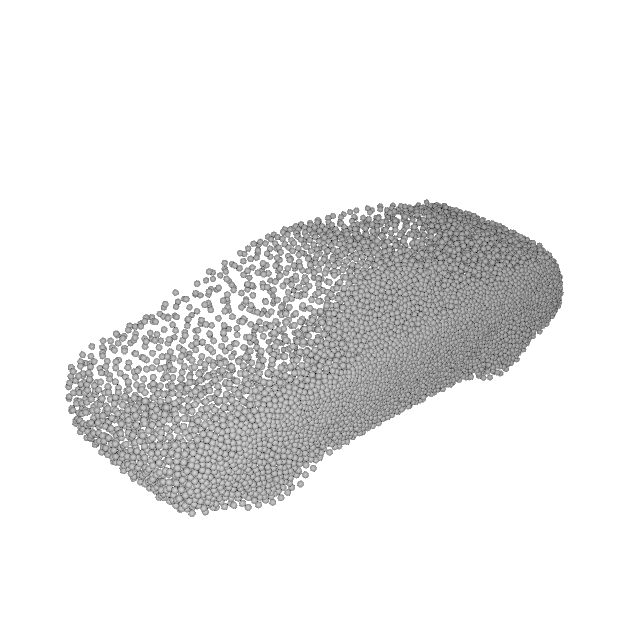}
    \end{subfigure}
    \begin{subfigure}[t]{0.18\textwidth}
        \caption*{Ours*}
        \vspace{-0.5em}
        \includegraphics[width=\textwidth]{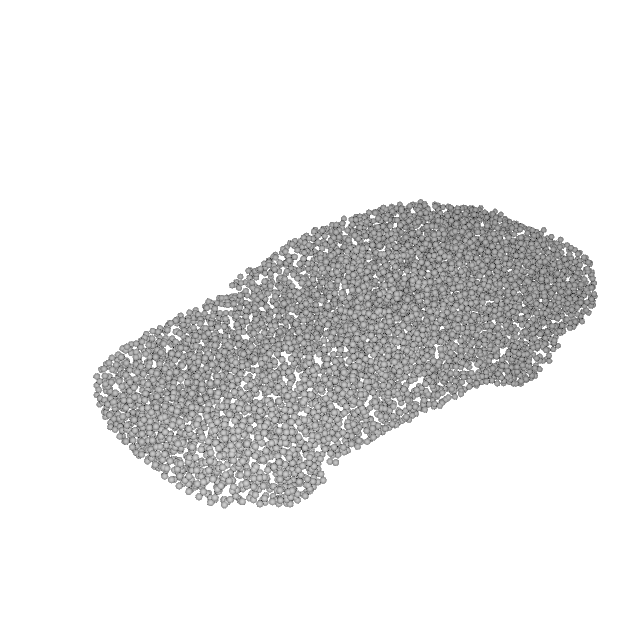}
    \end{subfigure}

    \vspace{-1.0em}

    \begin{subfigure}[t]{0.18\textwidth}
        \includegraphics[width=\textwidth]{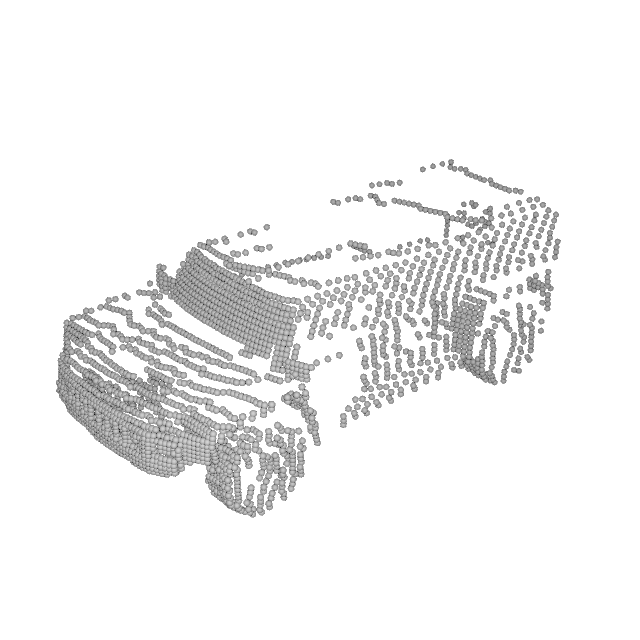}
    \end{subfigure}
    \begin{subfigure}[t]{0.18\textwidth}
        \includegraphics[width=\textwidth]{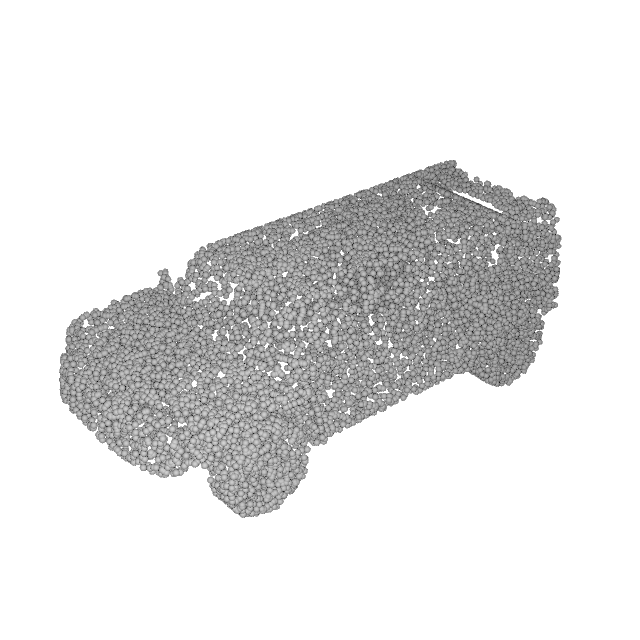}
    \end{subfigure}
    \begin{subfigure}[t]{0.18\textwidth}
        \includegraphics[width=\textwidth]{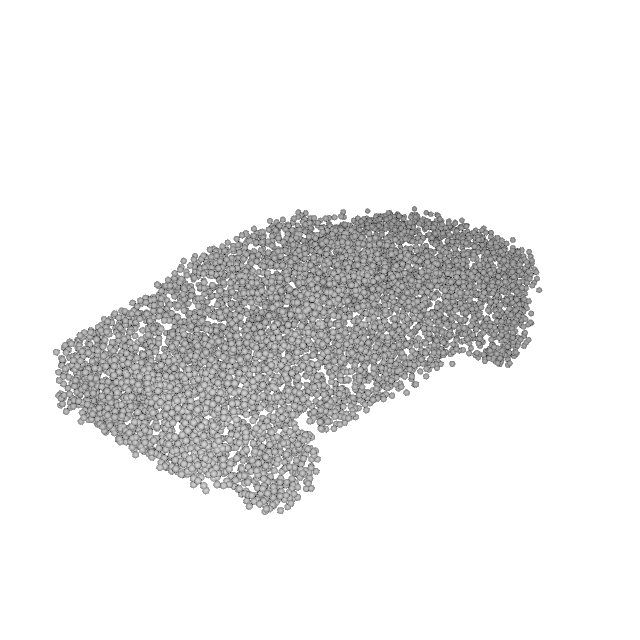}
    \end{subfigure}
    \begin{subfigure}[t]{0.18\textwidth}
        \includegraphics[width=\textwidth]{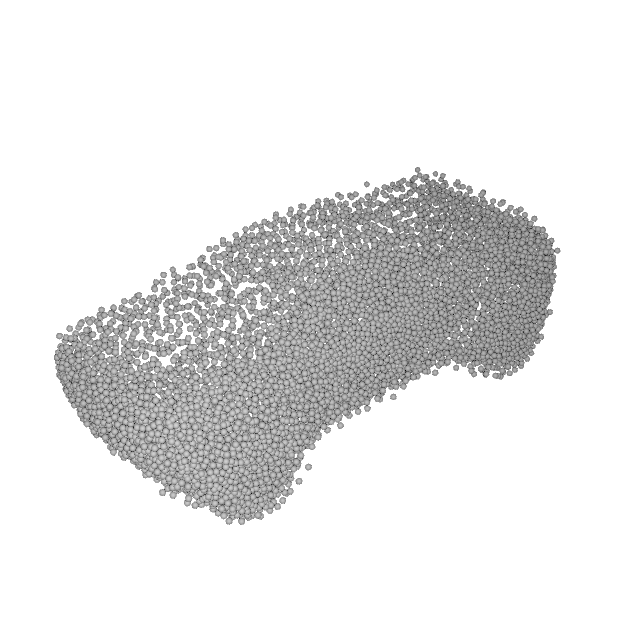}
    \end{subfigure}
    \begin{subfigure}[t]{0.18\textwidth}
        \includegraphics[width=\textwidth]{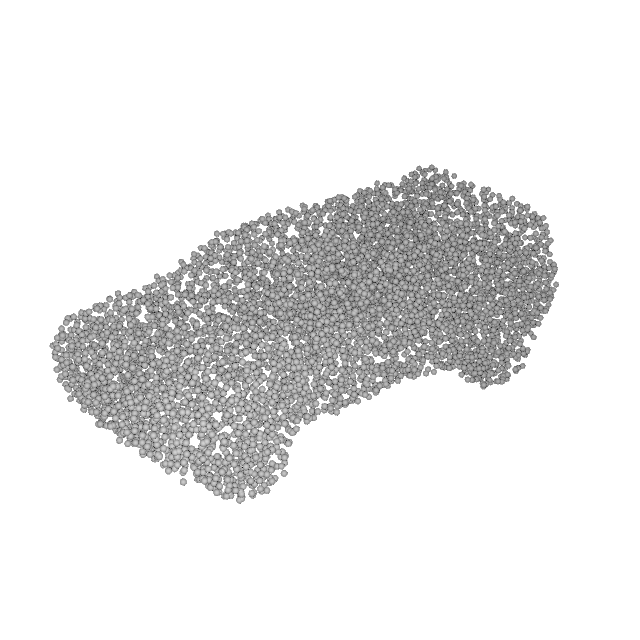}
    \end{subfigure}

    \vspace{-1.0em}

    \begin{subfigure}[t]{0.18\textwidth}
        \includegraphics[width=\textwidth]{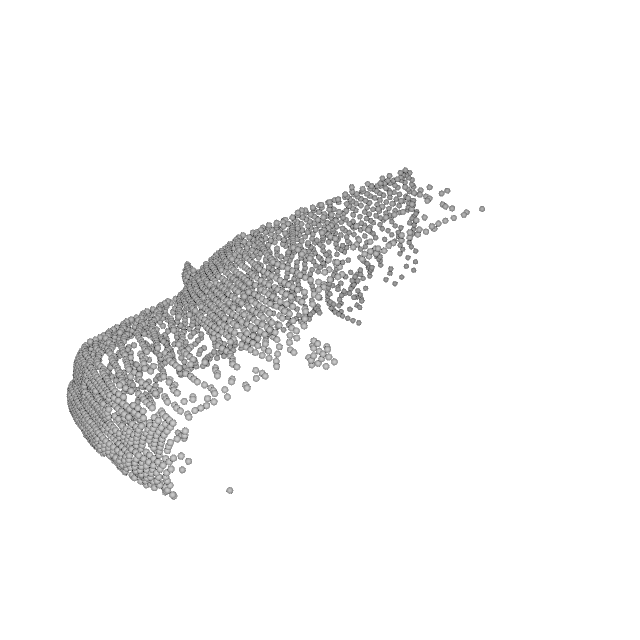}
    \end{subfigure}
    \begin{subfigure}[t]{0.18\textwidth}
        \includegraphics[width=\textwidth]{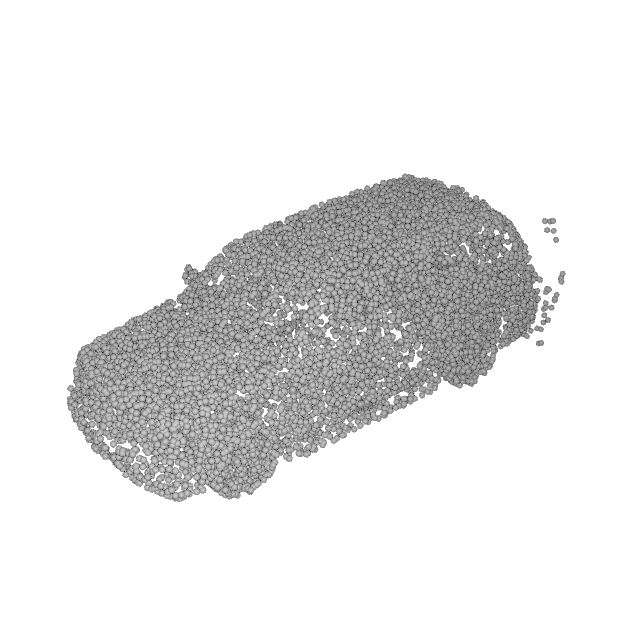}
    \end{subfigure}
    \begin{subfigure}[t]{0.18\textwidth}
        \includegraphics[width=\textwidth]{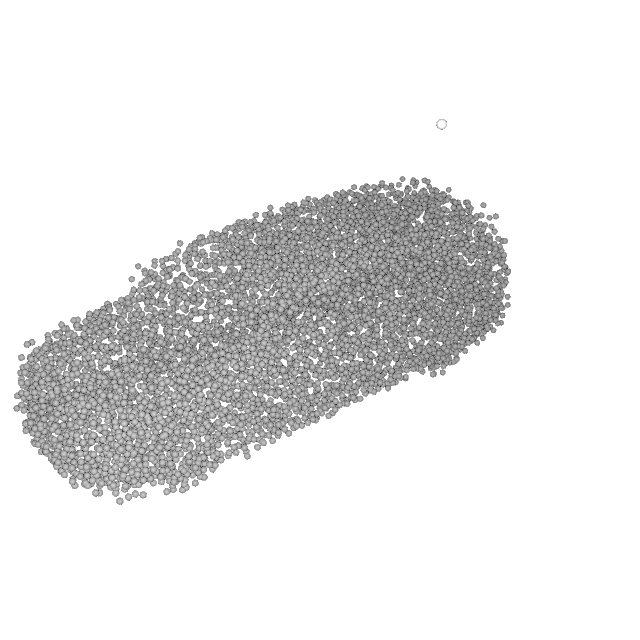}
    \end{subfigure}
    \begin{subfigure}[t]{0.18\textwidth}
        \includegraphics[width=\textwidth]{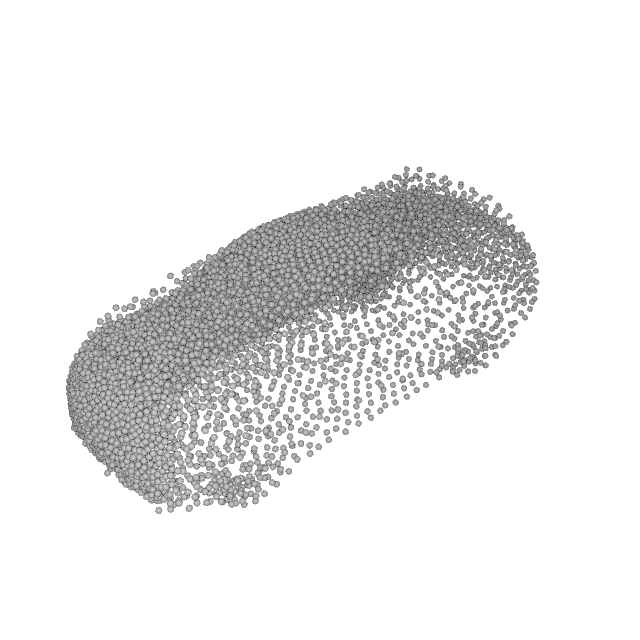}
    \end{subfigure}
    \begin{subfigure}[t]{0.18\textwidth}
        \includegraphics[width=\textwidth]{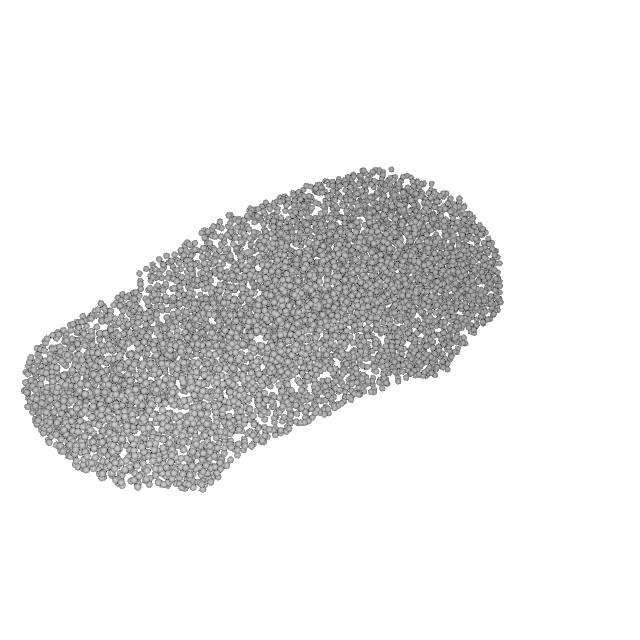}
    \end{subfigure}

    \vspace{-1.0em}

    \begin{subfigure}[t]{0.18\textwidth}
        \includegraphics[width=\textwidth]{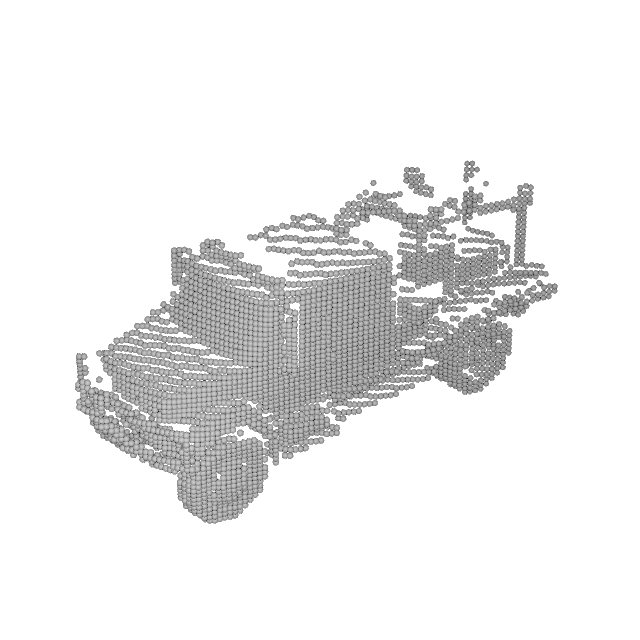}
    \end{subfigure}
    \begin{subfigure}[t]{0.18\textwidth}
        \includegraphics[width=\textwidth]{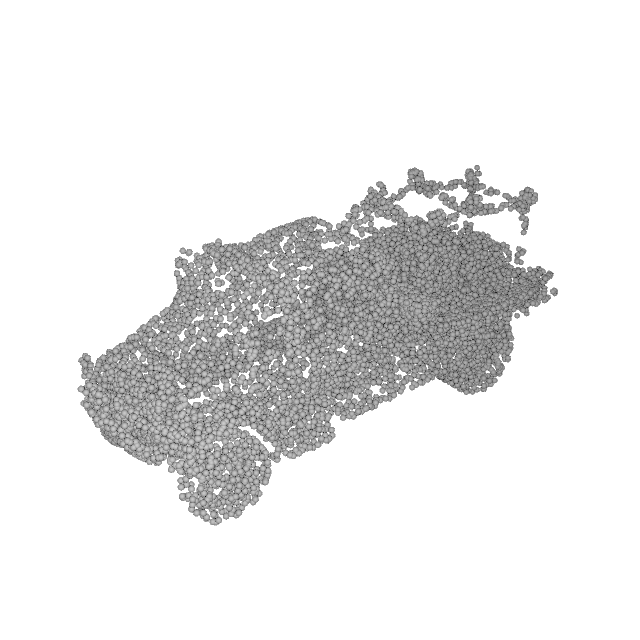}
    \end{subfigure}
    \begin{subfigure}[t]{0.18\textwidth}
        \includegraphics[width=\textwidth]{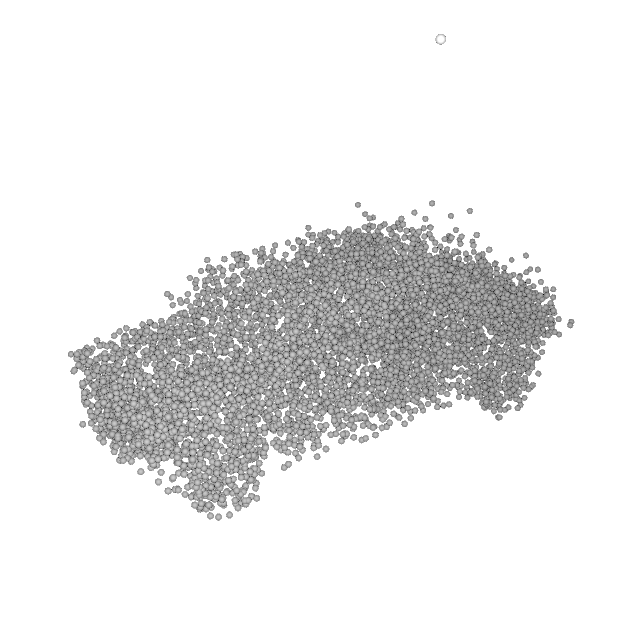}
    \end{subfigure}
    \begin{subfigure}[t]{0.18\textwidth}
        \includegraphics[width=\textwidth]{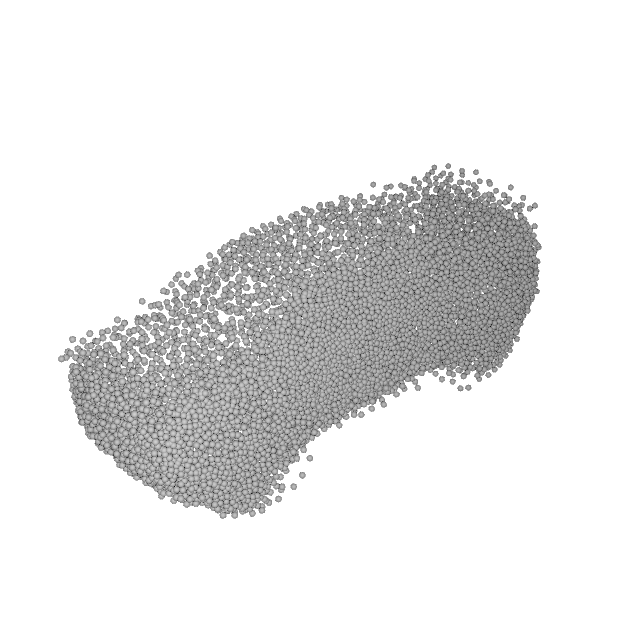}
    \end{subfigure}
    \begin{subfigure}[t]{0.18\textwidth}
        \includegraphics[width=\textwidth]{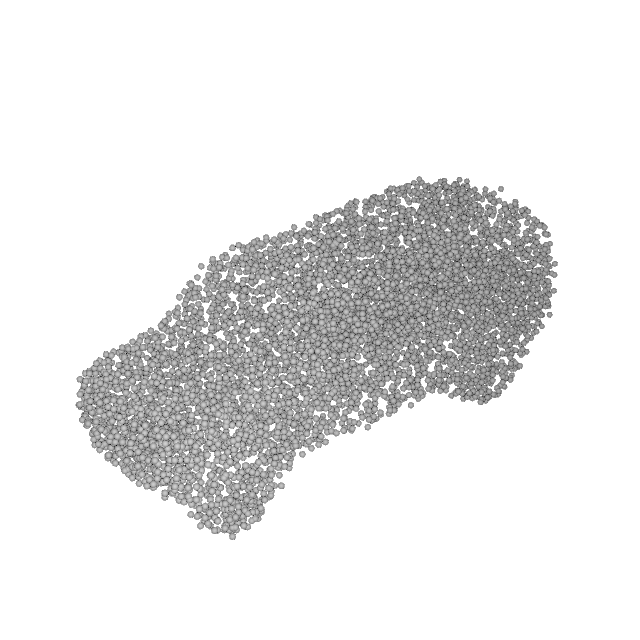}
    \end{subfigure}

    \caption{
        Qualitative results of 3D shape completion on the test set of ShapeNet.
        All the point clouds are transformed to the ground-truth canonical frame and visualized at a fixed viewpoint.
    }
    \label{fig:ours-2d}
\end{figure}

In the main paper, we have showcased that our approach can be extended to challenging partial point cloud registration on real datasets.
In this section, we demonstrate the results of this task on ShapeNet.
Concretely, we compare the relative pose between one view and the target view against the ground truth relative pose.
We argue that our evaluation protocal for pose estimation is better than that in DPC~\cite{insafutdinov2018unsupervised}, as they measure the pose error by first aligning the canonical pose learned with the groundtruth using ICP.
Compared to real datasets with over 80 scans per instance,
it is even challenging for synthetic data, since there are only 5 views per object in total for training.

We report the accuracy, median angle difference, and median translation MSE of our method, DPC, DPC$^\dagger$ in Table~\ref{tab:shapenet-pose}.
Our approach outperforms DPC and DPC$^\dagger$ by a large margin on all the categories.
For cars, we use a variant of our approach, where input and output points are both projected into 2D points and the chamfer distance between 2D projections is optimized.
Unlike chairs and planes, the front and back of cars look similar, which introduces more pose ambiguity and results in an oversmoothed canonical shape.
Thus, the variant is proposed to tackle the pose ambiguity caused by the symmetry of cars.
Fig~\ref{fig:ours-2d} shows the comparison between the variant (\emph{Ours$^*$}) and the original implementation.

\begin{table}[ht]
    \small
    \centering
    \setlength{\tabcolsep}{2pt}
    \renewcommand{\arraystretch}{1.1}
    \begin{tabular}{l|c|ccc}
        \toprule
        Category & Method        & Acc($\Delta \theta \leq 30^{\circ}$)            & Rot $\Delta \theta$ & Trans $\Delta t$ \\
        \midrule
        \multirow{2}{*}{Airplane}
                 & DPC           & 74.17          & 9.95                & -                \\
                 & DPC$^\dagger$ & 55.64          & 23.85               & 0.13             \\
                 & Ours          & \textbf{92.87} & \textbf{1.87}       & \textbf{0.01}    \\
        \midrule
        \multirow{2}{*}{Car}
                 & DPC           & 84.75          & 6.40                & -                \\
                 & DPC$^\dagger$ & 82.17          & 8.79                & 0.05             \\
                 & Ours*         & \textbf{91.03} & \textbf{2.46}       & -                \\
        \midrule
        \multirow{2}{*}{Chair}
                 & DPC           & 80.02          & 10.96               & -                \\
                 & DPC$^\dagger$ & 70.45          & 10.17               & 0.07             \\
                 & Ours          & \textbf{95.82} & \textbf{2.31}       & \textbf{0.02}    \\
        \bottomrule
    \end{tabular}
    \caption{Point cloud registration results on the test set of ShapeNet.
        \emph{Ours*} computes losses on projected input and output points.}
    \label{tab:shapenet-pose}
\end{table}

\section{Fine-tuning during inference}
\label{sec:skitti-truck}

\begin{figure}[ht]
    \centering
    \begin{subfigure}[t]{0.24\textwidth}
        \caption*{Input}
        \vspace{-0.5em}
        \includegraphics[width=\textwidth]{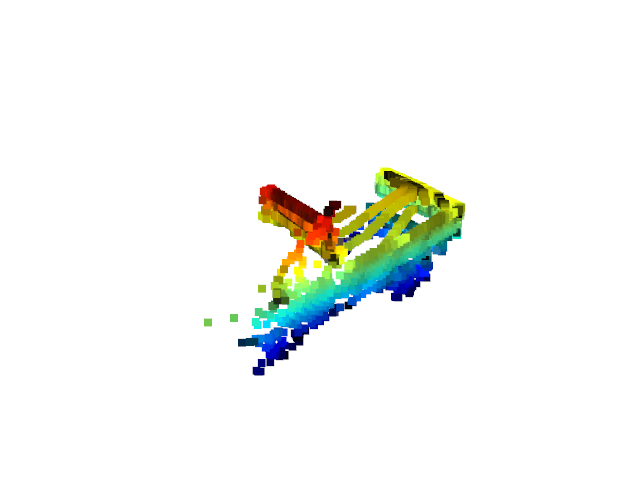}
    \end{subfigure}
    \begin{subfigure}[t]{0.24\textwidth}
        \caption*{Ground truth}
        \vspace{-0.5em}
        \includegraphics[width=\textwidth]{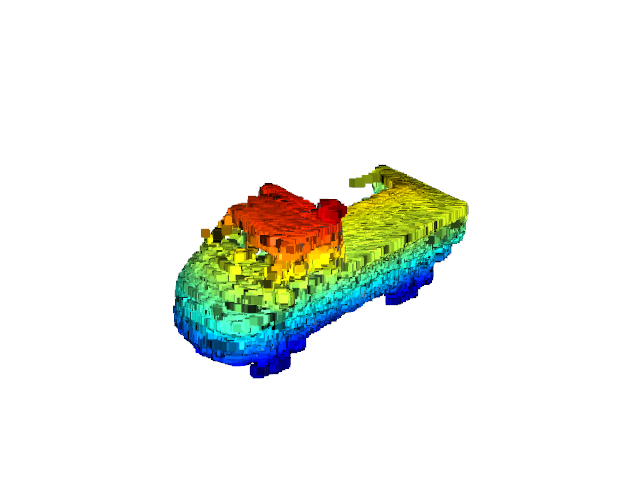}
    \end{subfigure}
    \begin{subfigure}[t]{0.24\textwidth}
        \caption*{Ours(shape)}
        \vspace{-0.5em}
        \includegraphics[width=\textwidth]{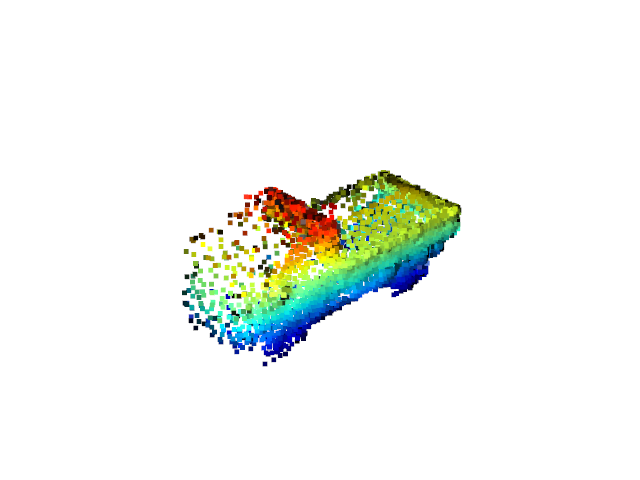}
    \end{subfigure}
    \begin{subfigure}[t]{0.24\textwidth}
        \caption*{Ours(fusion)}
        \vspace{-0.5em}
        \includegraphics[width=\textwidth]{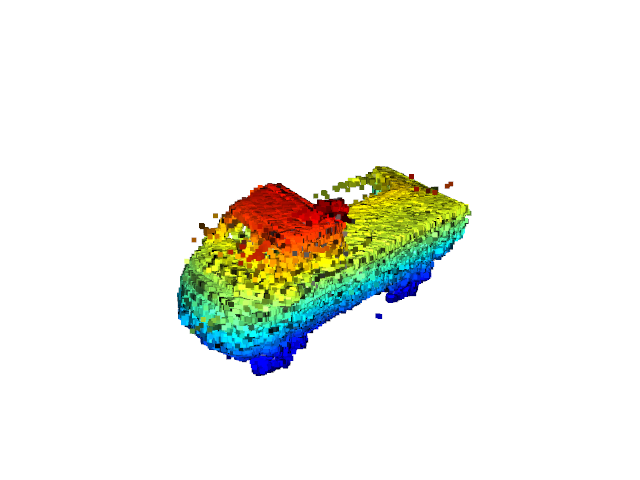}
    \end{subfigure}
    
    \vspace{-1.5em}
    
    \begin{subfigure}[t]{0.24\textwidth}
        \includegraphics[width=\textwidth]{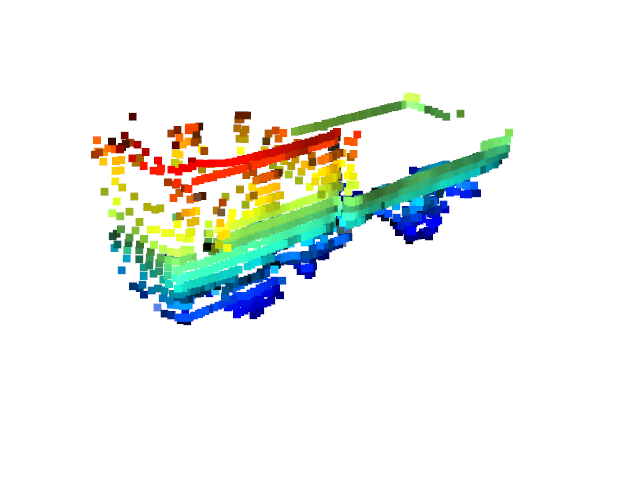}
    \end{subfigure}
    \begin{subfigure}[t]{0.24\textwidth}
        \includegraphics[width=\textwidth]{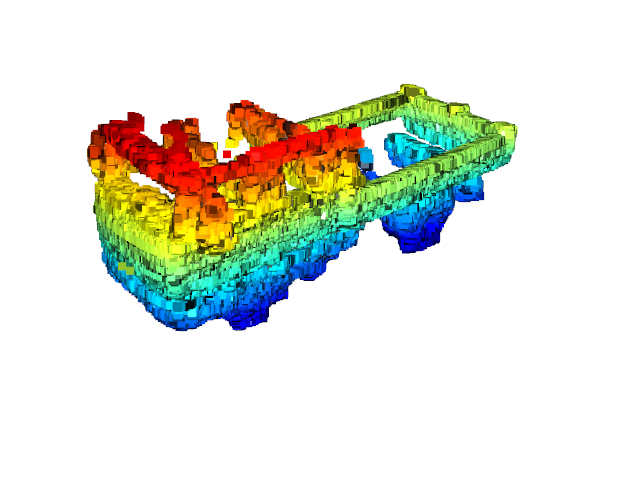}
    \end{subfigure}
    \begin{subfigure}[t]{0.24\textwidth}
        \includegraphics[width=\textwidth]{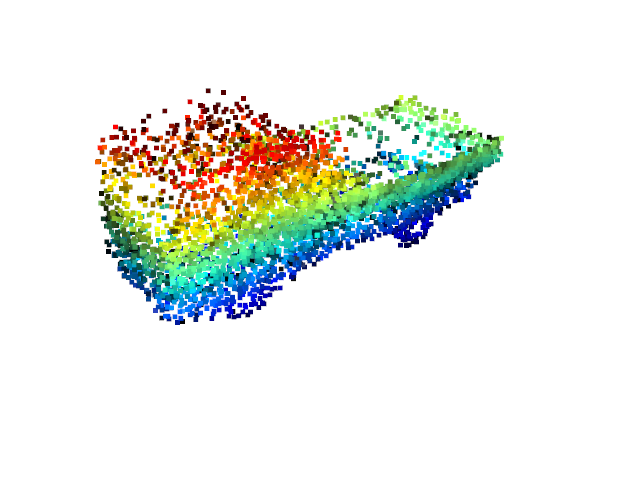}
    \end{subfigure}
    \begin{subfigure}[t]{0.24\textwidth}
        \includegraphics[width=\textwidth]{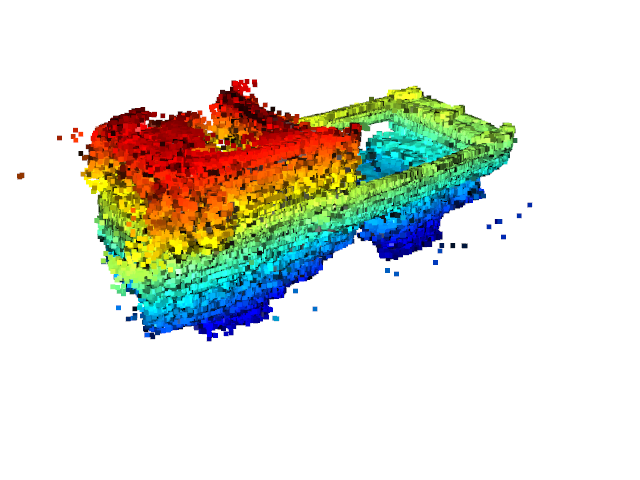}
    \end{subfigure}
    
    \vspace{-1.5em}
    
    \begin{subfigure}[t]{0.24\textwidth}
        \includegraphics[width=\textwidth]{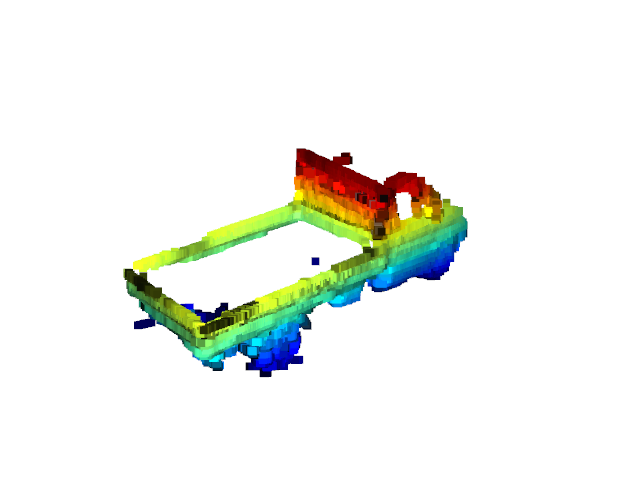}
    \end{subfigure}
    \begin{subfigure}[t]{0.24\textwidth}
        \includegraphics[width=\textwidth]{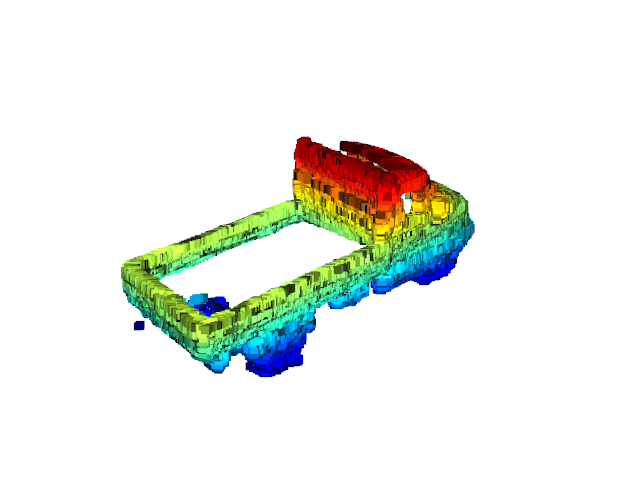}
    \end{subfigure}
    \begin{subfigure}[t]{0.24\textwidth}
        \includegraphics[width=\textwidth]{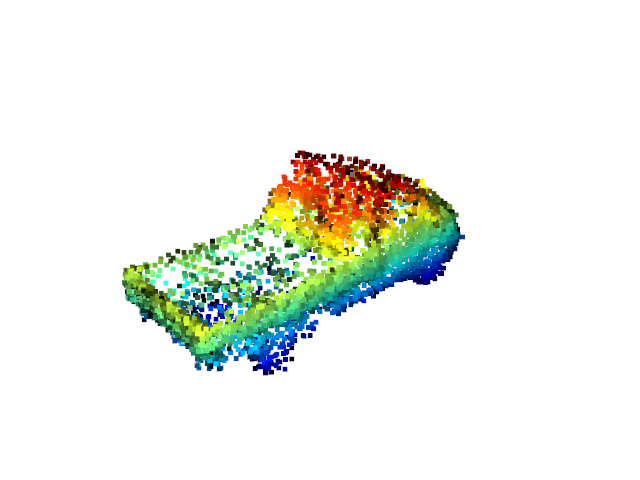}
    \end{subfigure}
    \begin{subfigure}[t]{0.24\textwidth}
        \includegraphics[width=\textwidth]{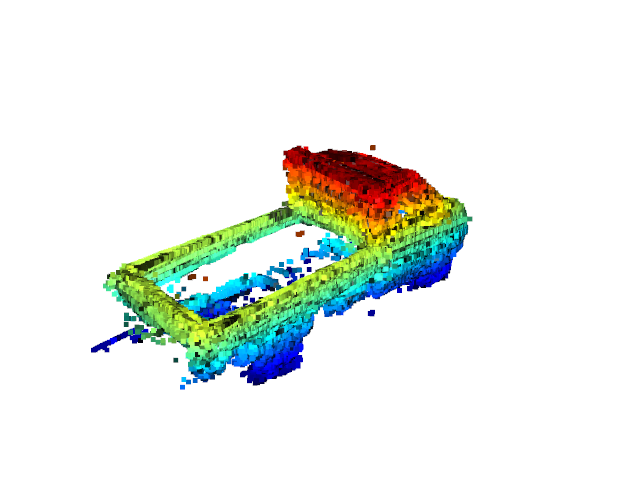}
    \end{subfigure}
    
    \vspace{-1.5em}
    
    \begin{subfigure}[t]{0.24\textwidth}
        \includegraphics[width=\textwidth]{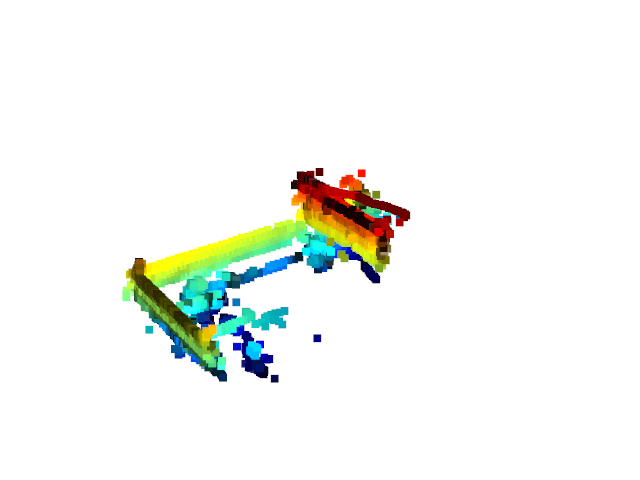}
    \end{subfigure}
    \begin{subfigure}[t]{0.24\textwidth}
        \includegraphics[width=\textwidth]{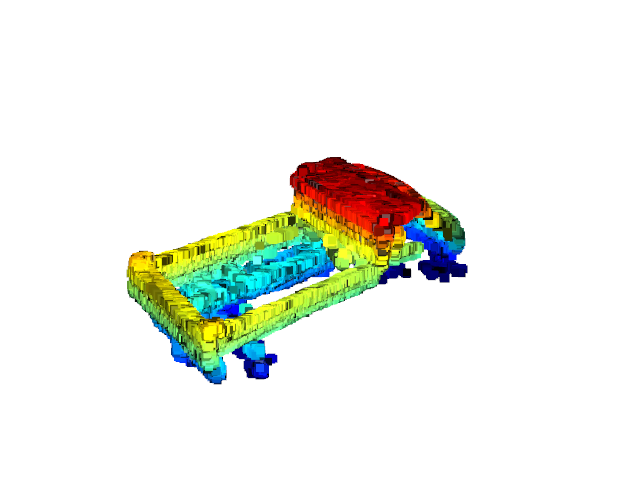}
    \end{subfigure}
    \begin{subfigure}[t]{0.24\textwidth}
        \includegraphics[width=\textwidth]{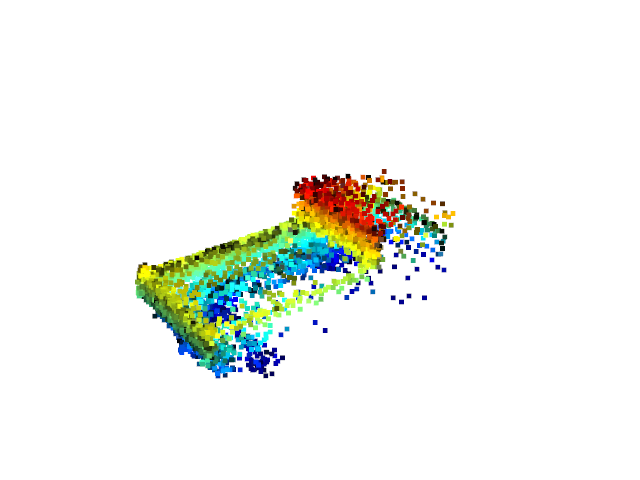}
    \end{subfigure}
    \begin{subfigure}[t]{0.24\textwidth}
        \includegraphics[width=\textwidth]{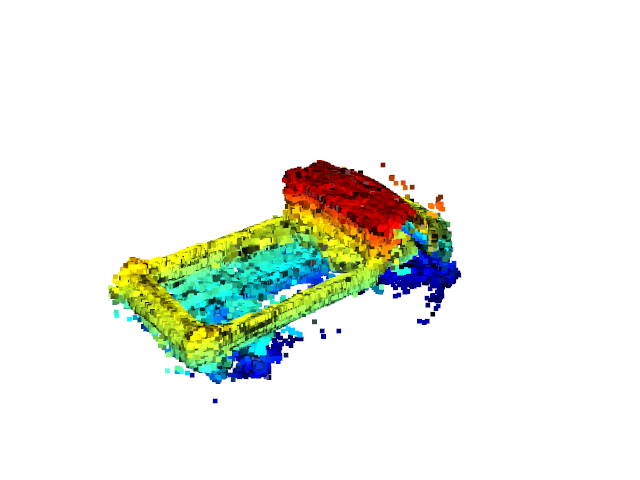}
    \end{subfigure}

    \caption{Qualitative results of our model fine-tuned on SemanticKITTI trucks.
        All the point clouds are transformed to the ground-truth canonical frame and visualized at a fixed viewpoint.
        We denote our approach for 3D shape completion and point cloud registration by \emph{Ours(shape)} and \emph{Ours(fusion)}.
    }
    \label{fig:skitti-truck}
\end{figure}

To demonstrate that our method can be applied to other categories in the wild, we experiment on parked trucks of Semantic KITTI.
Due to the limited amount of data (14 valid instances), we fine-tuned the model pre-trained on our 3D vehicle dataset.
The CD is 0.2942.
The pose accuracy is 86.74, the median angle difference is 2.08, and the median translation MSE is 0.15.
It indicates the flexibility of our method, which can be optimized during inference.
Some examples are visualized in Fig~\ref{fig:skitti-truck}.

\section{Clarification for the GT of our 3D vehicle dataset}
Note that we leverage symmetry to generate ground truth complete shapes of our 3D vehicle dataset.
However, for SemanticKITTI, due to lack of GT boxes, we use the point clouds fused over frames as ``partial'' GT.
Thus, we provide the quantitative results of shape completion on our 3D vehicle dataset evaluated by ``partial'' GT.
The chamfer distance of our method improves from 0.255 to 0.195, while local ICP and global ICP improve from 0.315 to 0.275 and from 0.309 to 0.274 respectively.
The ranking among different methods remains the same.
The performance of point cloud registration is not affected.

\section{More qualitative results}
\label{sec:more-qualitative}

\begin{figure}[h]
    \centering
    \begin{subfigure}[t]{0.15\textwidth}
        \caption*{Input(Ours)}
        \vspace{-0.5em}
        \includegraphics[width=\textwidth]{figures/shapenet_supp/car/929e5379fd9b49b8730895dbb38ad66f_0_input.png}
    \end{subfigure}
    \begin{subfigure}[t]{0.15\textwidth}
        \caption*{Input(DPC)}
        \vspace{-0.5em}
        \includegraphics[width=\textwidth]{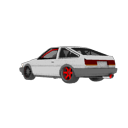}
    \end{subfigure}
    \begin{subfigure}[t]{0.15\textwidth}
        \caption*{GT}
        \vspace{-0.5em}
        \includegraphics[width=\textwidth]{figures/shapenet_supp/car/929e5379fd9b49b8730895dbb38ad66f_0_gt.png}
    \end{subfigure}
    \begin{subfigure}[t]{0.15\textwidth}
        \caption*{DPC}
        \vspace{-0.5em}
        \includegraphics[width=\textwidth]{figures/shapenet_supp/car/929e5379fd9b49b8730895dbb38ad66f_0_dpc.png}
    \end{subfigure}
    \begin{subfigure}[t]{0.15\textwidth}
        \caption*{DPC$^\dagger$}
        \vspace{-0.5em}
        \includegraphics[width=\textwidth]{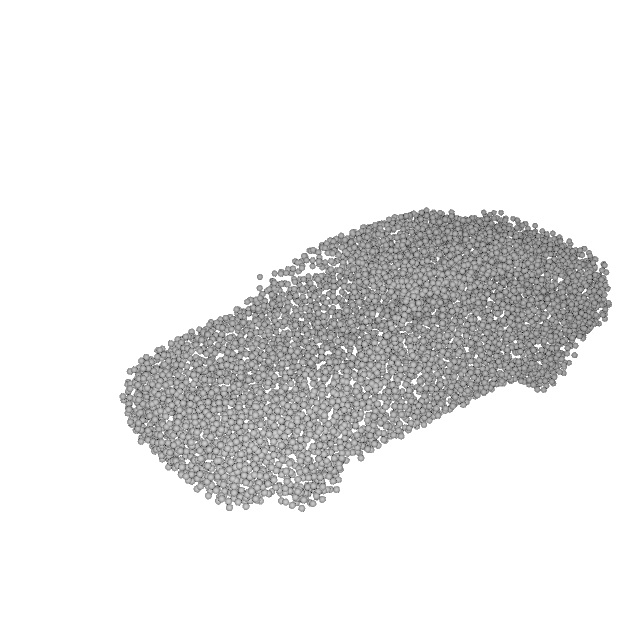}
    \end{subfigure}
    \begin{subfigure}[t]{0.15\textwidth}
        \caption*{Ours}
        \vspace{-0.5em}
        \includegraphics[width=\textwidth]{figures/shapenet_supp/car2/929e5379fd9b49b8730895dbb38ad66f_0_ours2.png}
    \end{subfigure}

    \vspace{-0.5em}

    \begin{subfigure}[t]{0.15\textwidth}
        \includegraphics[width=\textwidth]{figures/shapenet_supp/car/963a4fdf819cc5ab3174b45571ecff3d_2_input.png}
    \end{subfigure}
    \begin{subfigure}[t]{0.15\textwidth}
        \includegraphics[width=\textwidth]{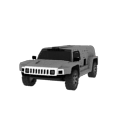}
    \end{subfigure}
    \begin{subfigure}[t]{0.15\textwidth}
        \includegraphics[width=\textwidth]{figures/shapenet_supp/car/963a4fdf819cc5ab3174b45571ecff3d_2_gt.png}
    \end{subfigure}
    \begin{subfigure}[t]{0.15\textwidth}
        \includegraphics[width=\textwidth]{figures/shapenet_supp/car/963a4fdf819cc5ab3174b45571ecff3d_2_dpc.png}
    \end{subfigure}
    \begin{subfigure}[t]{0.15\textwidth}
        \includegraphics[width=\textwidth]{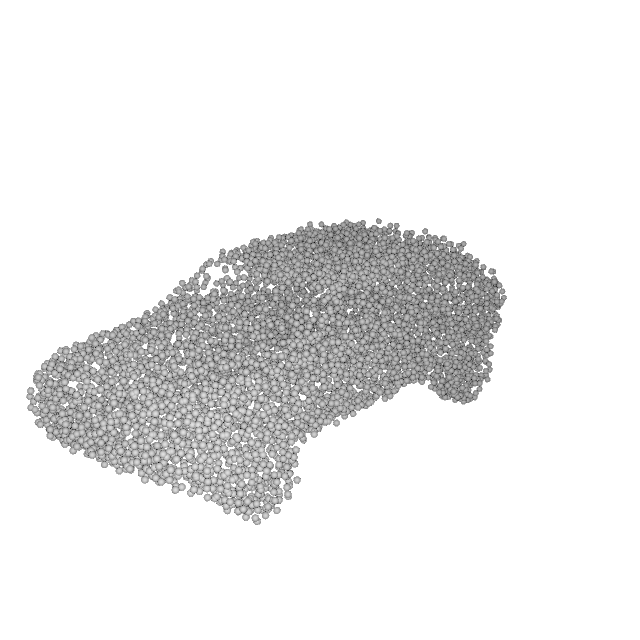}
    \end{subfigure}
    \begin{subfigure}[t]{0.15\textwidth}
        \includegraphics[width=\textwidth]{figures/shapenet_supp/car2/963a4fdf819cc5ab3174b45571ecff3d_2_ours2.png}
    \end{subfigure}

    \vspace{-0.5em}

    \begin{subfigure}[t]{0.15\textwidth}
        \includegraphics[width=\textwidth]{figures/shapenet_supp/car/ea1b46ad39c67e24597505fd7d99b613_2_input.png}
    \end{subfigure}
    \begin{subfigure}[t]{0.15\textwidth}
        \includegraphics[width=\textwidth]{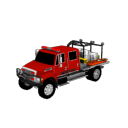}
    \end{subfigure}
    \begin{subfigure}[t]{0.15\textwidth}
        \includegraphics[width=\textwidth]{figures/shapenet_supp/car/ea1b46ad39c67e24597505fd7d99b613_2_gt.png}
    \end{subfigure}
    \begin{subfigure}[t]{0.15\textwidth}
        \includegraphics[width=\textwidth]{figures/shapenet_supp/car/ea1b46ad39c67e24597505fd7d99b613_2_dpc.png}
    \end{subfigure}
    \begin{subfigure}[t]{0.15\textwidth}
        \includegraphics[width=\textwidth]{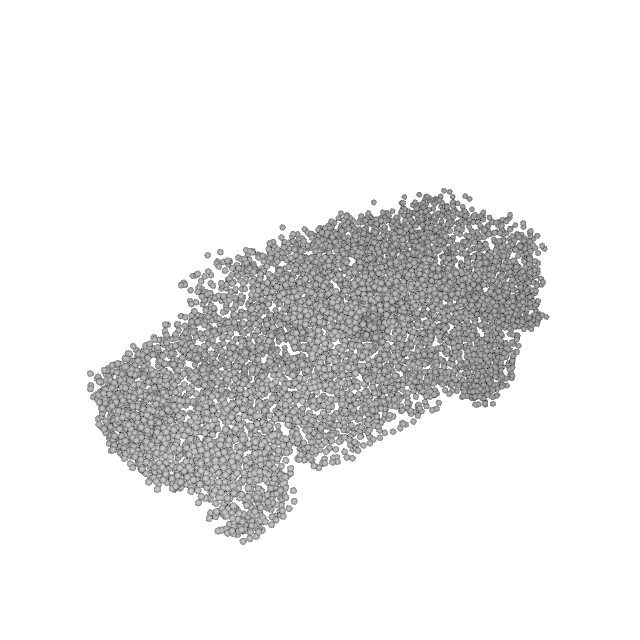}
    \end{subfigure}
    \begin{subfigure}[t]{0.15\textwidth}
        \includegraphics[width=\textwidth]{figures/shapenet_supp/car2/ea1b46ad39c67e24597505fd7d99b613_2_ours2.png}
    \end{subfigure}

    \vspace{-0.5em}

    \begin{subfigure}[t]{0.15\textwidth}
        \includegraphics[width=\textwidth]{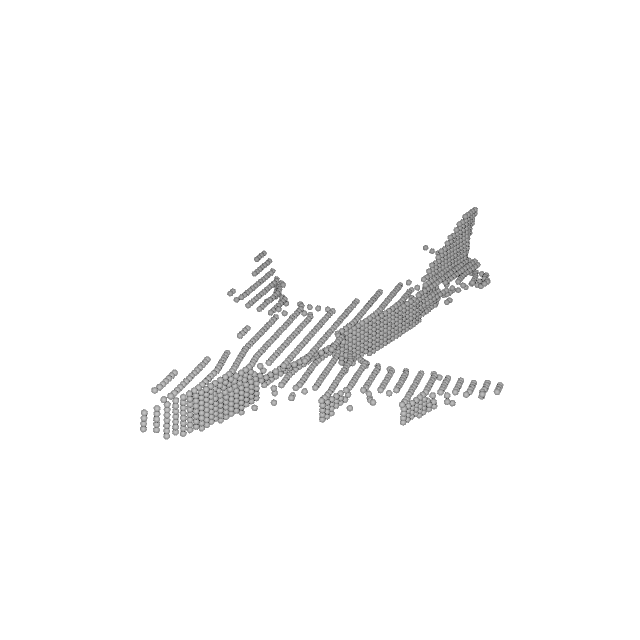}
    \end{subfigure}
    \begin{subfigure}[t]{0.15\textwidth}
        \includegraphics[width=\textwidth]{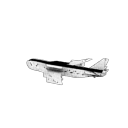}
    \end{subfigure}
    \begin{subfigure}[t]{0.15\textwidth}
        \includegraphics[width=\textwidth]{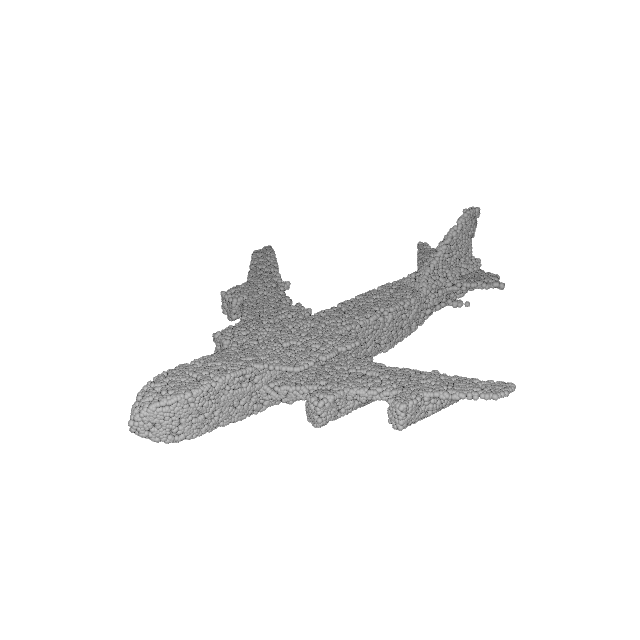}
    \end{subfigure}
    \begin{subfigure}[t]{0.15\textwidth}
        \includegraphics[width=\textwidth]{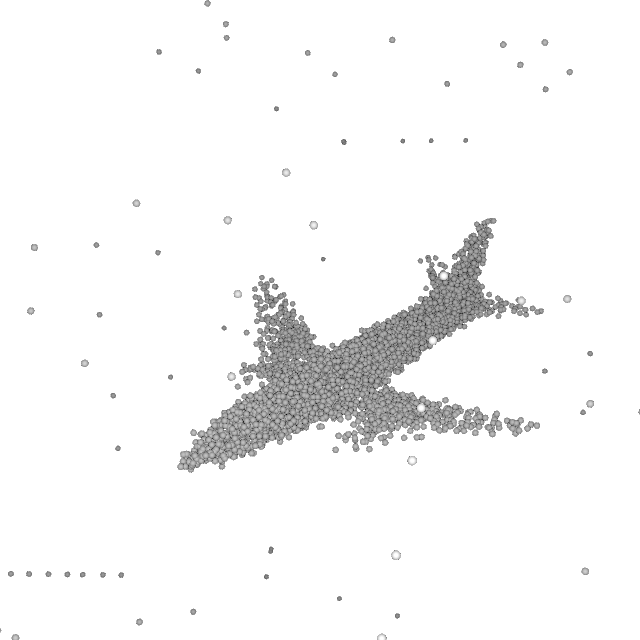}
    \end{subfigure}
    \begin{subfigure}[t]{0.15\textwidth}
        \includegraphics[width=\textwidth]{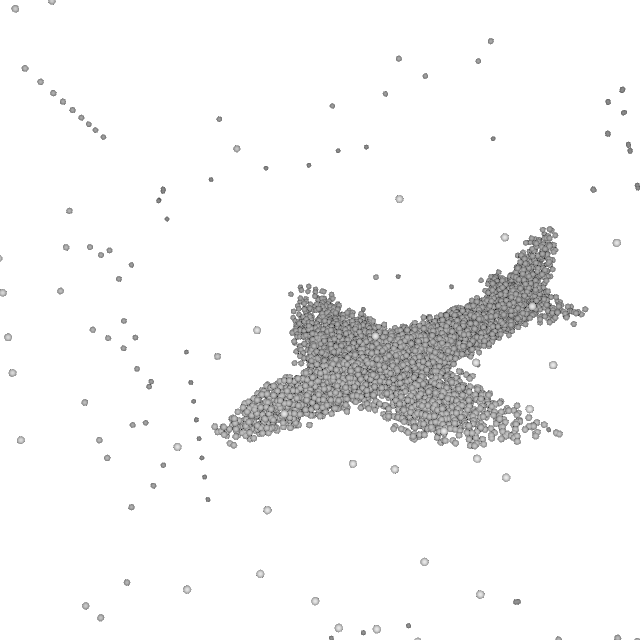}
    \end{subfigure}
    \begin{subfigure}[t]{0.15\textwidth}
        \includegraphics[width=\textwidth]{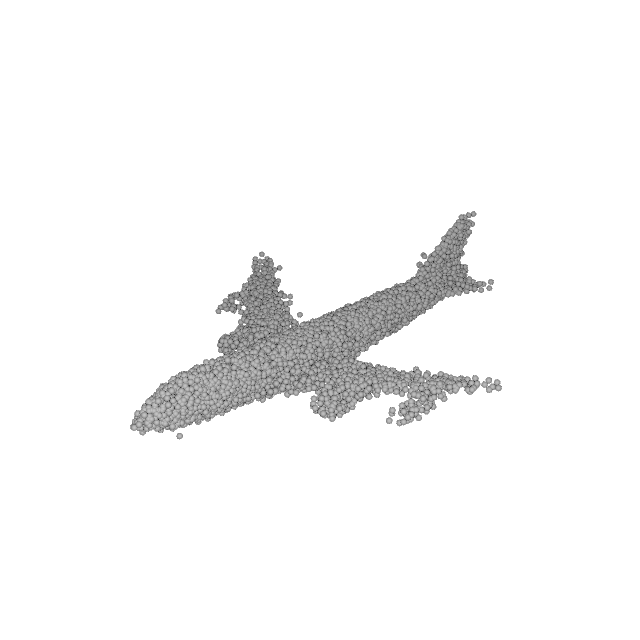}
    \end{subfigure}

    \vspace{-0.5em}

    \begin{subfigure}[t]{0.15\textwidth}
        \includegraphics[width=\textwidth]{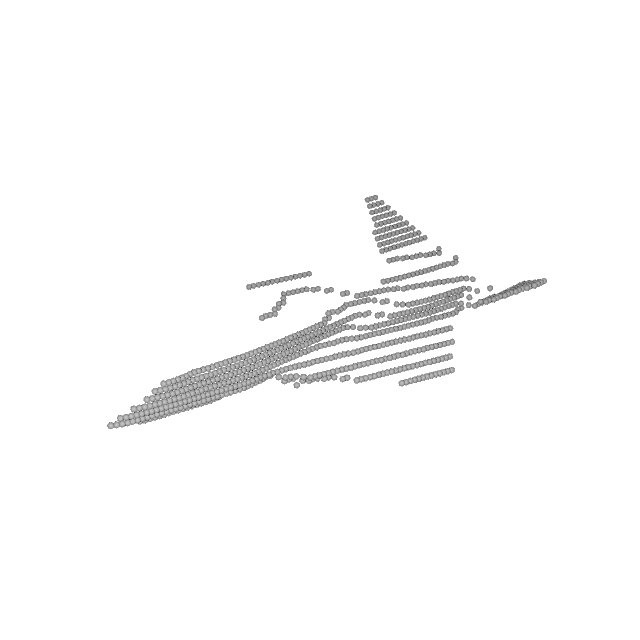}
    \end{subfigure}
    \begin{subfigure}[t]{0.15\textwidth}
        \includegraphics[width=\textwidth]{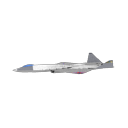}
    \end{subfigure}
    \begin{subfigure}[t]{0.15\textwidth}
        \includegraphics[width=\textwidth]{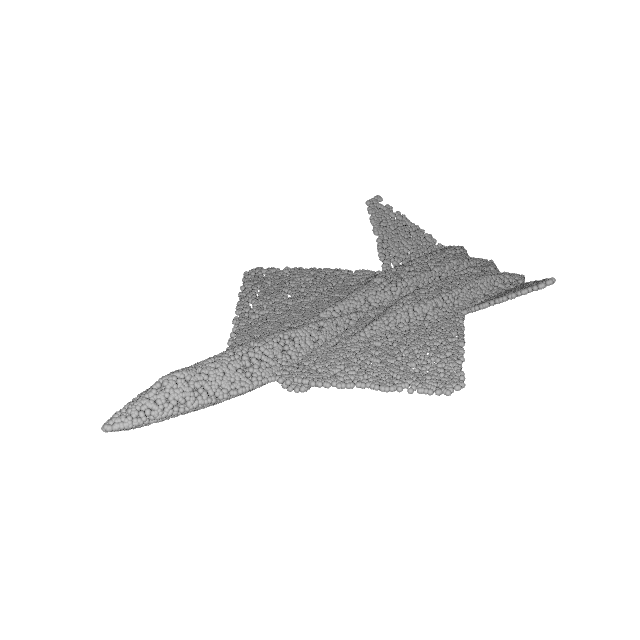}
    \end{subfigure}
    \begin{subfigure}[t]{0.15\textwidth}
        \includegraphics[width=\textwidth]{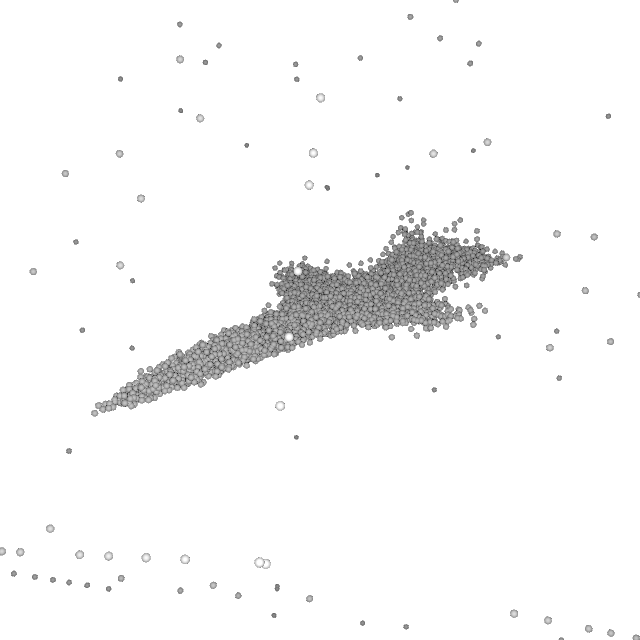}
    \end{subfigure}
    \begin{subfigure}[t]{0.15\textwidth}
        \includegraphics[width=\textwidth]{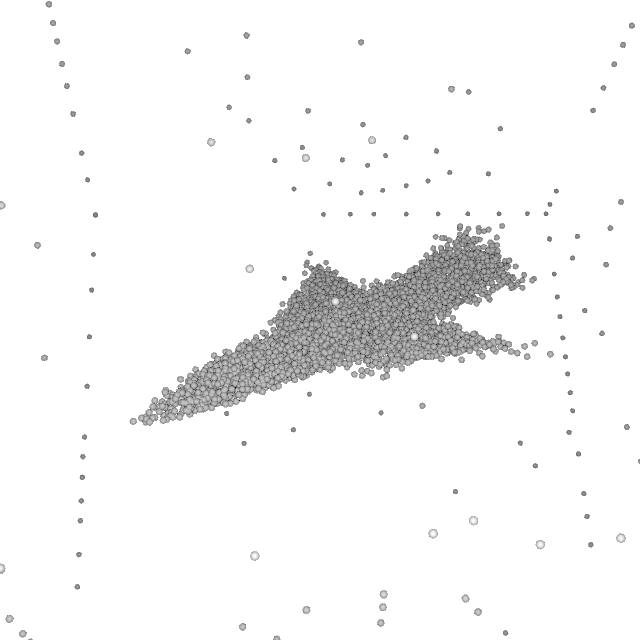}
    \end{subfigure}
    \begin{subfigure}[t]{0.15\textwidth}
        \includegraphics[width=\textwidth]{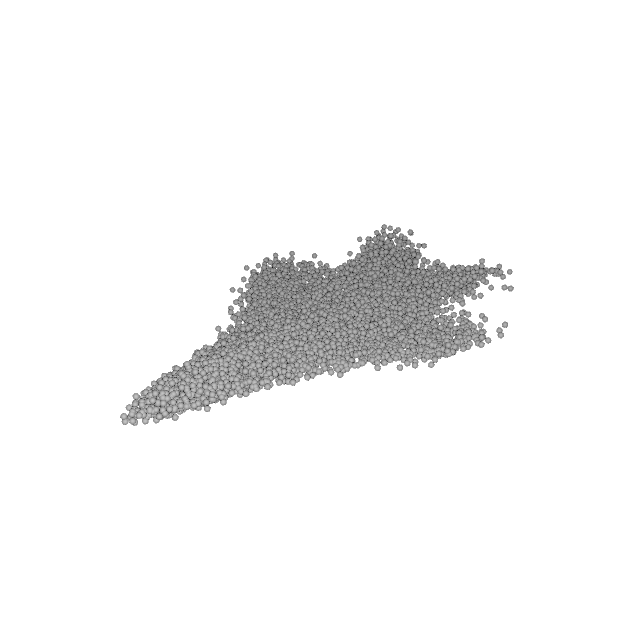}
    \end{subfigure}

    \vspace{-0.5em}

    \begin{subfigure}[t]{0.15\textwidth}
        \includegraphics[width=\textwidth]{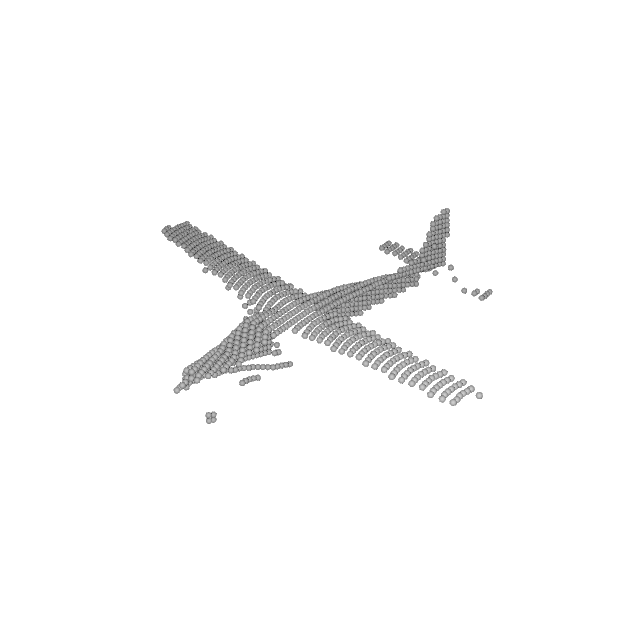}
    \end{subfigure}
    \begin{subfigure}[t]{0.15\textwidth}
        \includegraphics[width=\textwidth]{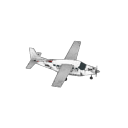}
    \end{subfigure}
    \begin{subfigure}[t]{0.15\textwidth}
        \includegraphics[width=\textwidth]{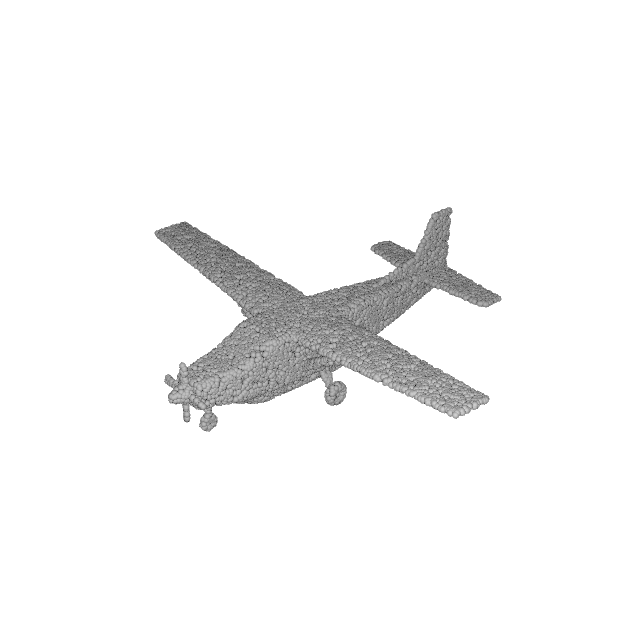}
    \end{subfigure}
    \begin{subfigure}[t]{0.15\textwidth}
        \includegraphics[width=\textwidth]{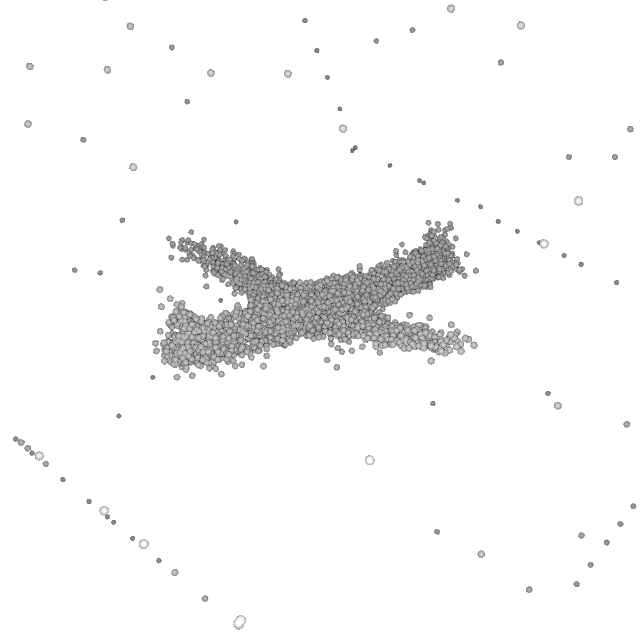}
    \end{subfigure}
    \begin{subfigure}[t]{0.15\textwidth}
        \includegraphics[width=\textwidth]{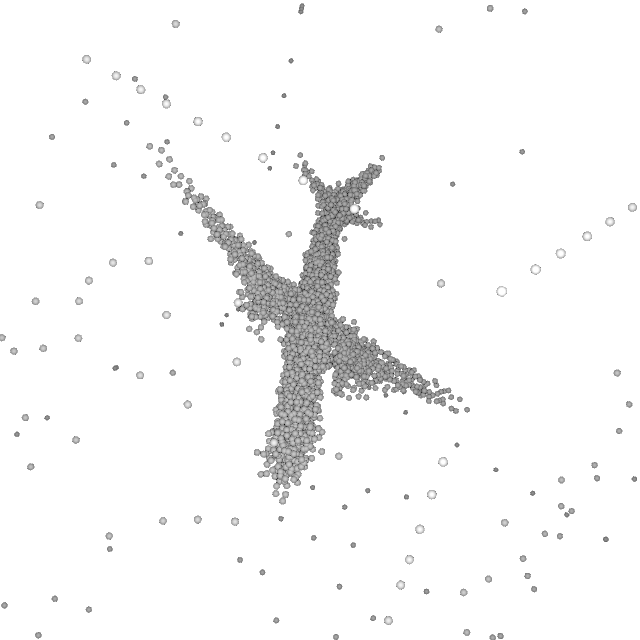}
    \end{subfigure}
    \begin{subfigure}[t]{0.15\textwidth}
        \includegraphics[width=\textwidth]{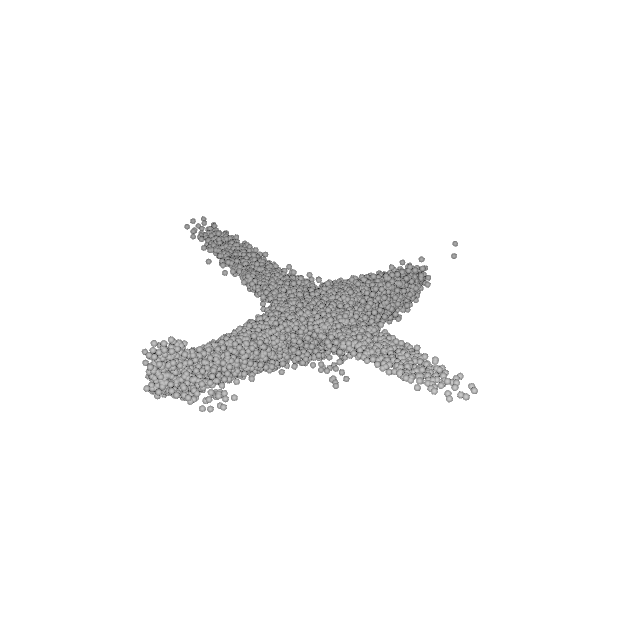}
    \end{subfigure}

    \vspace{-0.5em}

    \begin{subfigure}[t]{0.15\textwidth}
        \includegraphics[width=\textwidth]{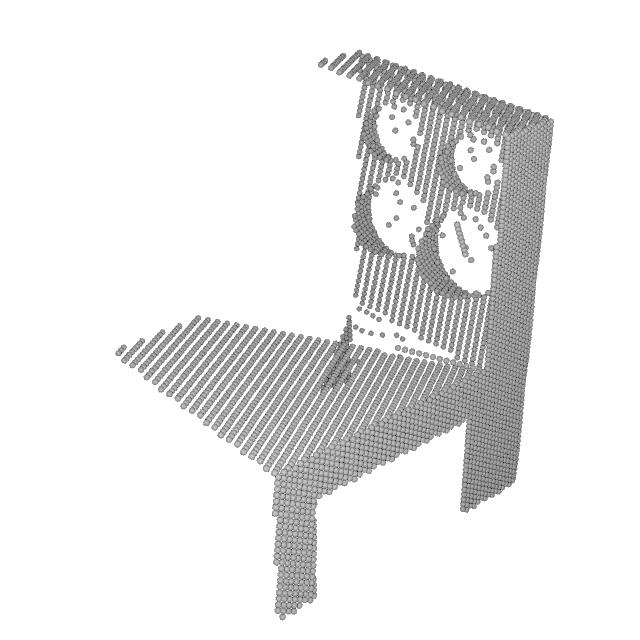}
    \end{subfigure}
    \begin{subfigure}[t]{0.15\textwidth}
        \includegraphics[width=\textwidth]{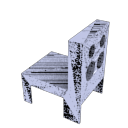}
    \end{subfigure}
    \begin{subfigure}[t]{0.15\textwidth}
        \includegraphics[width=\textwidth]{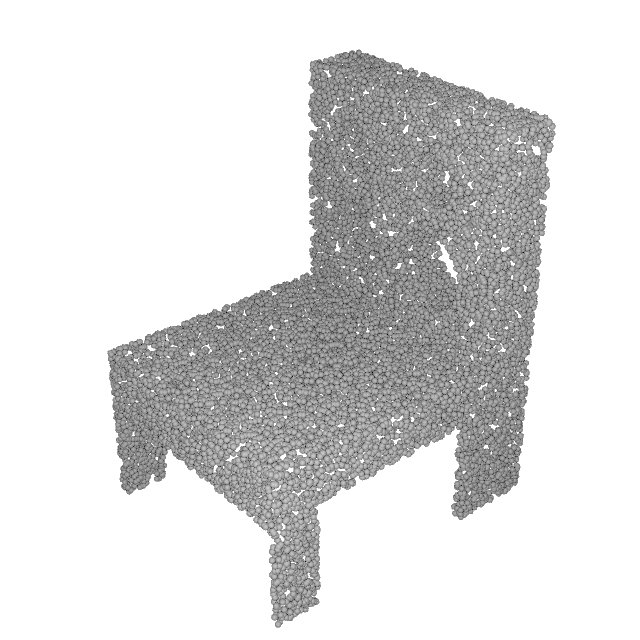}
    \end{subfigure}
    \begin{subfigure}[t]{0.15\textwidth}
        \includegraphics[width=\textwidth]{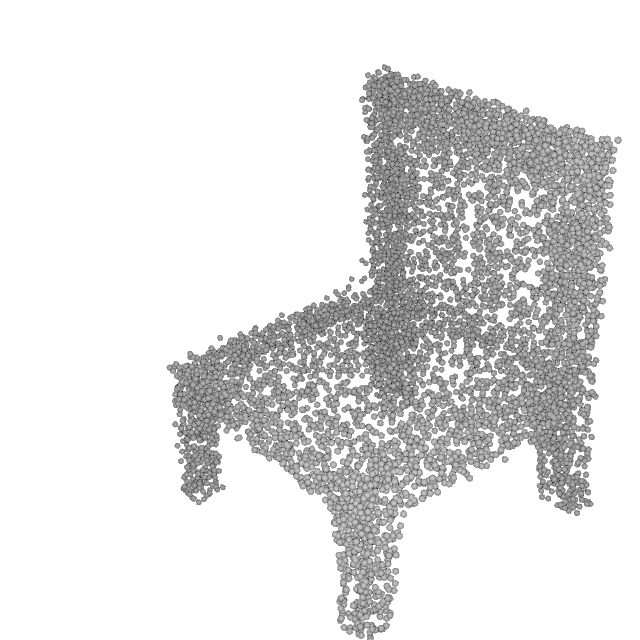}
    \end{subfigure}
    \begin{subfigure}[t]{0.15\textwidth}
        \includegraphics[width=\textwidth]{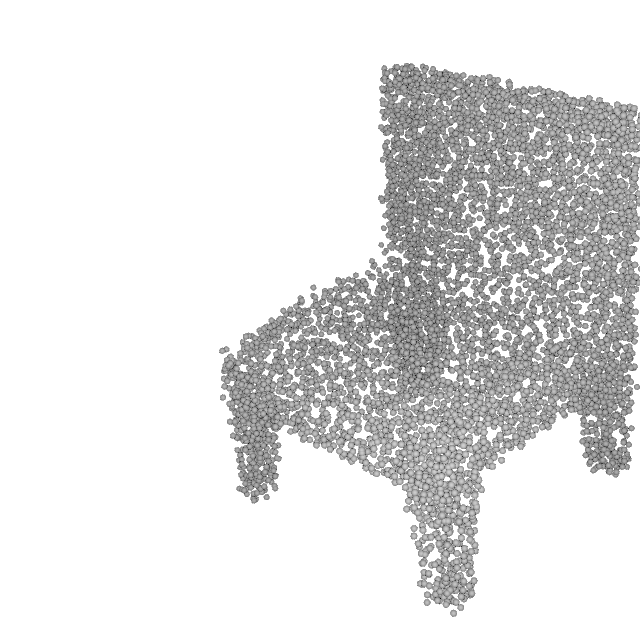}
    \end{subfigure}
    \begin{subfigure}[t]{0.15\textwidth}
        \includegraphics[width=\textwidth]{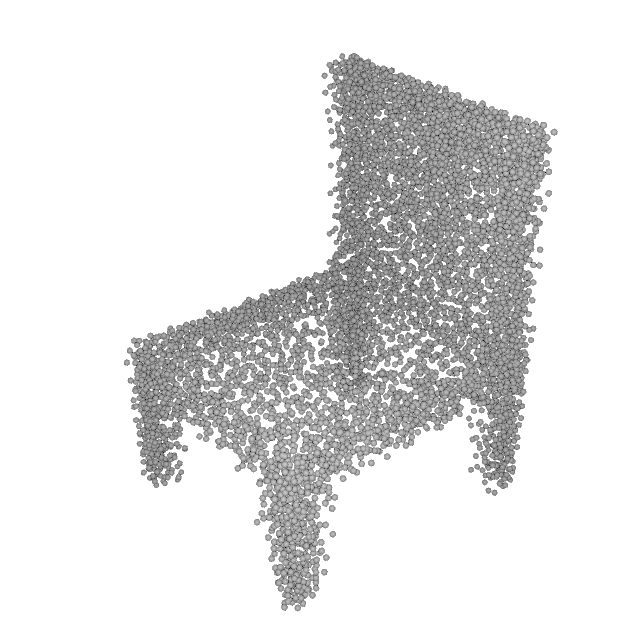}
    \end{subfigure}

    \begin{subfigure}[t]{0.15\textwidth}
        \includegraphics[width=\textwidth]{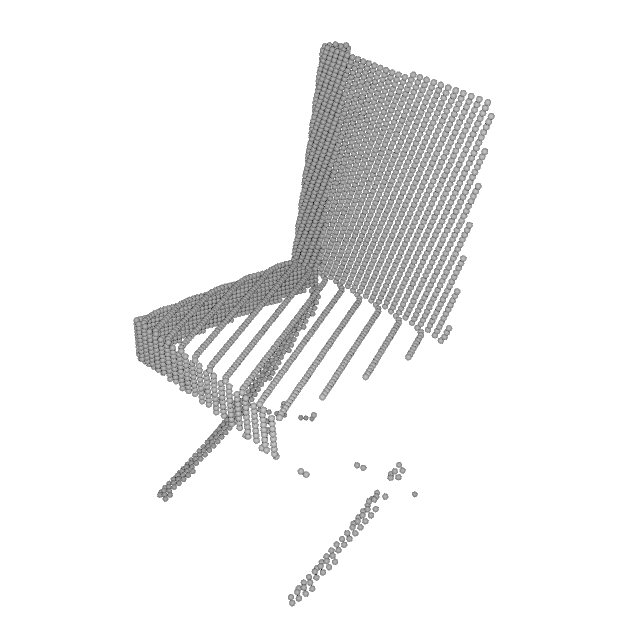}
    \end{subfigure}
    \begin{subfigure}[t]{0.15\textwidth}
        \includegraphics[width=\textwidth]{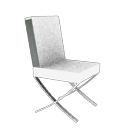}
    \end{subfigure}
    \begin{subfigure}[t]{0.15\textwidth}
        \includegraphics[width=\textwidth]{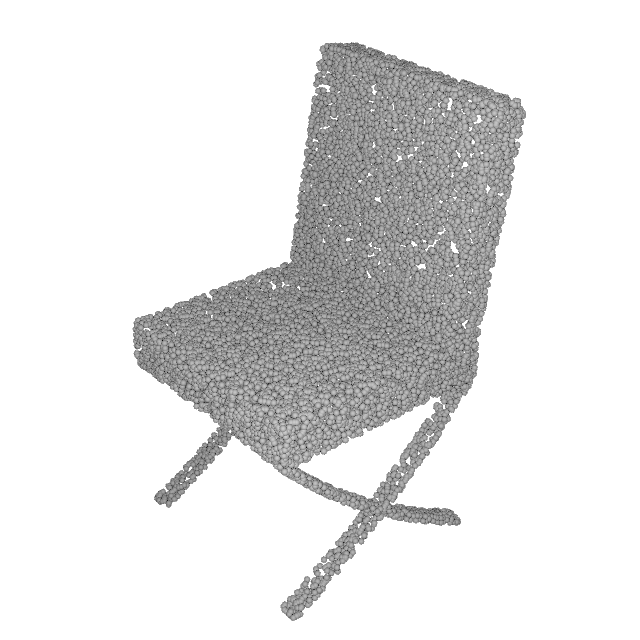}
    \end{subfigure}
    \begin{subfigure}[t]{0.15\textwidth}
        \includegraphics[width=\textwidth]{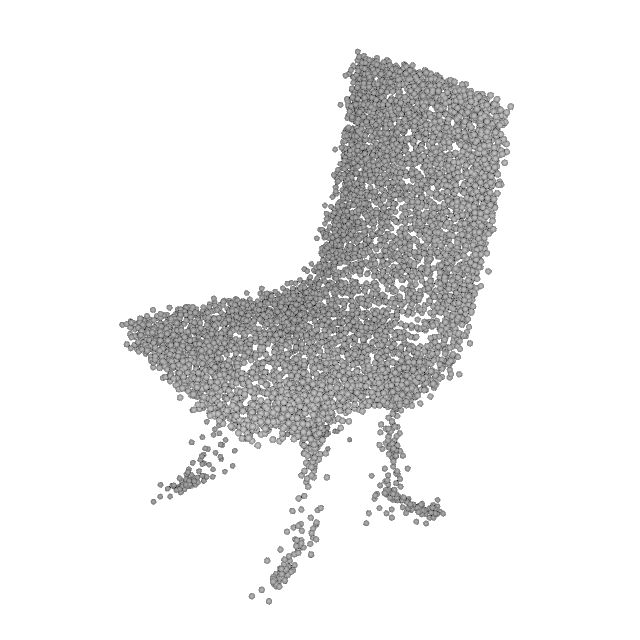}
    \end{subfigure}
    \begin{subfigure}[t]{0.15\textwidth}
        \includegraphics[width=\textwidth]{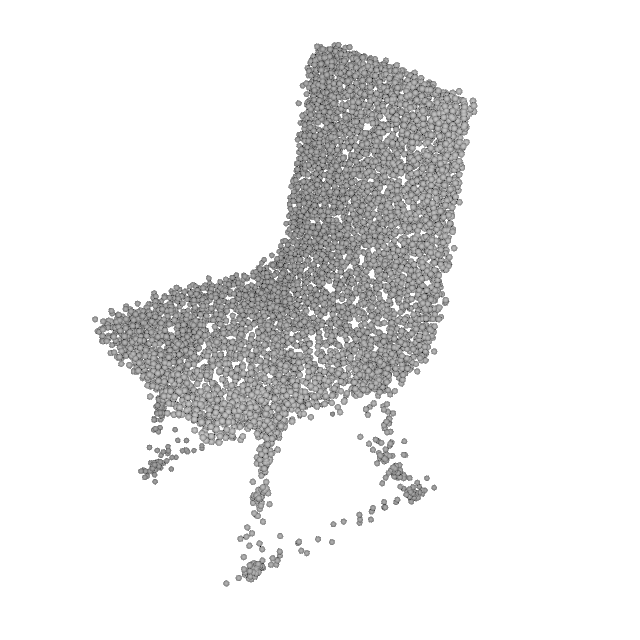}
    \end{subfigure}
    \begin{subfigure}[t]{0.15\textwidth}
        \includegraphics[width=\textwidth]{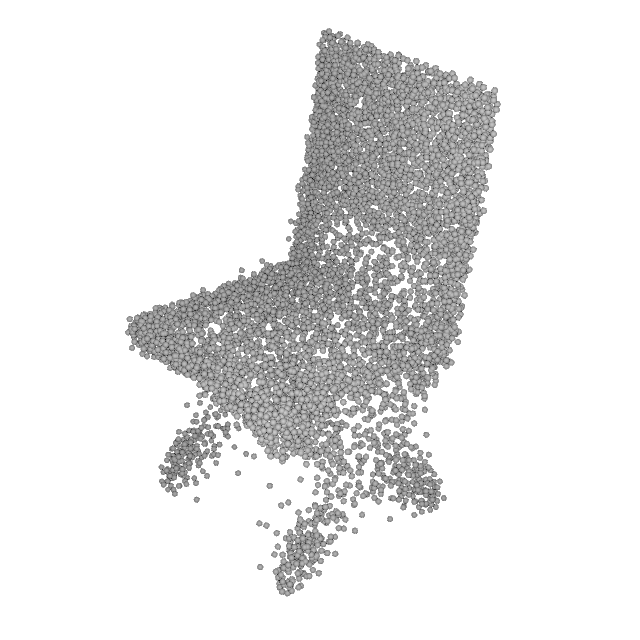}
    \end{subfigure}

    \vspace{-0.5em}

    \begin{subfigure}[t]{0.15\textwidth}
        \includegraphics[width=\textwidth]{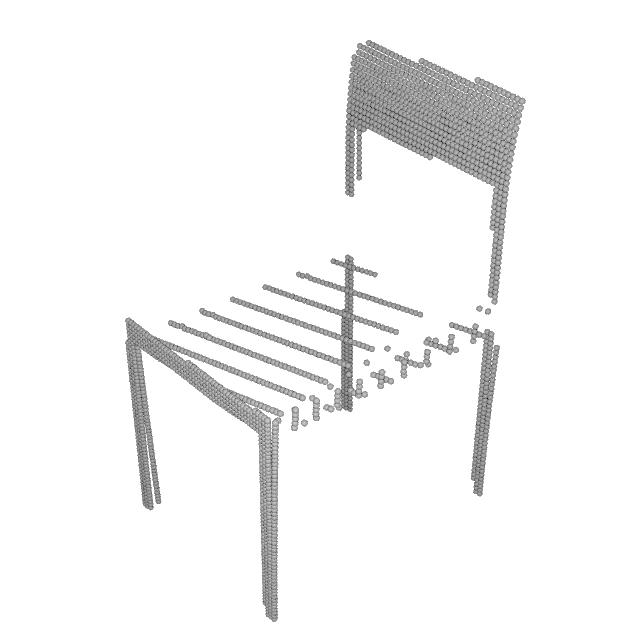}
    \end{subfigure}
    \begin{subfigure}[t]{0.15\textwidth}
        \includegraphics[width=\textwidth]{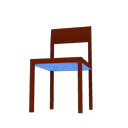}
    \end{subfigure}
    \begin{subfigure}[t]{0.15\textwidth}
        \includegraphics[width=\textwidth]{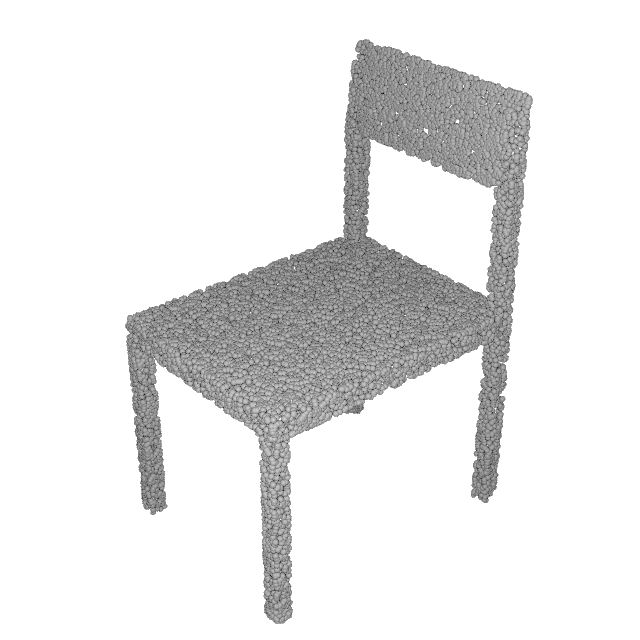}
    \end{subfigure}
    \begin{subfigure}[t]{0.15\textwidth}
        \includegraphics[width=\textwidth]{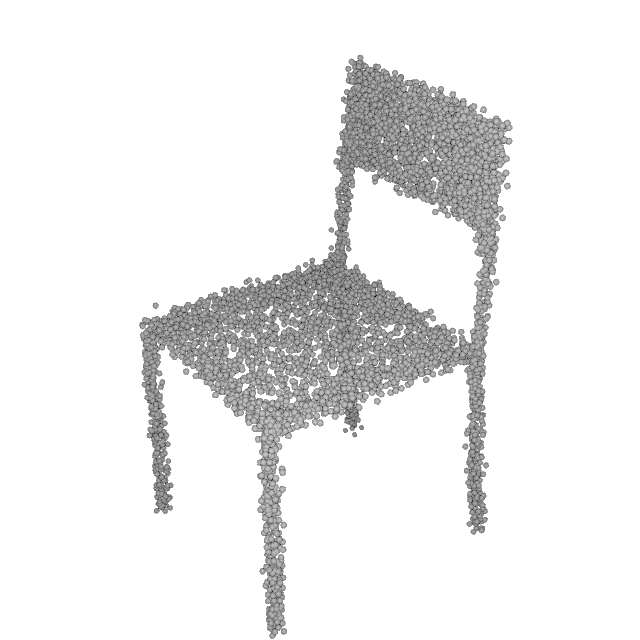}
    \end{subfigure}
    \begin{subfigure}[t]{0.15\textwidth}
        \includegraphics[width=\textwidth]{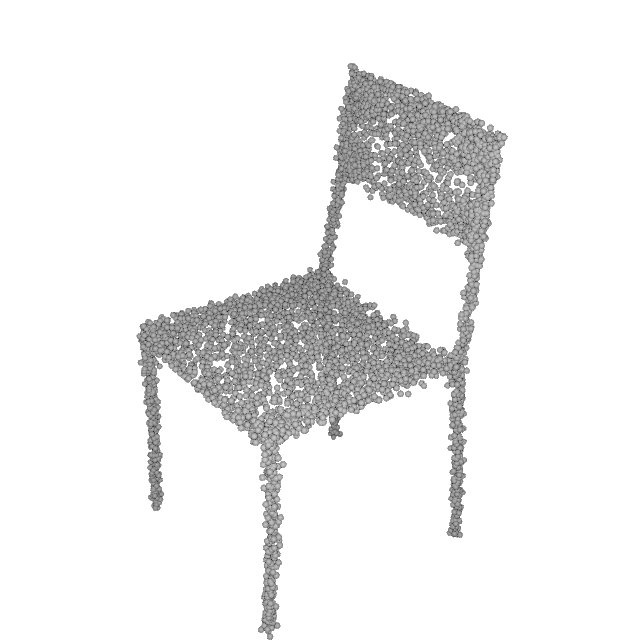}
    \end{subfigure}
    \begin{subfigure}[t]{0.15\textwidth}
        \includegraphics[width=\textwidth]{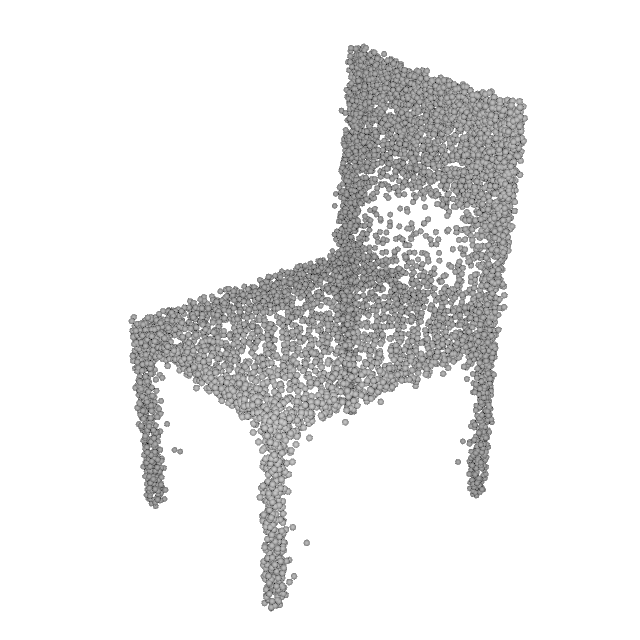}
    \end{subfigure}

    \caption{Qualitative results of 3D shape completion on the test set of ShapeNet.
        All the point clouds are transformed to the ground-truth canonical frame and visualized at a fixed viewpoint.
        For cars, we use the variant of our method.
    }
    \label{fig:shapenet-supp}
\end{figure}

To better understand how our method performs compared to baselines, we visualize more results in this section.
Fig~\ref{fig:shapenet-supp} demonstrates more qualitative results on ShapeNet.
It can be observed that shapes and poses estimated by our method are more accurate than DPC and DPC$^\dagger$, especially for chairs and planes.
Since planes are usually flat, DPC and its variant suffer from sparse 2D observations and generate many artifacts.

\begin{figure}[h]
    \centering
    \begin{subfigure}[t]{0.16\textwidth}
        \caption*{Input}
        \vspace{-0.5em}
        \includegraphics[width=\textwidth]{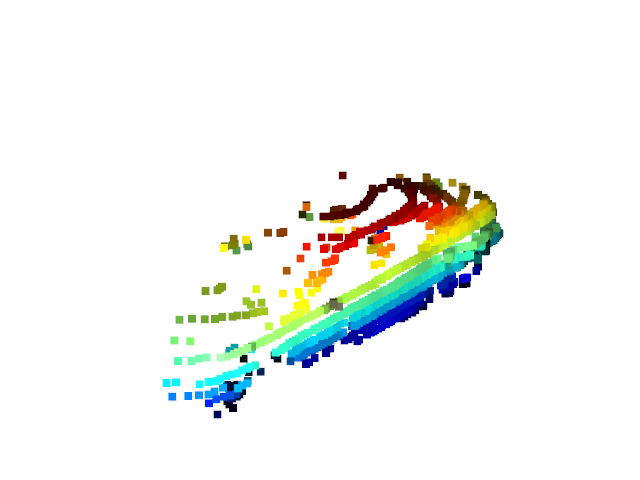}
    \end{subfigure}
    \begin{subfigure}[t]{0.16\textwidth}
        \caption*{Ground truth}
        \vspace{-0.5em}
        \includegraphics[width=\textwidth]{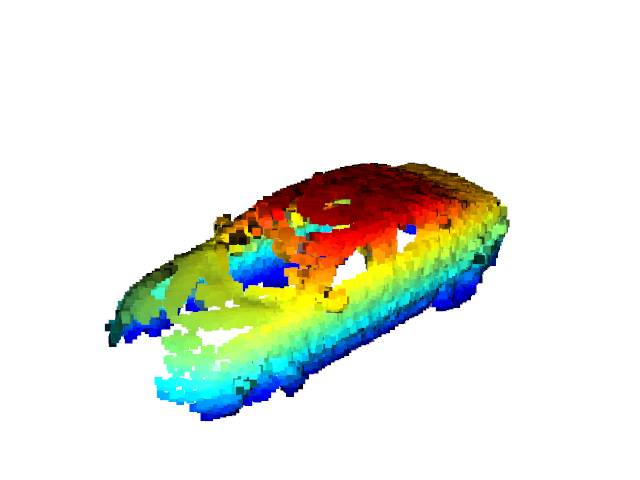}
    \end{subfigure}
    \begin{subfigure}[t]{0.16\textwidth}
        \caption*{Local-ICP}
        \vspace{-0.5em}
        \includegraphics[width=\textwidth]{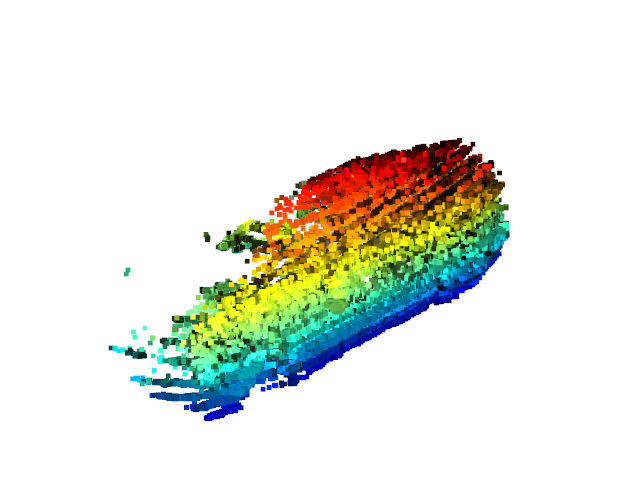}
    \end{subfigure}
    \begin{subfigure}[t]{0.16\textwidth}
        \caption*{Global-ICP}
        \vspace{-0.5em}
        \includegraphics[width=\textwidth]{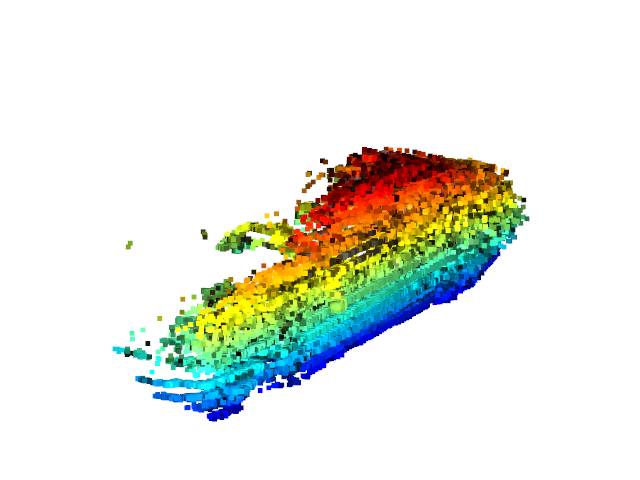}
    \end{subfigure}
    \begin{subfigure}[t]{0.16\textwidth}
        \caption*{Ours(shape)}
        \vspace{-0.5em}
        \includegraphics[width=\textwidth]{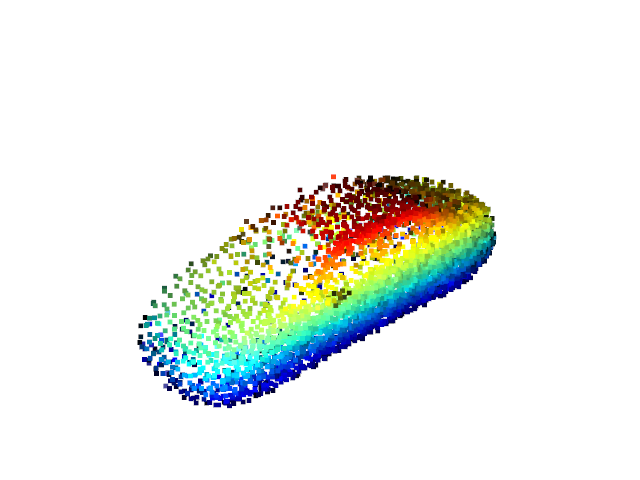}
    \end{subfigure}
    \begin{subfigure}[t]{0.16\textwidth}
        \caption*{Ours(registration)}
        \vspace{-0.5em}
        \includegraphics[width=\textwidth]{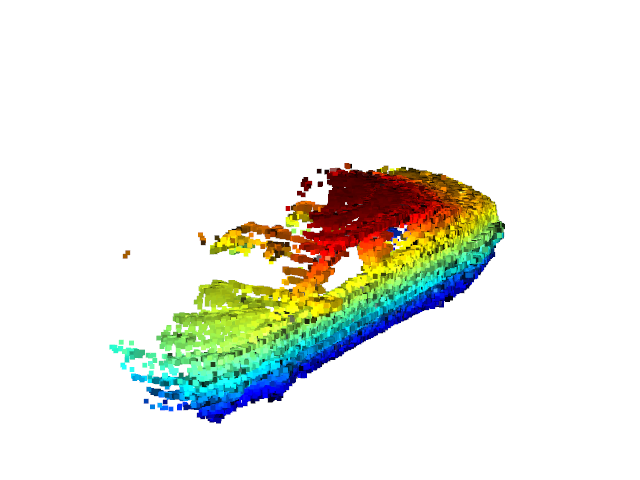}
    \end{subfigure}

    \begin{subfigure}[t]{0.16\textwidth}
        \includegraphics[width=\textwidth]{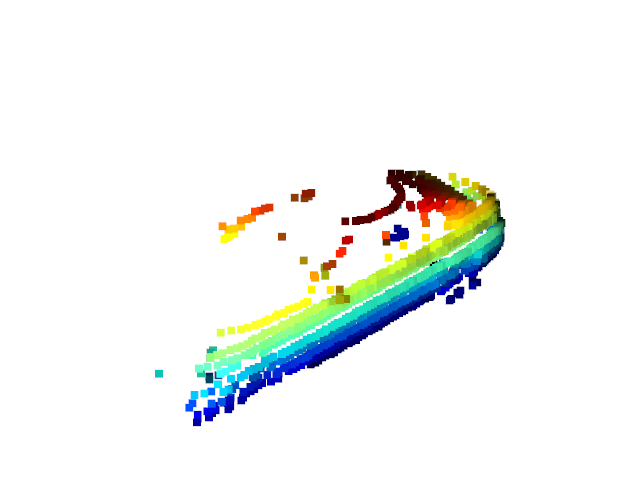}
    \end{subfigure}
    \begin{subfigure}[t]{0.16\textwidth}
        \includegraphics[width=\textwidth]{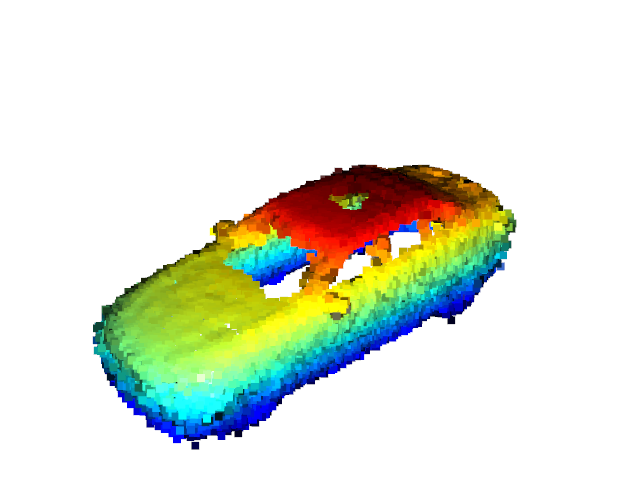}
    \end{subfigure}
    \begin{subfigure}[t]{0.16\textwidth}
        \includegraphics[width=\textwidth]{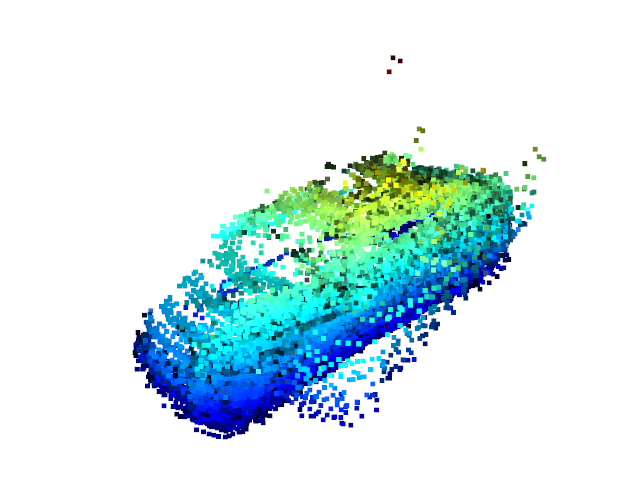}
    \end{subfigure}
    \begin{subfigure}[t]{0.16\textwidth}
        \includegraphics[width=\textwidth]{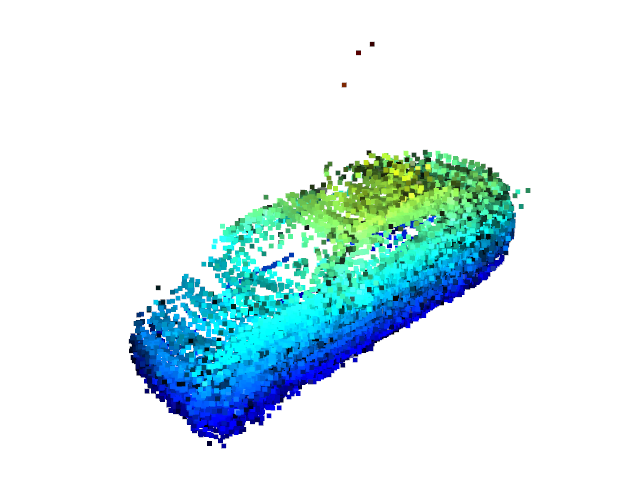}
    \end{subfigure}
    \begin{subfigure}[t]{0.16\textwidth}
        \includegraphics[width=\textwidth]{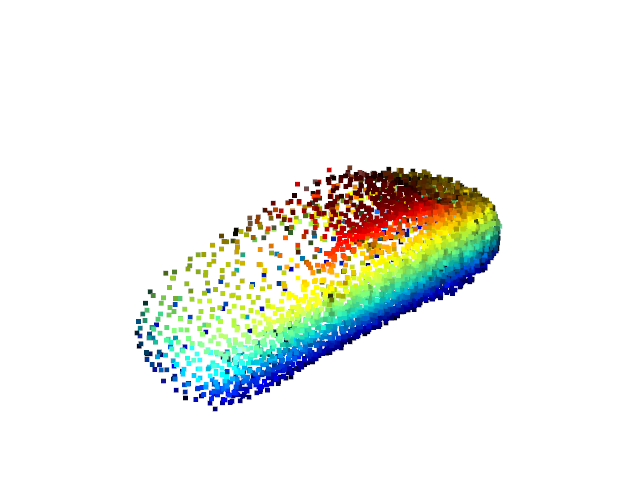}
    \end{subfigure}
    \begin{subfigure}[t]{0.16\textwidth}
        \includegraphics[width=\textwidth]{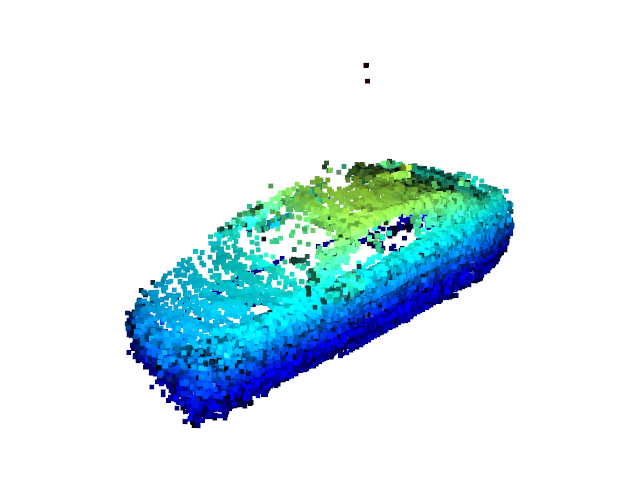}
    \end{subfigure}

    \begin{subfigure}[t]{0.16\textwidth}
        \includegraphics[width=\textwidth]{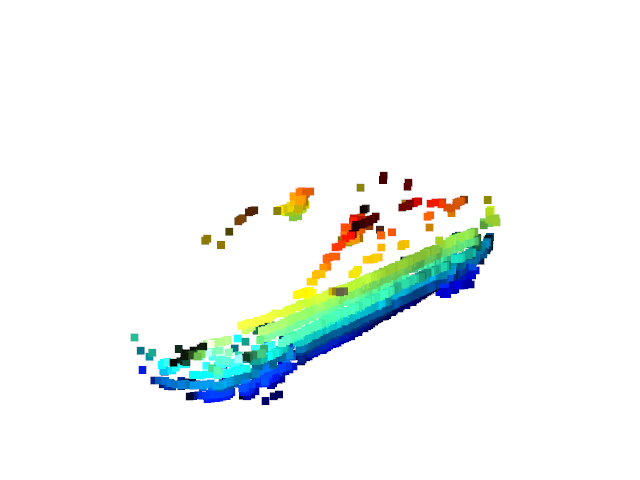}
    \end{subfigure}
    \begin{subfigure}[t]{0.16\textwidth}
        \includegraphics[width=\textwidth]{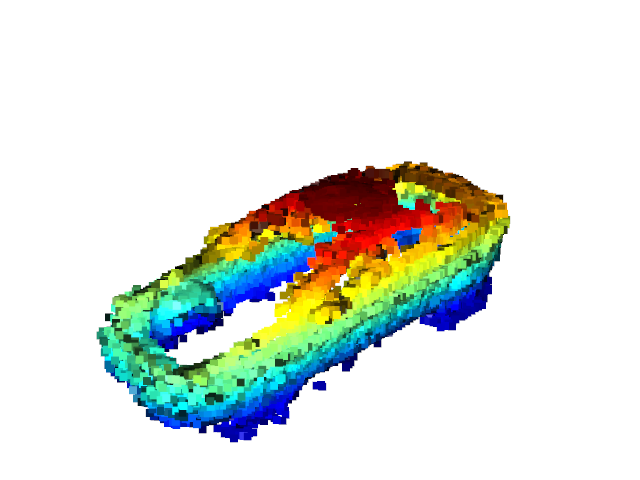}
    \end{subfigure}
    \begin{subfigure}[t]{0.16\textwidth}
        \includegraphics[width=\textwidth]{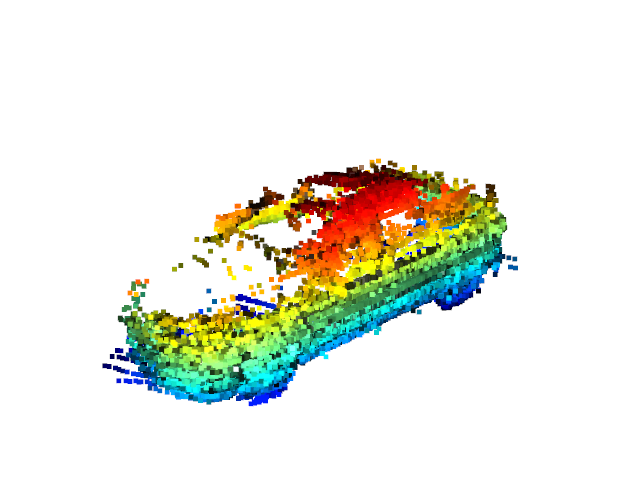}
    \end{subfigure}
    \begin{subfigure}[t]{0.16\textwidth}
        \includegraphics[width=\textwidth]{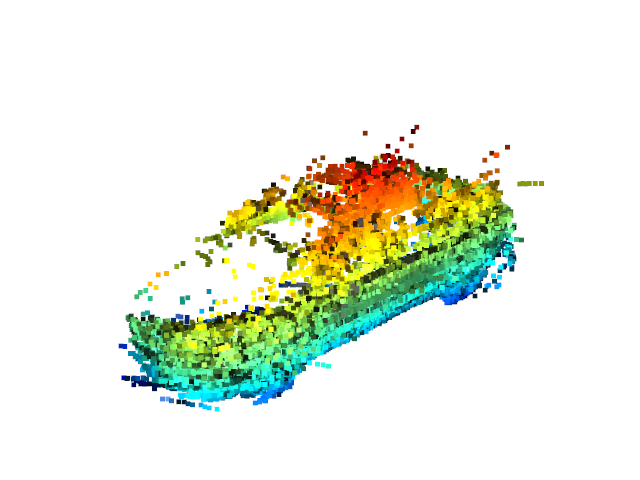}
    \end{subfigure}
    \begin{subfigure}[t]{0.16\textwidth}
        \includegraphics[width=\textwidth]{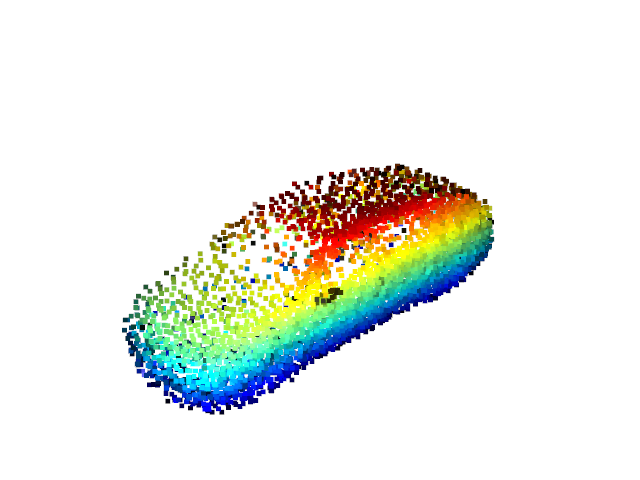}
    \end{subfigure}
    \begin{subfigure}[t]{0.16\textwidth}
        \includegraphics[width=\textwidth]{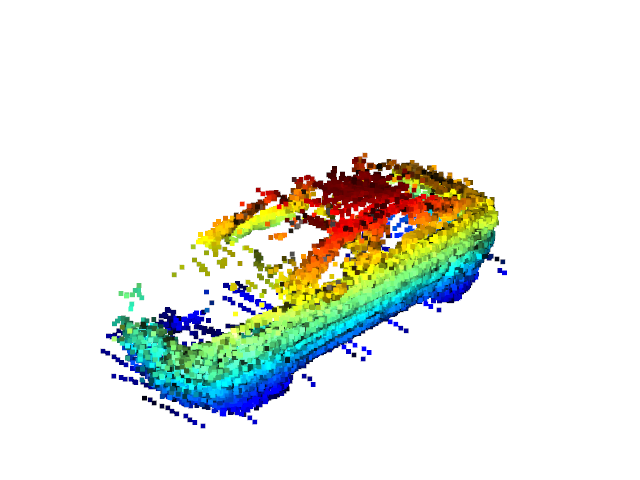}
    \end{subfigure}

    \begin{subfigure}[t]{0.16\textwidth}
        \includegraphics[width=\textwidth]{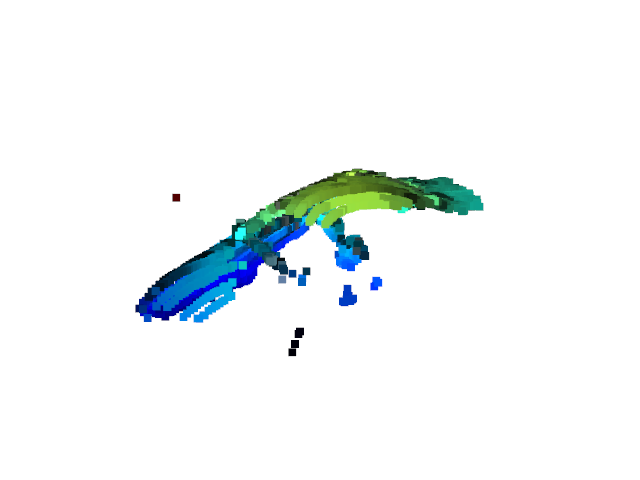}
    \end{subfigure}
    \begin{subfigure}[t]{0.16\textwidth}
        \includegraphics[width=\textwidth]{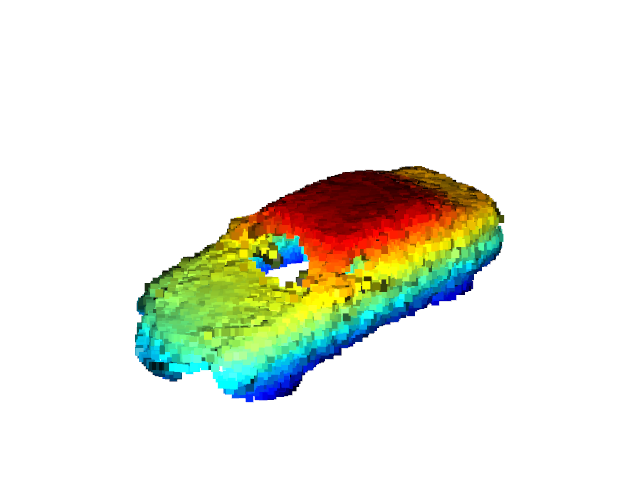}
    \end{subfigure}
    \begin{subfigure}[t]{0.16\textwidth}
        \includegraphics[width=\textwidth]{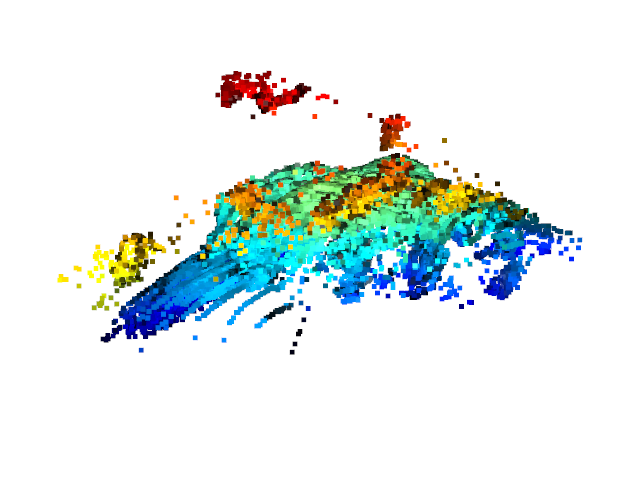}
    \end{subfigure}
    \begin{subfigure}[t]{0.16\textwidth}
        \includegraphics[width=\textwidth]{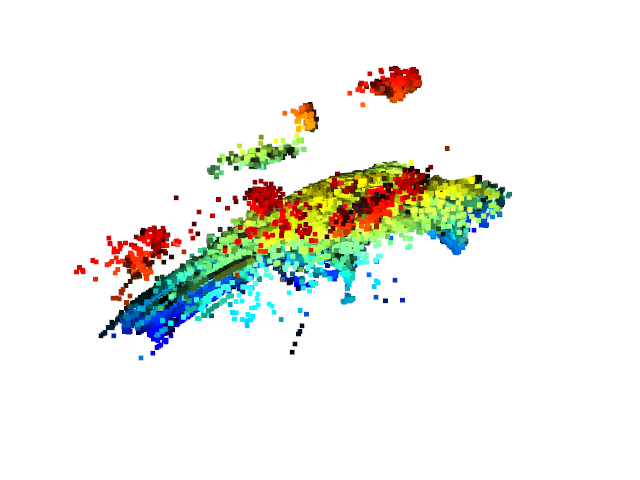}
    \end{subfigure}
    \begin{subfigure}[t]{0.16\textwidth}
        \includegraphics[width=\textwidth]{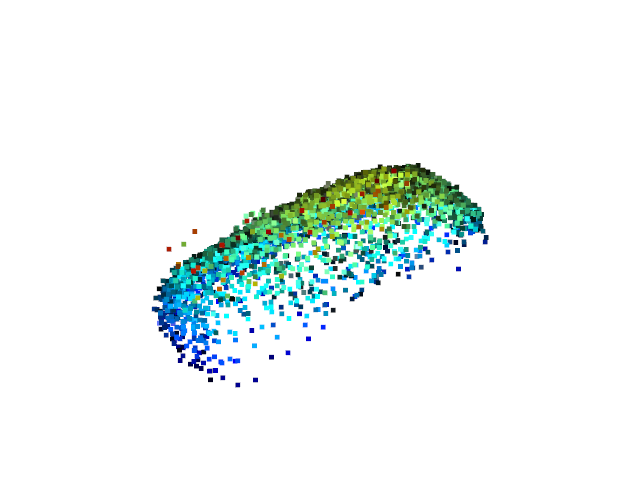}
    \end{subfigure}
    \begin{subfigure}[t]{0.16\textwidth}
        \includegraphics[width=\textwidth]{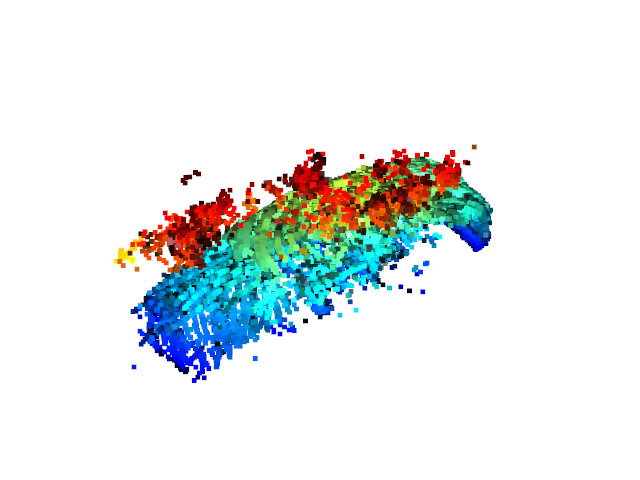}
    \end{subfigure}

    \begin{subfigure}[t]{0.16\textwidth}
        \includegraphics[width=\textwidth]{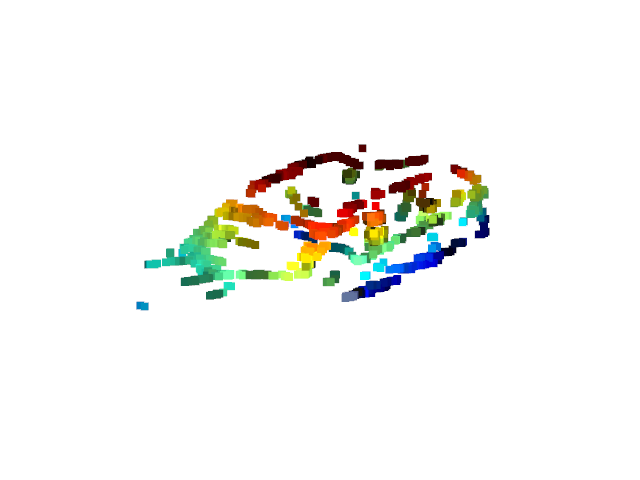}
    \end{subfigure}
    \begin{subfigure}[t]{0.16\textwidth}
        \includegraphics[width=\textwidth]{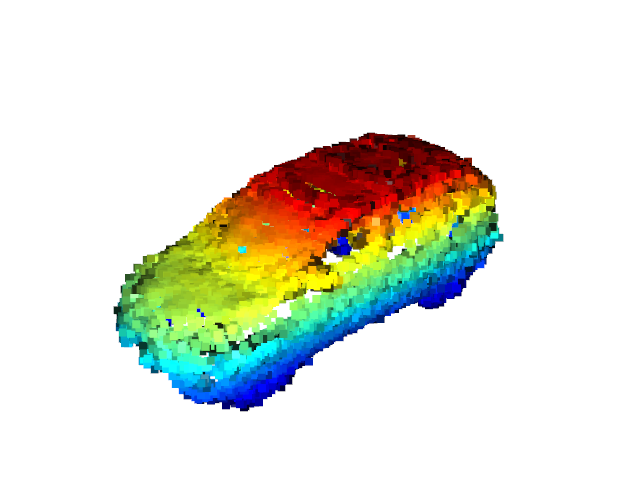}
    \end{subfigure}
    \begin{subfigure}[t]{0.16\textwidth}
        \includegraphics[width=\textwidth]{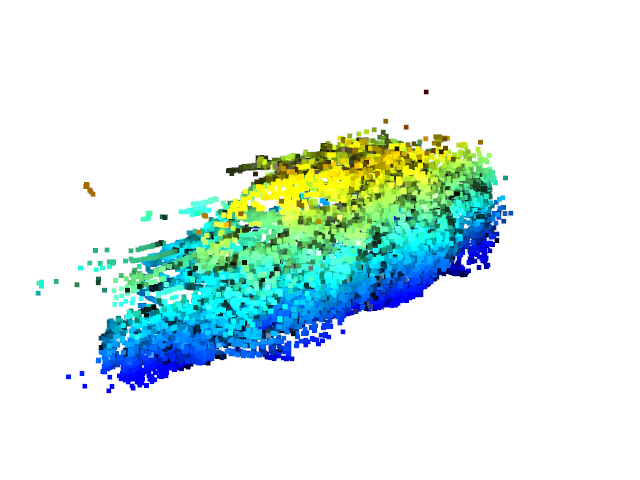}
    \end{subfigure}
    \begin{subfigure}[t]{0.16\textwidth}
        \includegraphics[width=\textwidth]{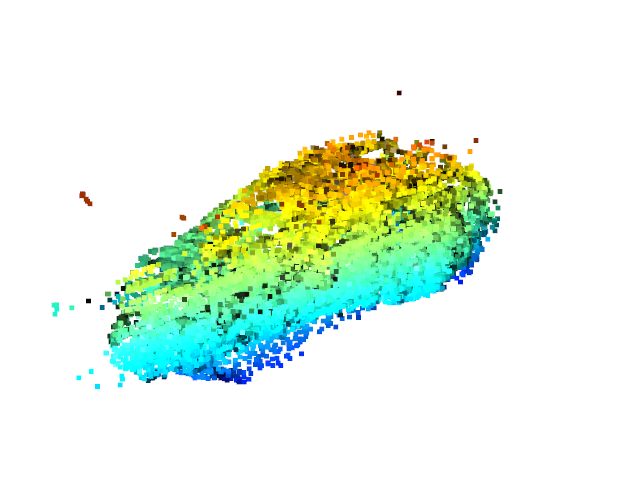}
    \end{subfigure}
    \begin{subfigure}[t]{0.16\textwidth}
        \includegraphics[width=\textwidth]{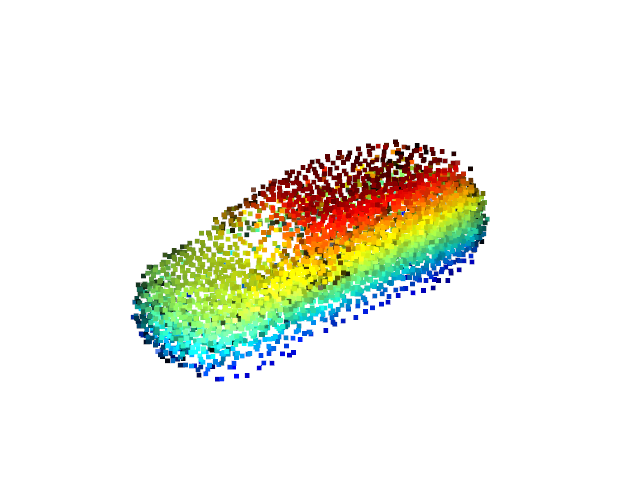}
    \end{subfigure}
    \begin{subfigure}[t]{0.16\textwidth}
        \includegraphics[width=\textwidth]{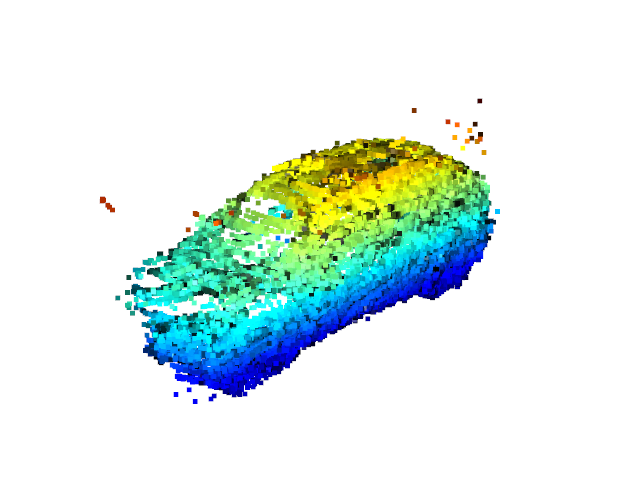}
    \end{subfigure}

    \begin{subfigure}[t]{0.16\textwidth}
        \includegraphics[width=\textwidth]{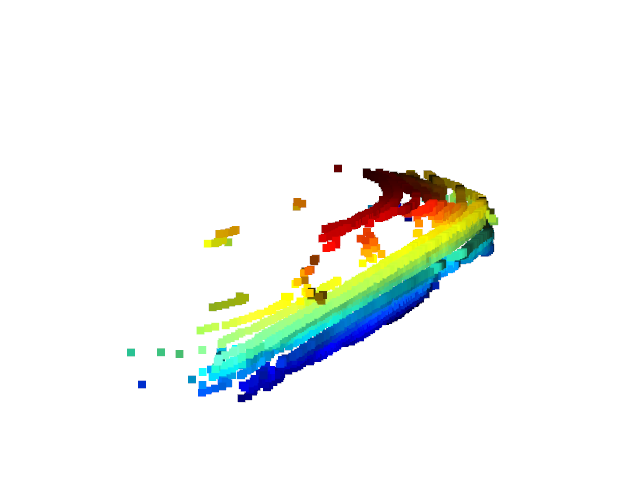}
    \end{subfigure}
    \begin{subfigure}[t]{0.16\textwidth}
        \includegraphics[width=\textwidth]{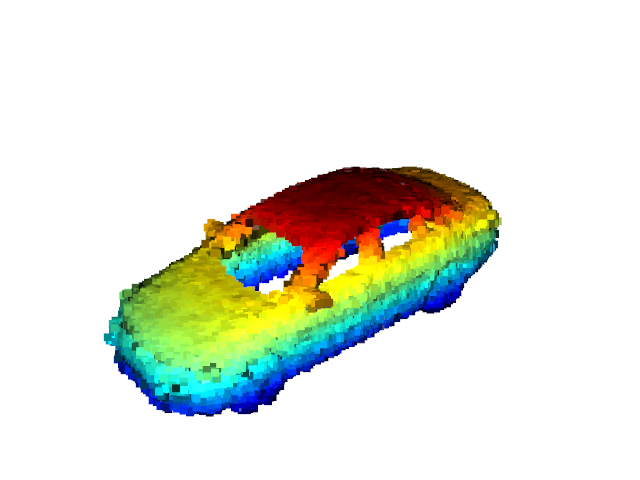}
    \end{subfigure}
    \begin{subfigure}[t]{0.16\textwidth}
        \includegraphics[width=\textwidth]{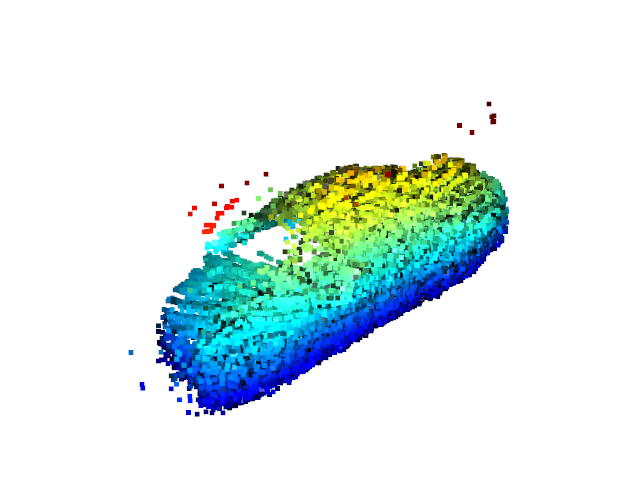}
    \end{subfigure}
    \begin{subfigure}[t]{0.16\textwidth}
        \includegraphics[width=\textwidth]{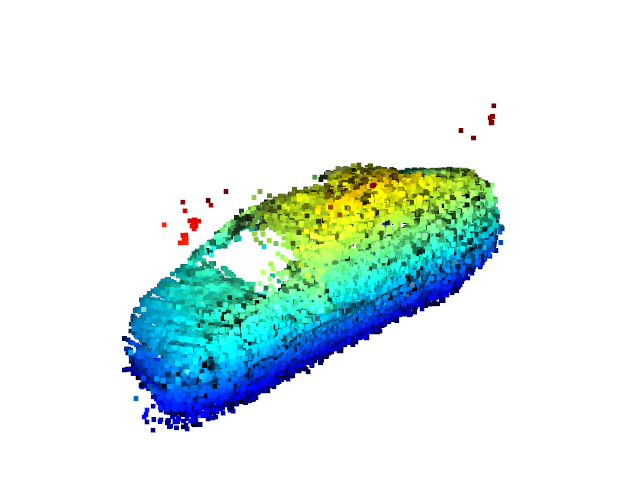}
    \end{subfigure}
    \begin{subfigure}[t]{0.16\textwidth}
        \includegraphics[width=\textwidth]{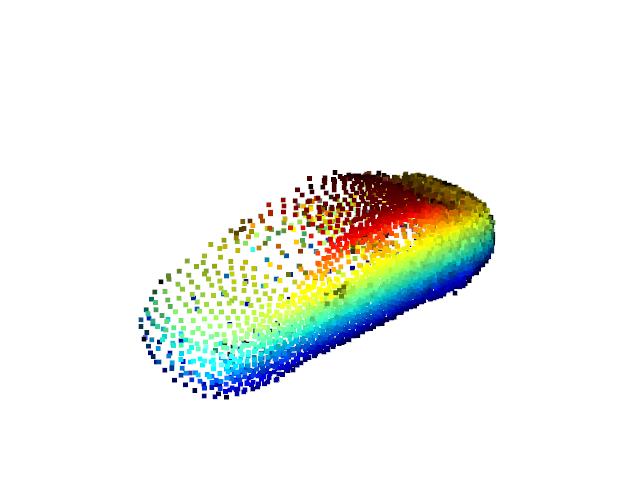}
    \end{subfigure}
    \begin{subfigure}[t]{0.16\textwidth}
        \includegraphics[width=\textwidth]{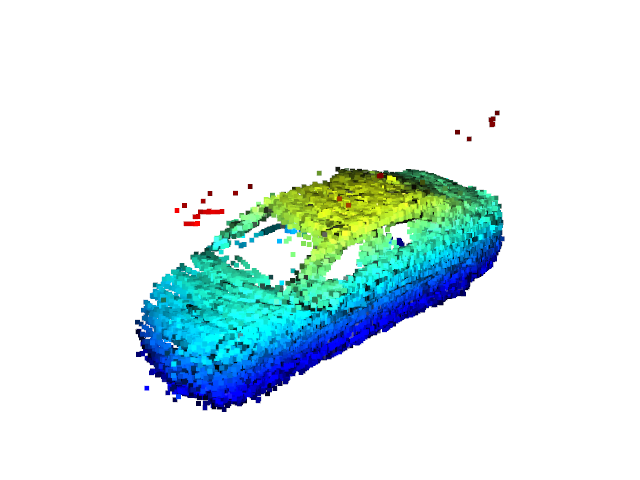}
    \end{subfigure}

    \begin{subfigure}[t]{0.16\textwidth}
        \includegraphics[width=\textwidth]{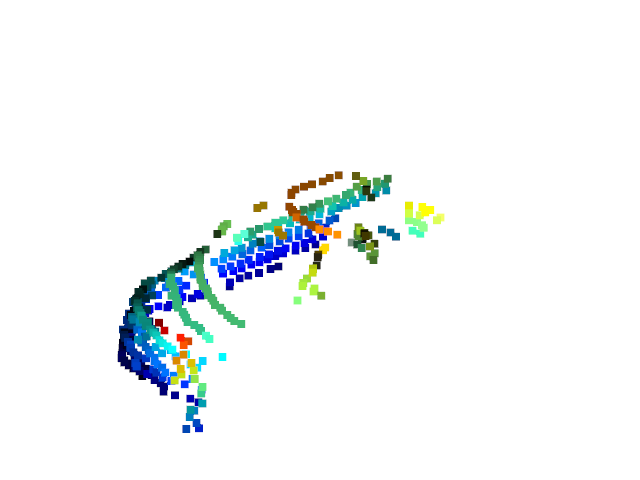}
    \end{subfigure}
    \begin{subfigure}[t]{0.16\textwidth}
        \includegraphics[width=\textwidth]{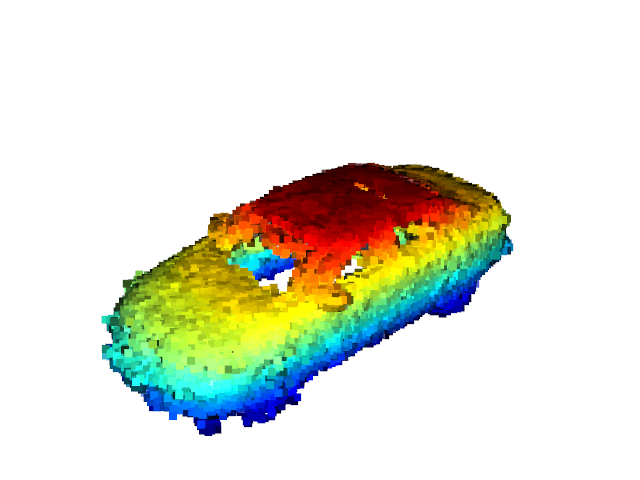}
    \end{subfigure}
    \begin{subfigure}[t]{0.16\textwidth}
        \includegraphics[width=\textwidth]{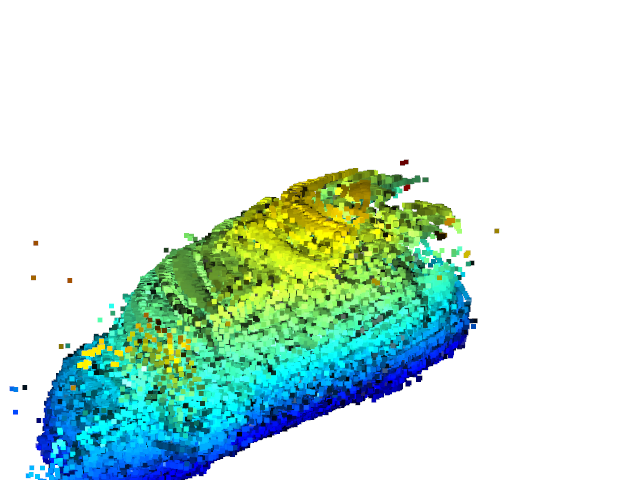}
    \end{subfigure}
    \begin{subfigure}[t]{0.16\textwidth}
        \includegraphics[width=\textwidth]{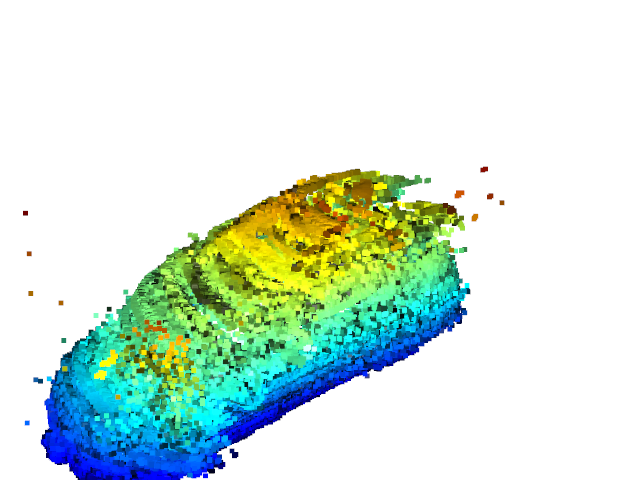}
    \end{subfigure}
    \begin{subfigure}[t]{0.16\textwidth}
        \includegraphics[width=\textwidth]{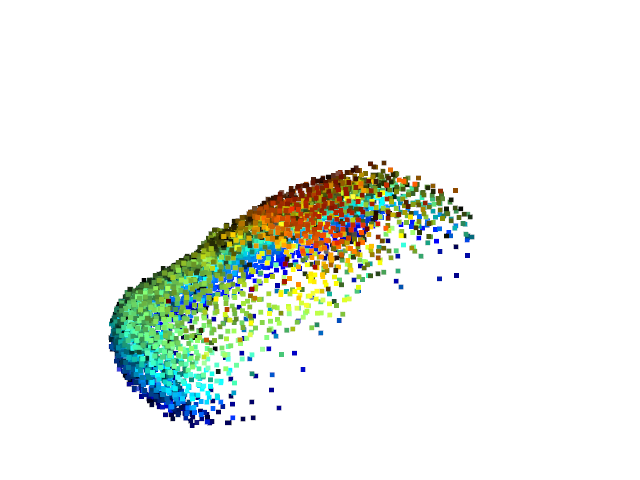}
    \end{subfigure}
    \begin{subfigure}[t]{0.16\textwidth}
        \includegraphics[width=\textwidth]{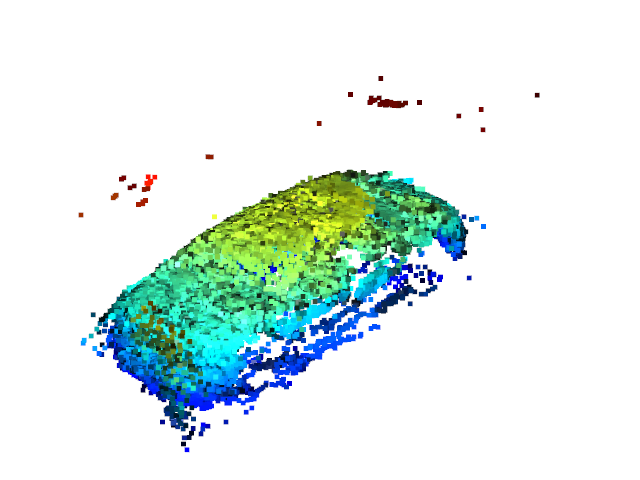}
    \end{subfigure}

    \begin{subfigure}[t]{0.16\textwidth}
        \includegraphics[width=\textwidth]{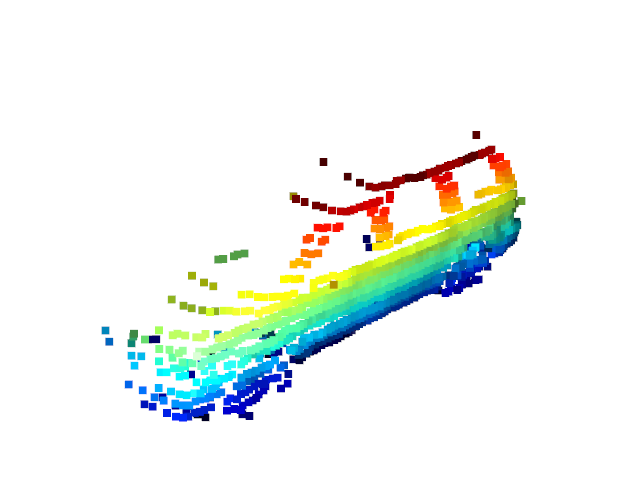}
    \end{subfigure}
    \begin{subfigure}[t]{0.16\textwidth}
        \includegraphics[width=\textwidth]{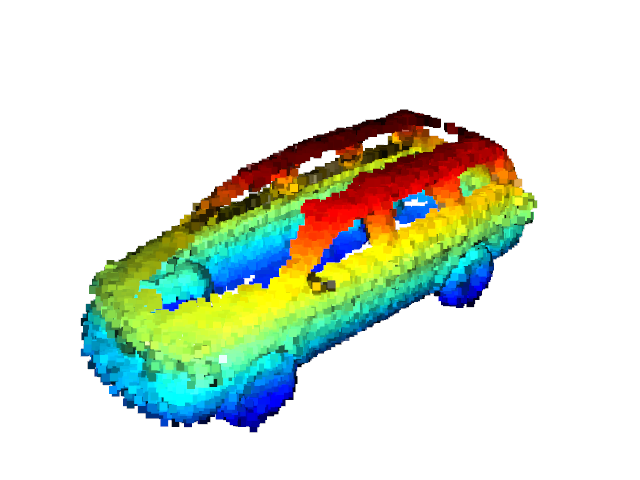}
    \end{subfigure}
    \begin{subfigure}[t]{0.16\textwidth}
        \includegraphics[width=\textwidth]{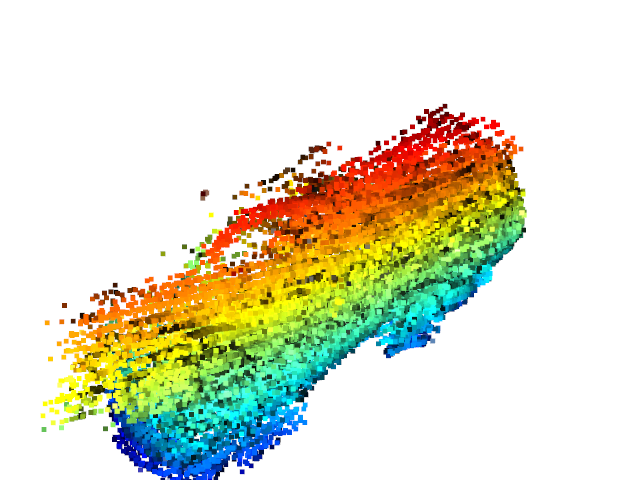}
    \end{subfigure}
    \begin{subfigure}[t]{0.16\textwidth}
        \includegraphics[width=\textwidth]{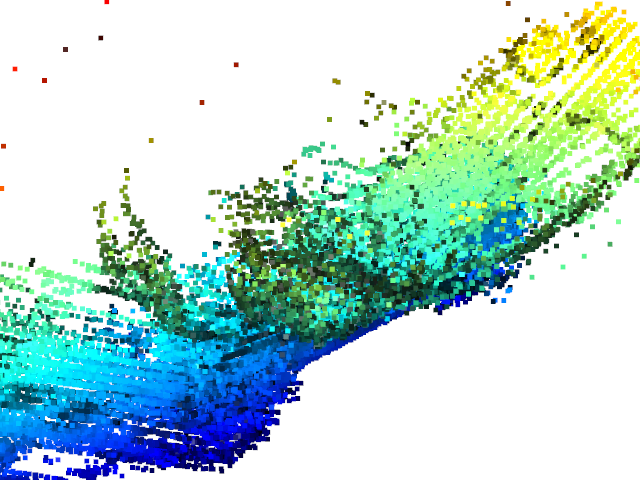}
    \end{subfigure}
    \begin{subfigure}[t]{0.16\textwidth}
        \includegraphics[width=\textwidth]{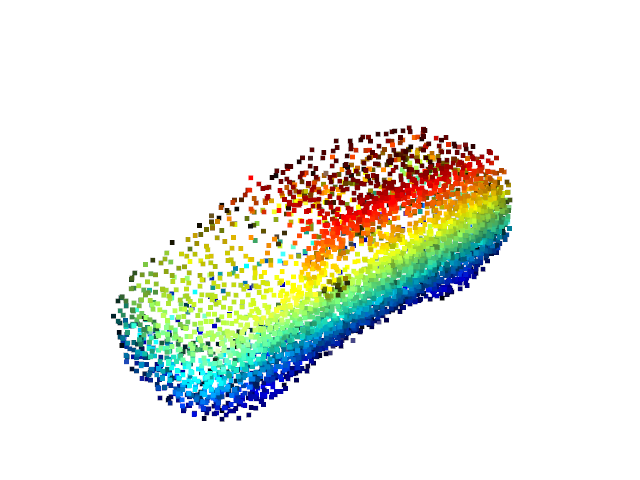}
    \end{subfigure}
    \begin{subfigure}[t]{0.16\textwidth}
        \includegraphics[width=\textwidth]{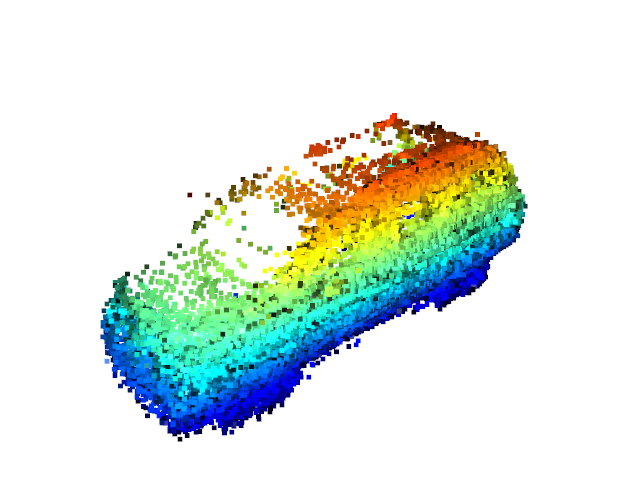}
    \end{subfigure}

    \begin{subfigure}[t]{0.16\textwidth}
        \includegraphics[width=\textwidth]{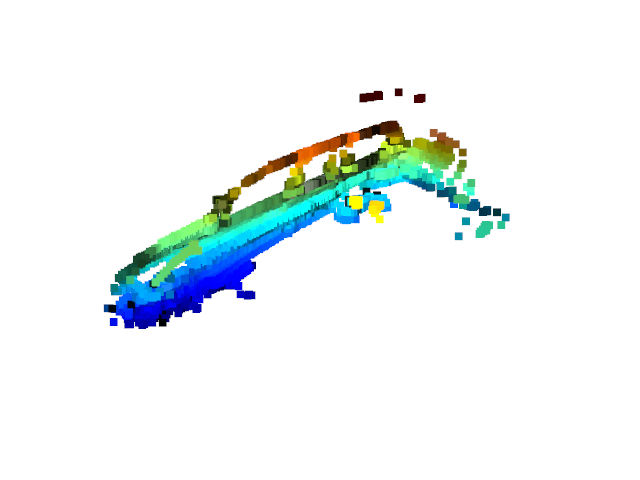}
    \end{subfigure}
    \begin{subfigure}[t]{0.16\textwidth}
        \includegraphics[width=\textwidth]{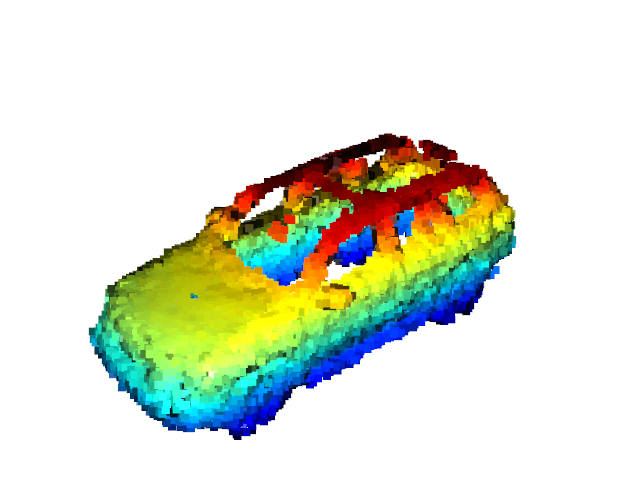}
    \end{subfigure}
    \begin{subfigure}[t]{0.16\textwidth}
        \includegraphics[width=\textwidth]{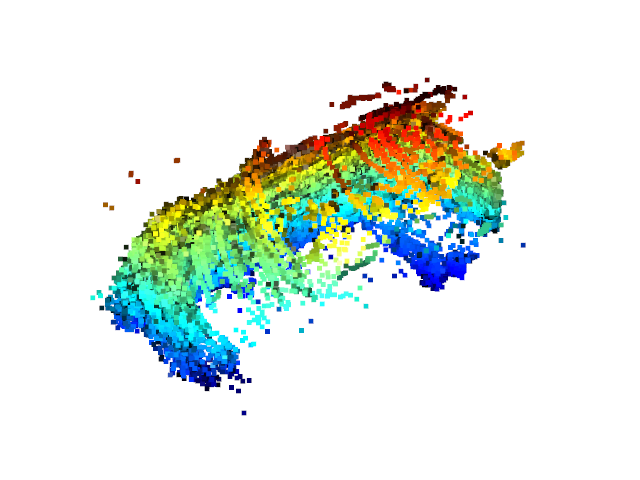}
    \end{subfigure}
    \begin{subfigure}[t]{0.16\textwidth}
        \includegraphics[width=\textwidth]{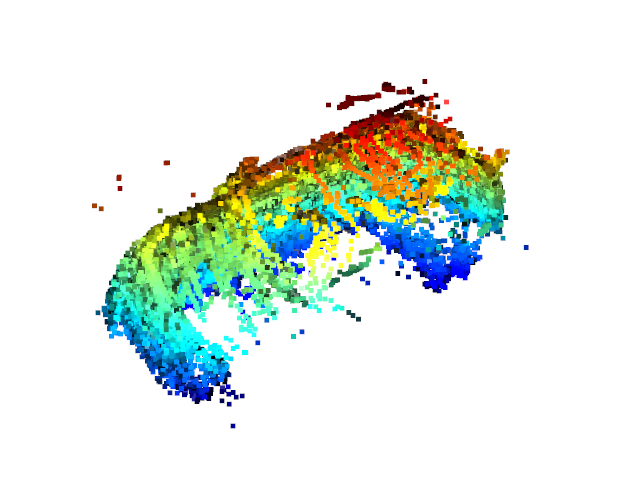}
    \end{subfigure}
    \begin{subfigure}[t]{0.16\textwidth}
        \includegraphics[width=\textwidth]{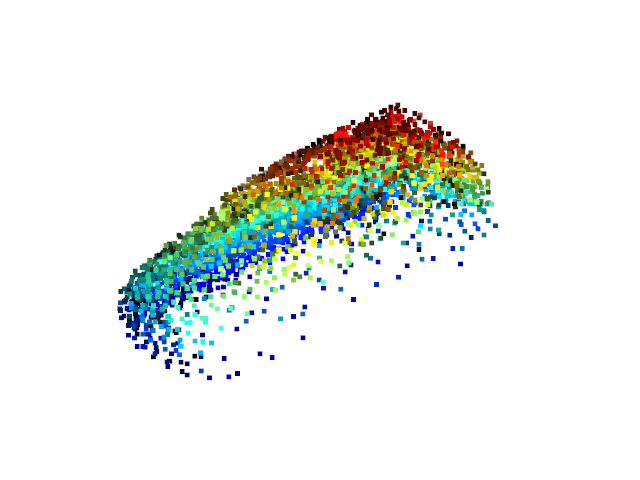}
    \end{subfigure}
    \begin{subfigure}[t]{0.16\textwidth}
        \includegraphics[width=\textwidth]{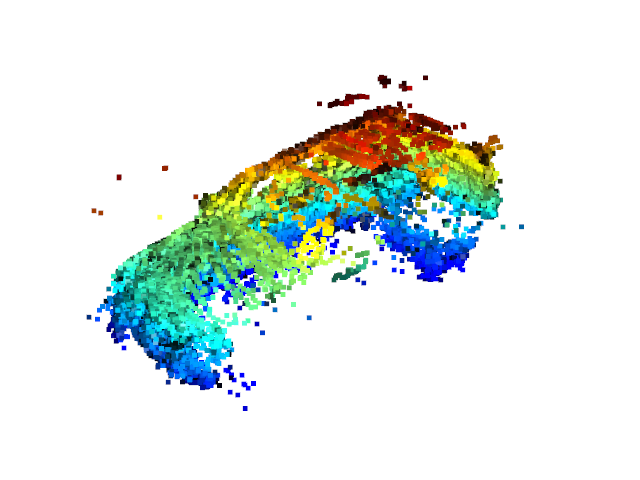}
    \end{subfigure}

    \caption{Qualitative results of our method compared against ground-truth and ICP on our 3D vehicle dataset.
        All the point clouds are transformed to the ground-truth canonical frame and visualized at a fixed viewpoint.
        We denote our approach for 3D shape completion and point cloud registration by \emph{Ours(shape)} and \emph{Ours(registration)}.
    }
    \label{fig:tor4d-supp}
\end{figure}

\begin{figure}[h]
    \centering
    \begin{subfigure}[t]{0.16\textwidth}
        \caption*{Input}
        \vspace{-0.5em}
        \includegraphics[width=\textwidth]{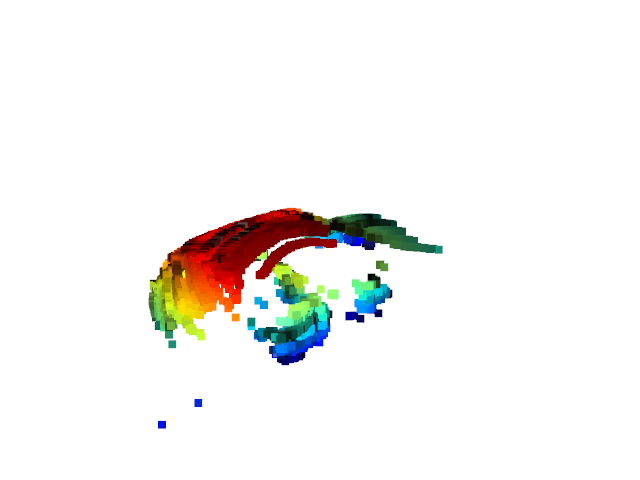}
    \end{subfigure}
    \begin{subfigure}[t]{0.16\textwidth}
        \caption*{Ground truth}
        \vspace{-0.5em}
        \includegraphics[width=\textwidth]{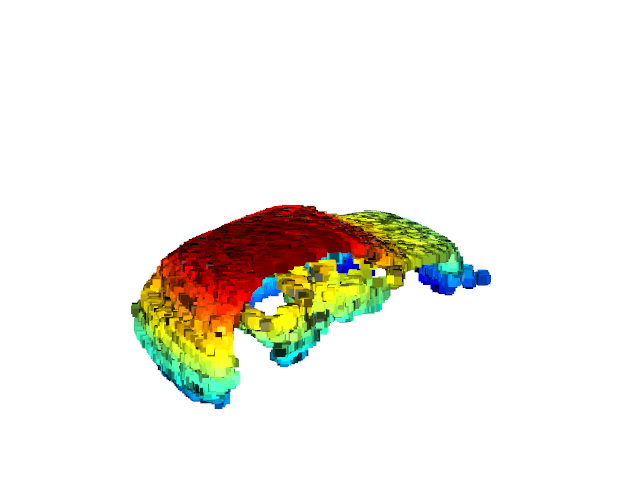}
    \end{subfigure}
    \begin{subfigure}[t]{0.16\textwidth}
        \caption*{Local-ICP}
        \vspace{-0.5em}
        \includegraphics[width=\textwidth]{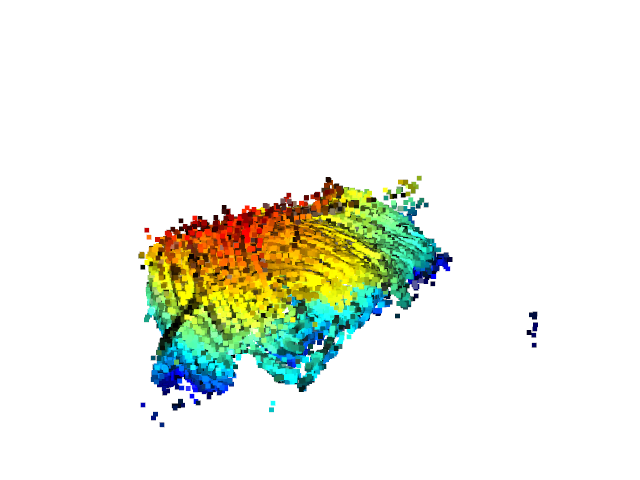}
    \end{subfigure}
    \begin{subfigure}[t]{0.16\textwidth}
        \caption*{Global-ICP}
        \vspace{-0.5em}
        \includegraphics[width=\textwidth]{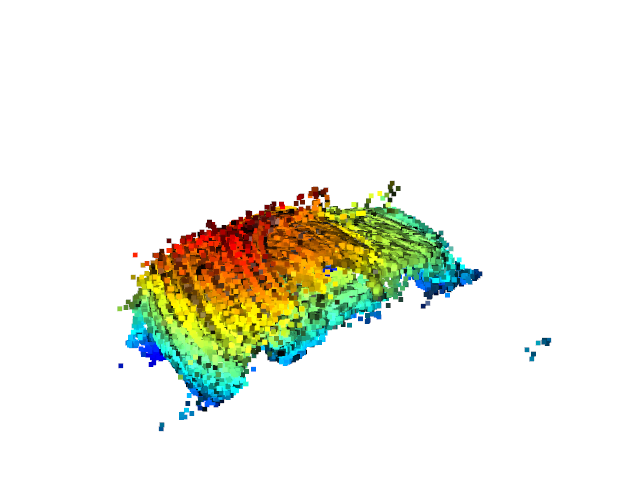}
    \end{subfigure}
    \begin{subfigure}[t]{0.16\textwidth}
        \caption*{Ours(shape)}
        \vspace{-0.5em}
        \includegraphics[width=\textwidth]{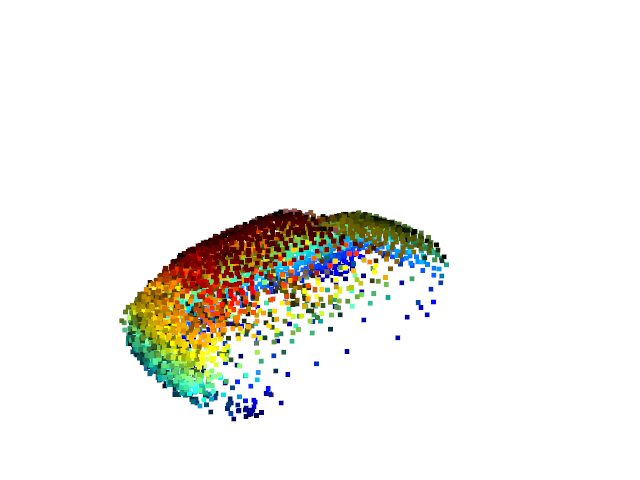}
    \end{subfigure}
    \begin{subfigure}[t]{0.16\textwidth}
        \caption*{Ours(registration)}
        \vspace{-0.5em}
        \includegraphics[width=\textwidth]{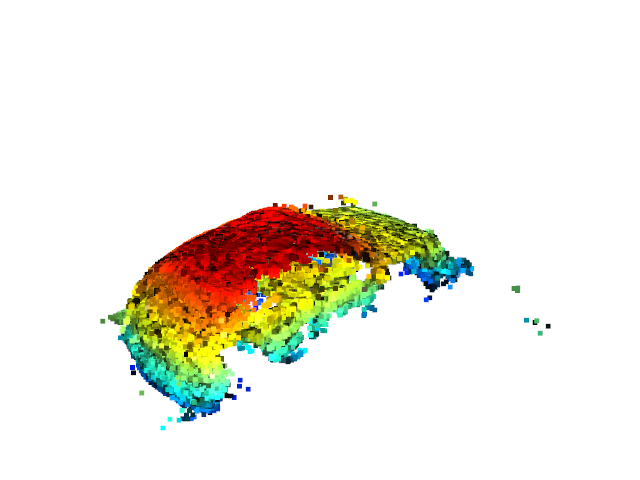}
    \end{subfigure}

    \begin{subfigure}[t]{0.16\textwidth}
        \includegraphics[width=\textwidth]{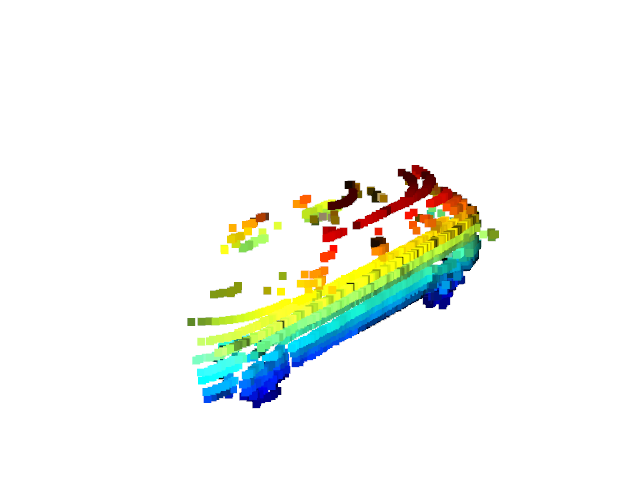}
    \end{subfigure}
    \begin{subfigure}[t]{0.16\textwidth}
        \includegraphics[width=\textwidth]{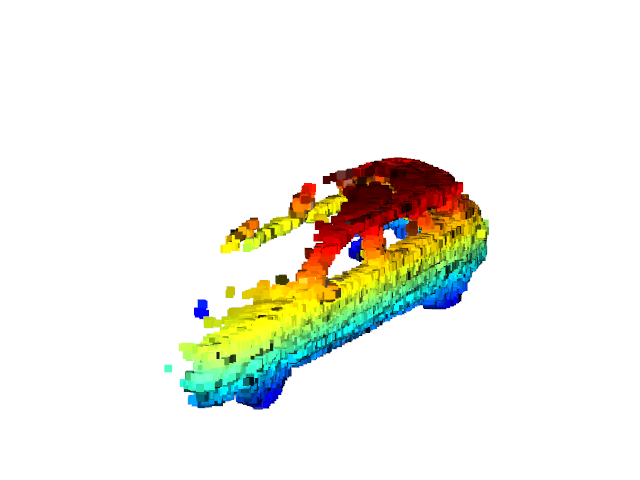}
    \end{subfigure}
    \begin{subfigure}[t]{0.16\textwidth}
        \includegraphics[width=\textwidth]{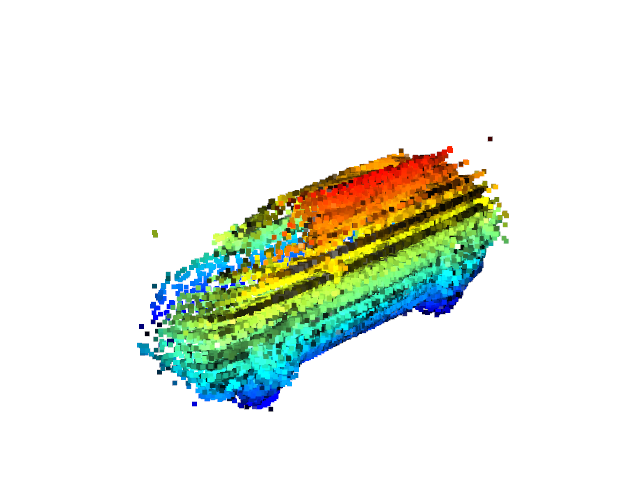}
    \end{subfigure}
    \begin{subfigure}[t]{0.16\textwidth}
        \includegraphics[width=\textwidth]{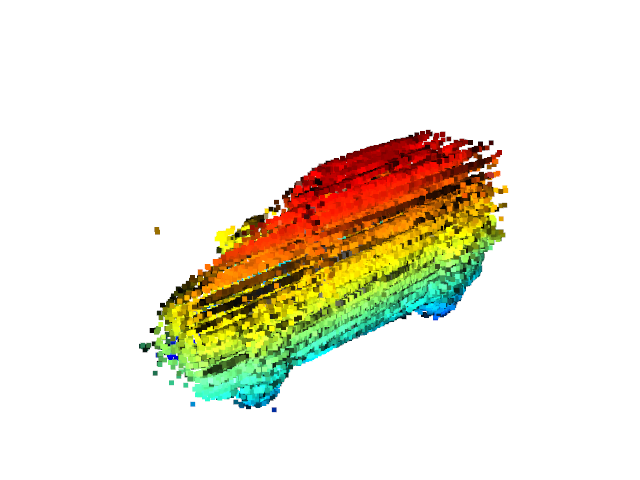}
    \end{subfigure}
    \begin{subfigure}[t]{0.16\textwidth}
        \includegraphics[width=\textwidth]{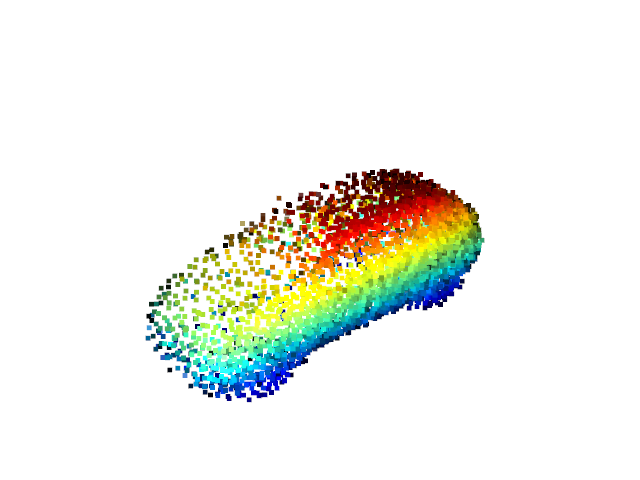}
    \end{subfigure}
    \begin{subfigure}[t]{0.16\textwidth}
        \includegraphics[width=\textwidth]{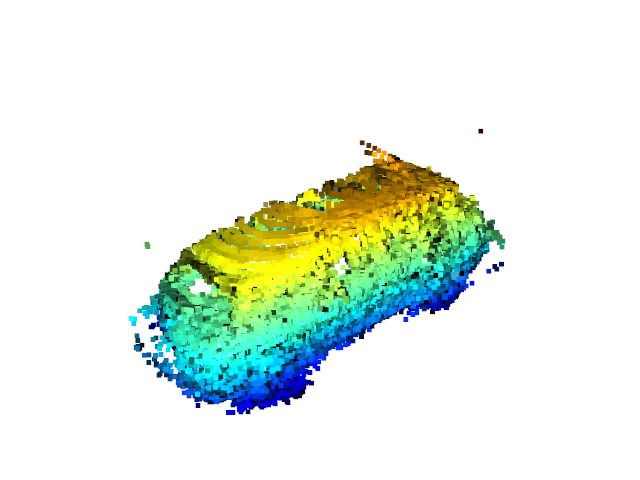}
    \end{subfigure}

    \begin{subfigure}[t]{0.16\textwidth}
        \includegraphics[width=\textwidth]{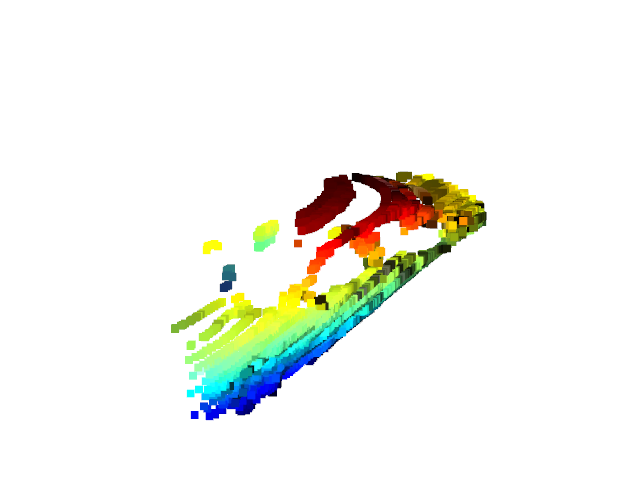}
    \end{subfigure}
    \begin{subfigure}[t]{0.16\textwidth}
        \includegraphics[width=\textwidth]{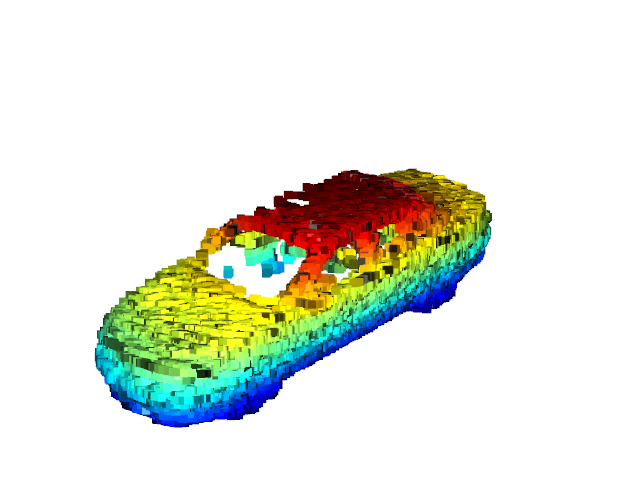}
    \end{subfigure}
    \begin{subfigure}[t]{0.16\textwidth}
        \includegraphics[width=\textwidth]{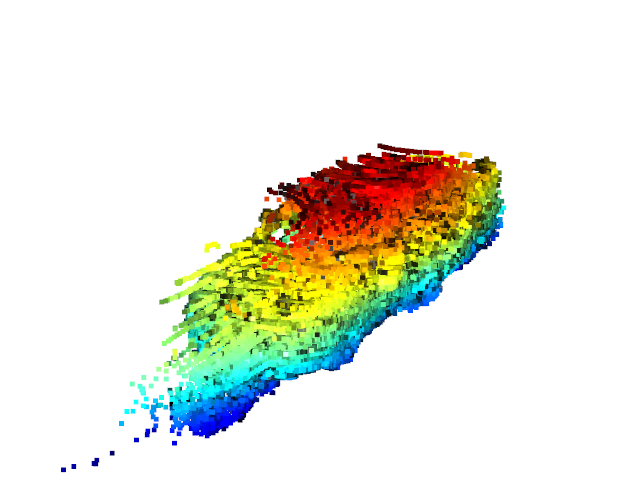}
    \end{subfigure}
    \begin{subfigure}[t]{0.16\textwidth}
        \includegraphics[width=\textwidth]{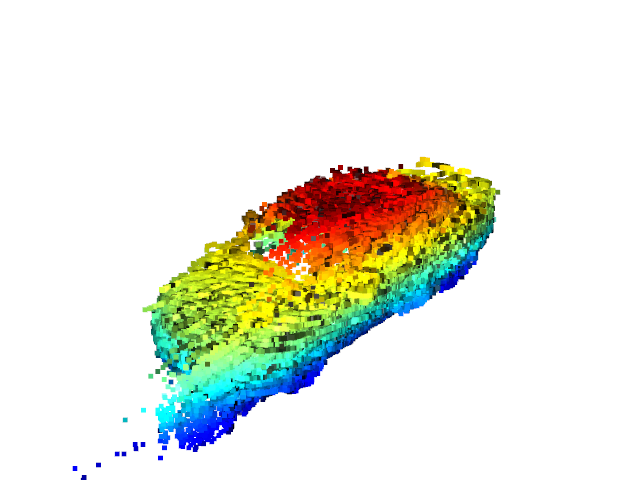}
    \end{subfigure}
    \begin{subfigure}[t]{0.16\textwidth}
        \includegraphics[width=\textwidth]{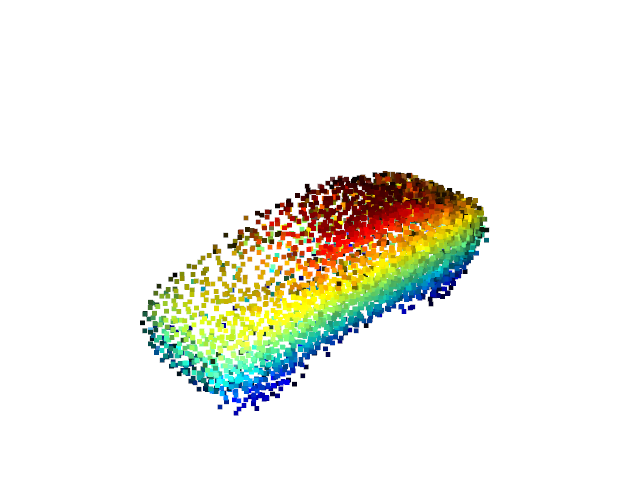}
    \end{subfigure}
    \begin{subfigure}[t]{0.16\textwidth}
        \includegraphics[width=\textwidth]{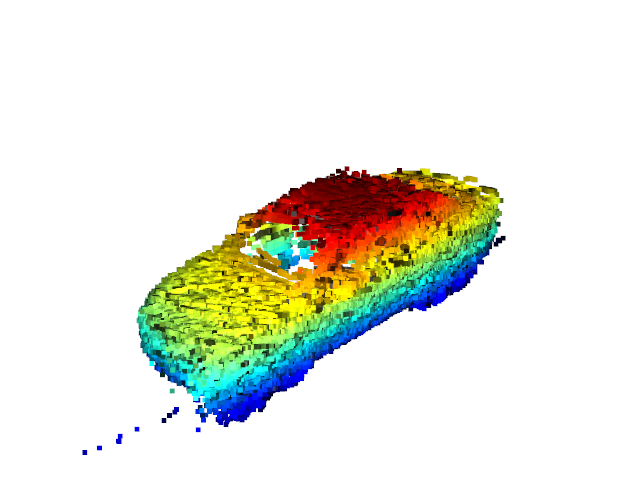}
    \end{subfigure}

    \begin{subfigure}[t]{0.16\textwidth}
        \includegraphics[width=\textwidth]{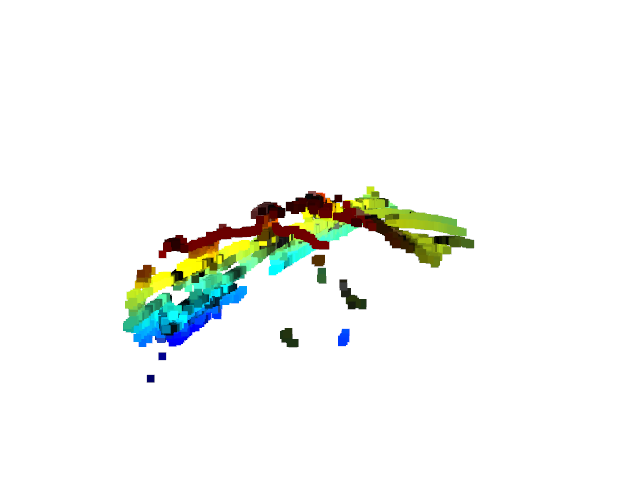}
    \end{subfigure}
    \begin{subfigure}[t]{0.16\textwidth}
        \includegraphics[width=\textwidth]{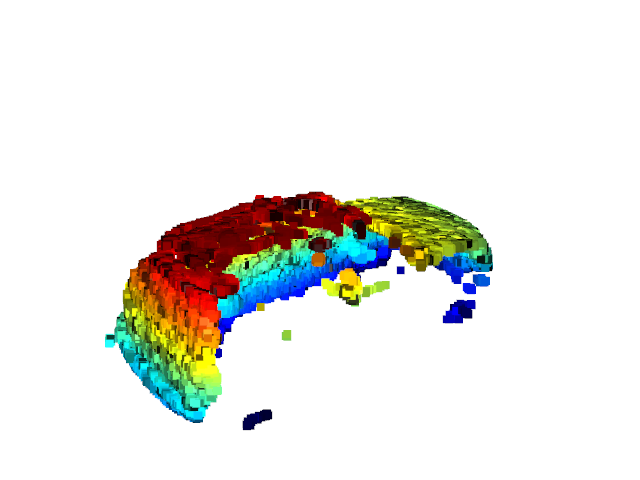}
    \end{subfigure}
    \begin{subfigure}[t]{0.16\textwidth}
        \includegraphics[width=\textwidth]{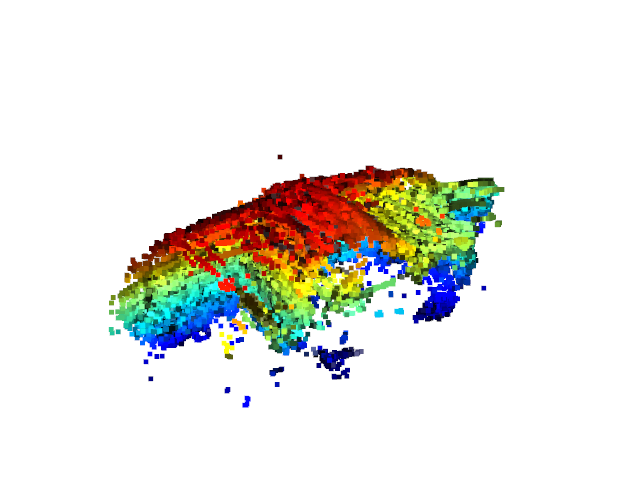}
    \end{subfigure}
    \begin{subfigure}[t]{0.16\textwidth}
        \includegraphics[width=\textwidth]{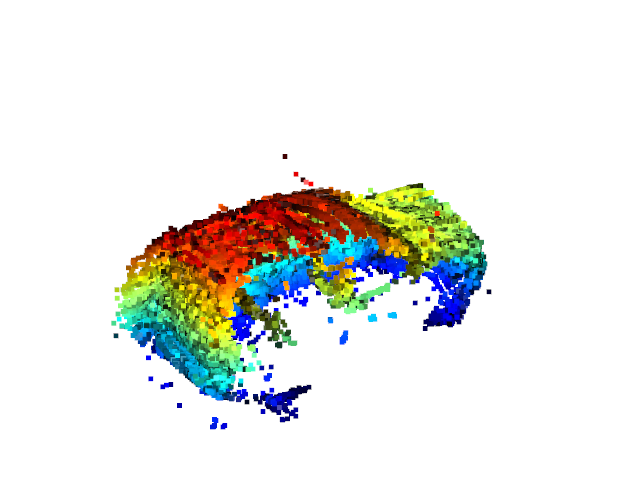}
    \end{subfigure}
    \begin{subfigure}[t]{0.16\textwidth}
        \includegraphics[width=\textwidth]{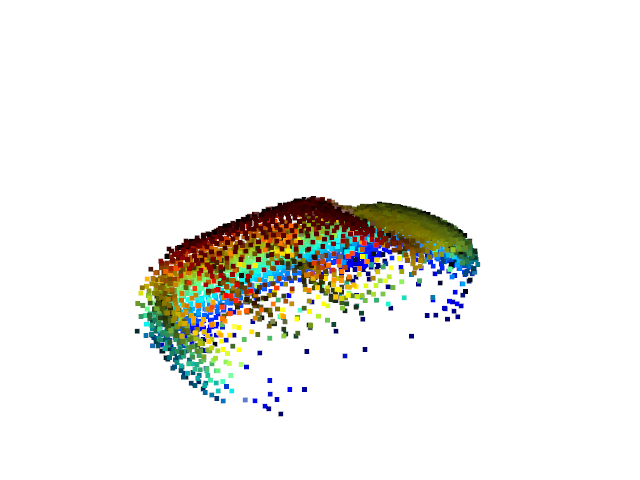}
    \end{subfigure}
    \begin{subfigure}[t]{0.16\textwidth}
        \includegraphics[width=\textwidth]{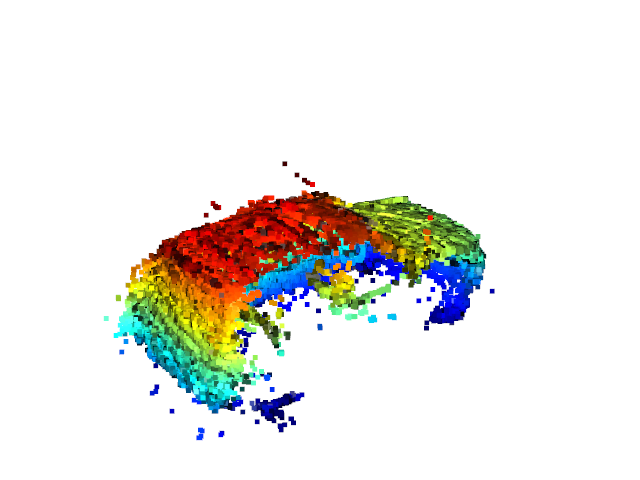}
    \end{subfigure}

    \begin{subfigure}[t]{0.16\textwidth}
        \includegraphics[width=\textwidth]{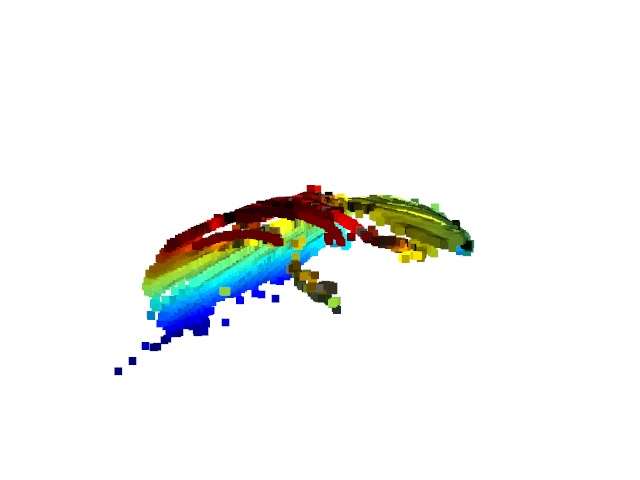}
    \end{subfigure}
    \begin{subfigure}[t]{0.16\textwidth}
        \includegraphics[width=\textwidth]{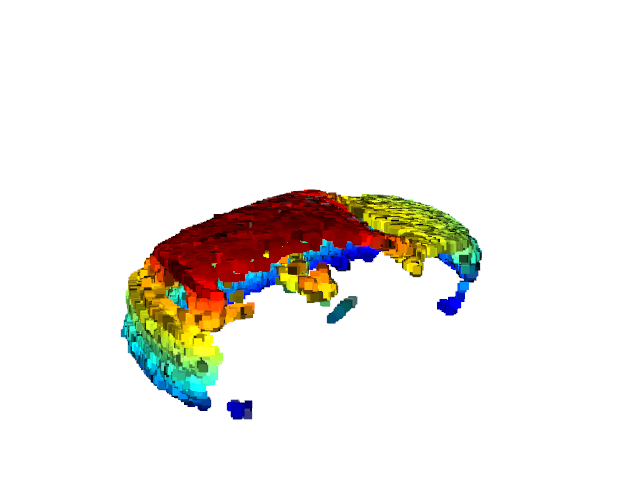}
    \end{subfigure}
    \begin{subfigure}[t]{0.16\textwidth}
        \includegraphics[width=\textwidth]{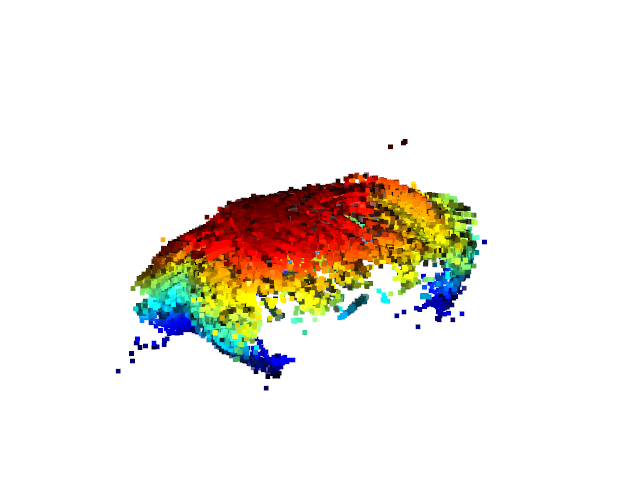}
    \end{subfigure}
    \begin{subfigure}[t]{0.16\textwidth}
        \includegraphics[width=\textwidth]{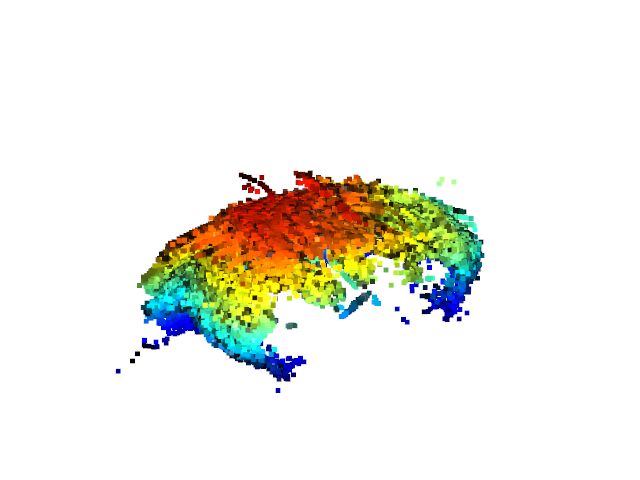}
    \end{subfigure}
    \begin{subfigure}[t]{0.16\textwidth}
        \includegraphics[width=\textwidth]{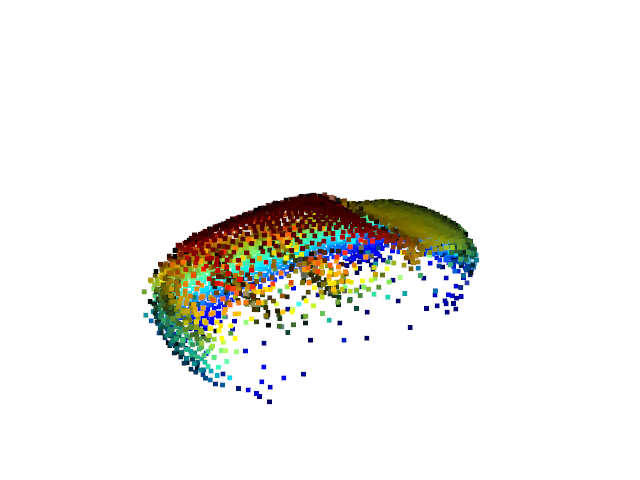}
    \end{subfigure}
    \begin{subfigure}[t]{0.16\textwidth}
        \includegraphics[width=\textwidth]{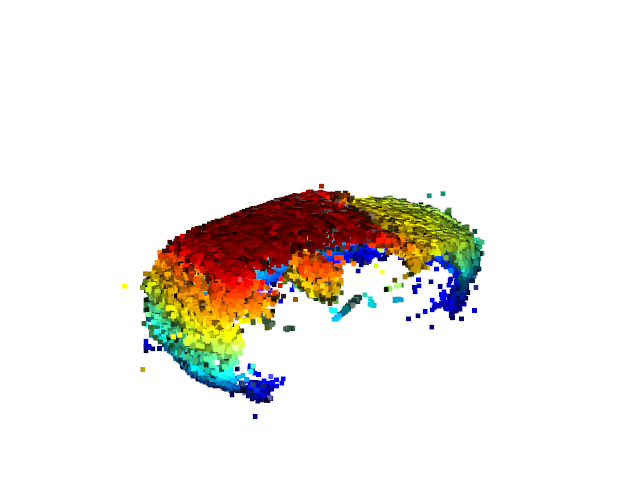}
    \end{subfigure}

    \begin{subfigure}[t]{0.16\textwidth}
        \includegraphics[width=\textwidth]{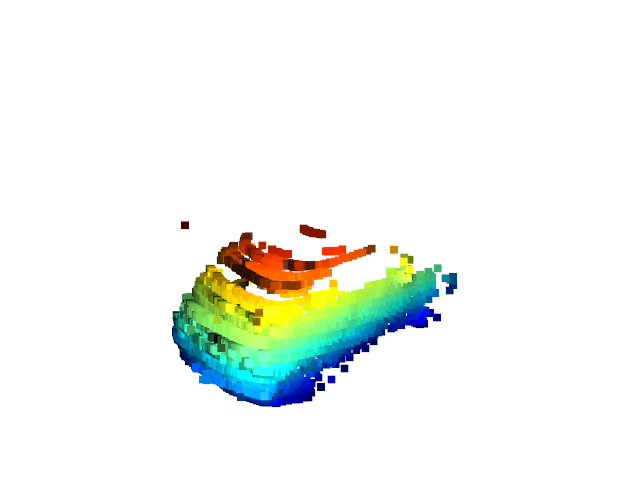}
    \end{subfigure}
    \begin{subfigure}[t]{0.16\textwidth}
        \includegraphics[width=\textwidth]{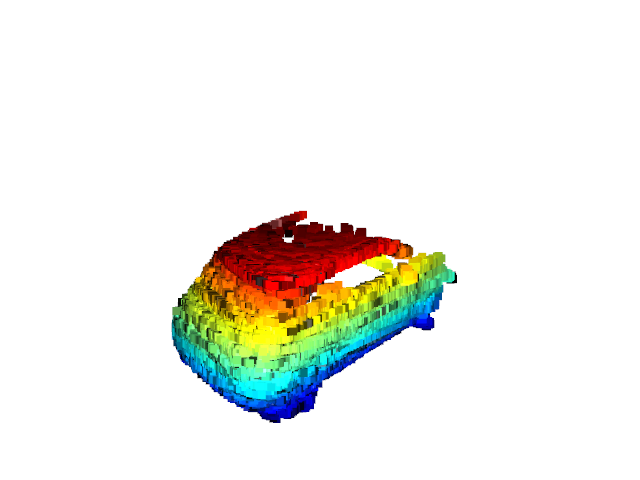}
    \end{subfigure}
    \begin{subfigure}[t]{0.16\textwidth}
        \includegraphics[width=\textwidth]{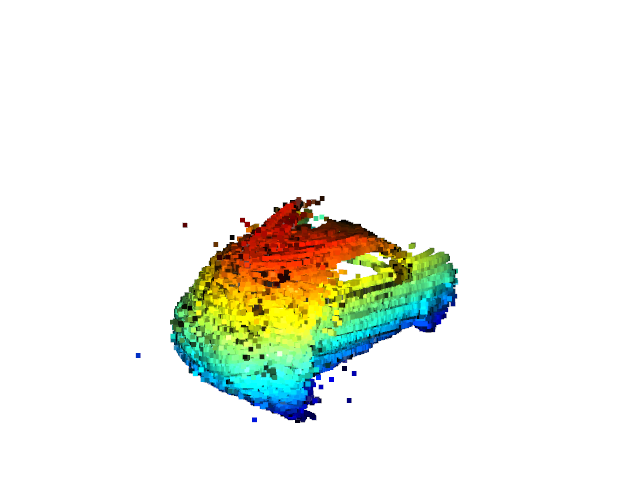}
    \end{subfigure}
    \begin{subfigure}[t]{0.16\textwidth}
        \includegraphics[width=\textwidth]{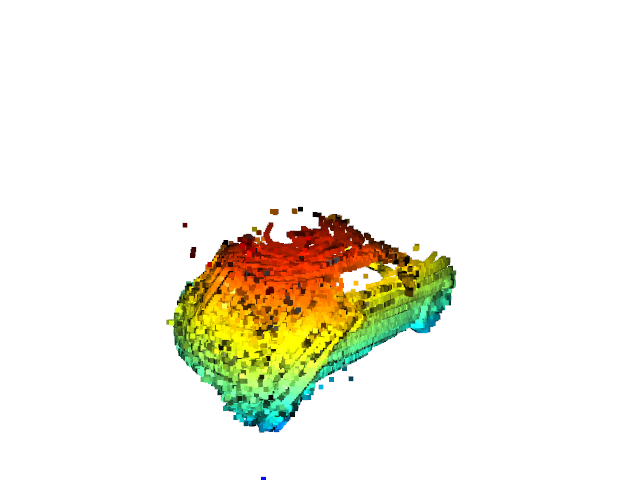}
    \end{subfigure}
    \begin{subfigure}[t]{0.16\textwidth}
        \includegraphics[width=\textwidth]{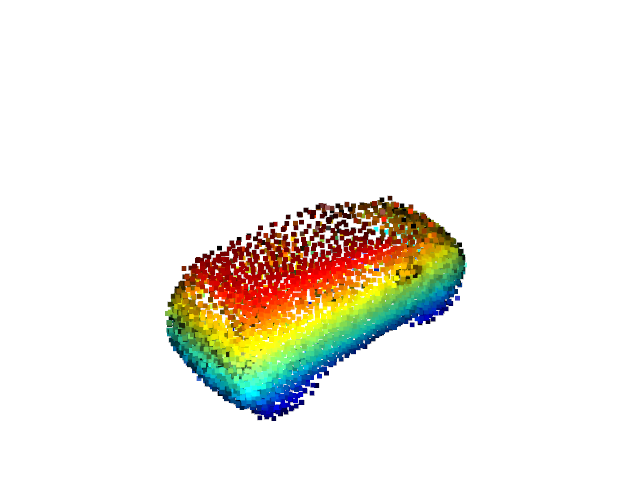}
    \end{subfigure}
    \begin{subfigure}[t]{0.16\textwidth}
        \includegraphics[width=\textwidth]{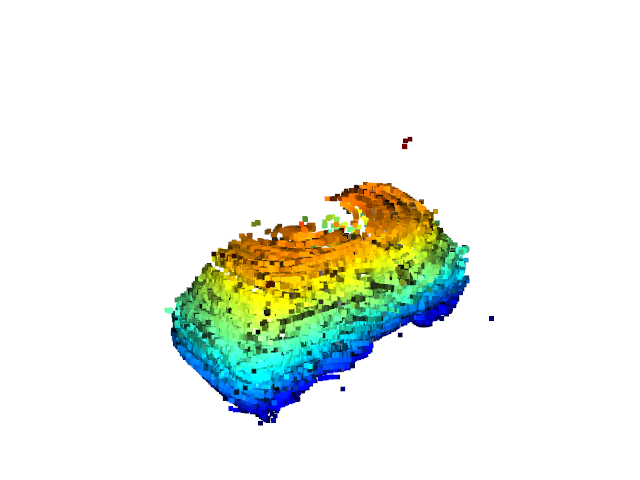}
    \end{subfigure}

    \begin{subfigure}[t]{0.16\textwidth}
        \includegraphics[width=\textwidth]{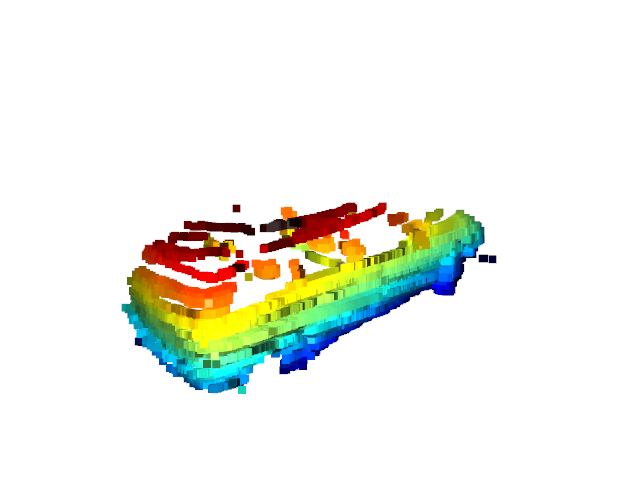}
    \end{subfigure}
    \begin{subfigure}[t]{0.16\textwidth}
        \includegraphics[width=\textwidth]{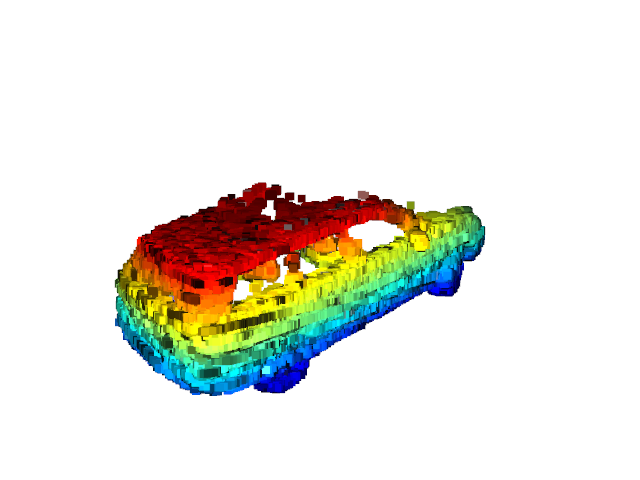}
    \end{subfigure}
    \begin{subfigure}[t]{0.16\textwidth}
        \includegraphics[width=\textwidth]{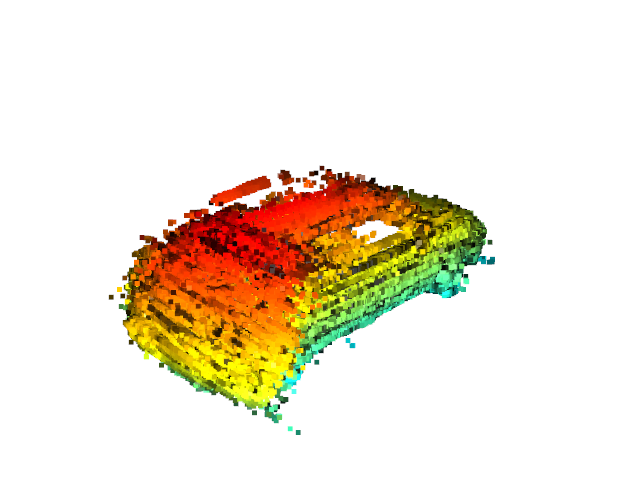}
    \end{subfigure}
    \begin{subfigure}[t]{0.16\textwidth}
        \includegraphics[width=\textwidth]{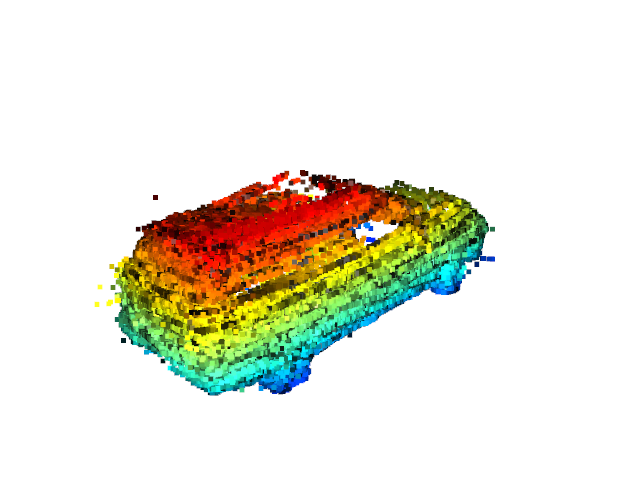}
    \end{subfigure}
    \begin{subfigure}[t]{0.16\textwidth}
        \includegraphics[width=\textwidth]{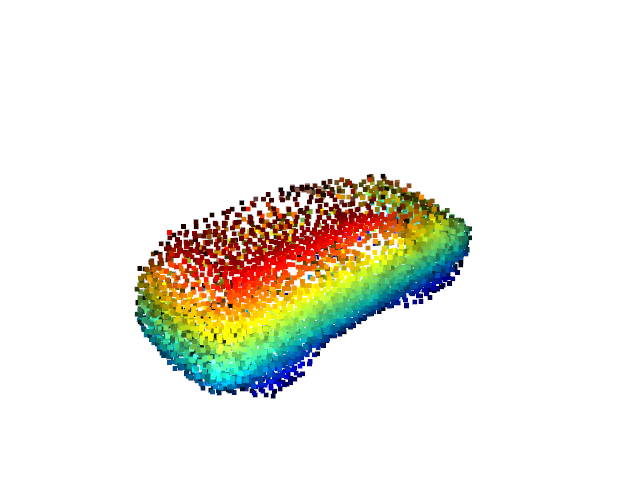}
    \end{subfigure}
    \begin{subfigure}[t]{0.16\textwidth}
        \includegraphics[width=\textwidth]{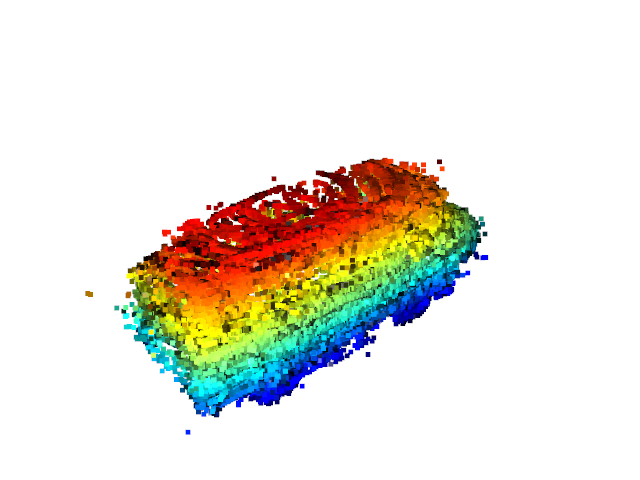}
    \end{subfigure}

    \begin{subfigure}[t]{0.16\textwidth}
        \includegraphics[width=\textwidth]{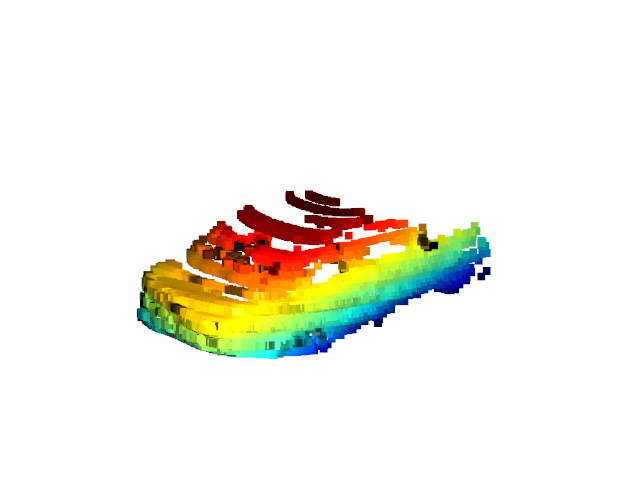}
    \end{subfigure}
    \begin{subfigure}[t]{0.16\textwidth}
        \includegraphics[width=\textwidth]{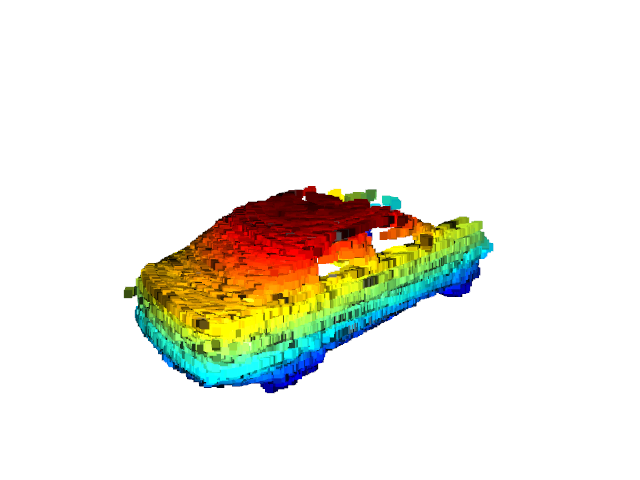}
    \end{subfigure}
    \begin{subfigure}[t]{0.16\textwidth}
        \includegraphics[width=\textwidth]{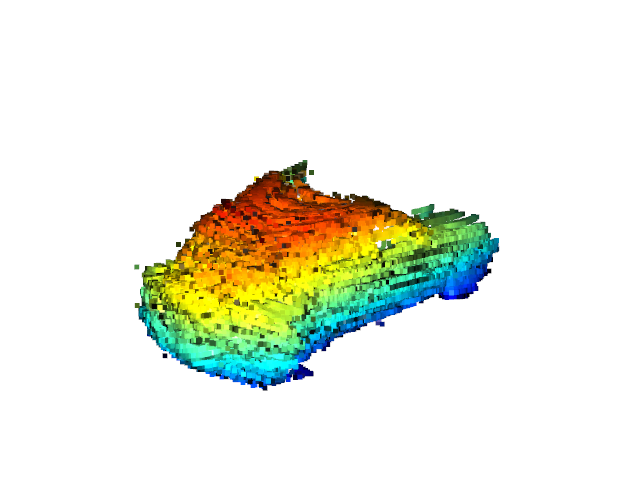}
    \end{subfigure}
    \begin{subfigure}[t]{0.16\textwidth}
        \includegraphics[width=\textwidth]{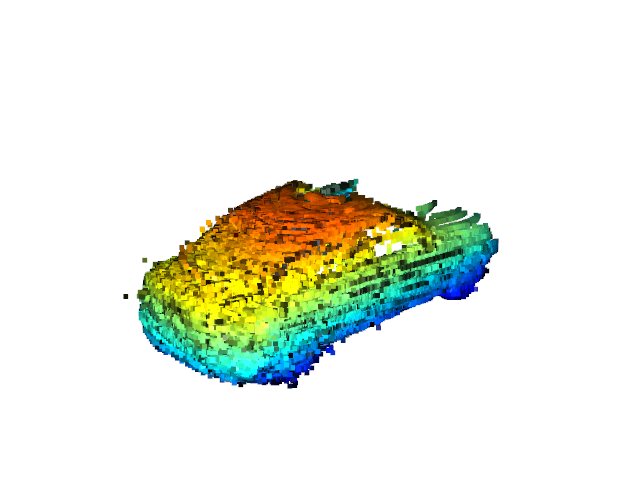}
    \end{subfigure}
    \begin{subfigure}[t]{0.16\textwidth}
        \includegraphics[width=\textwidth]{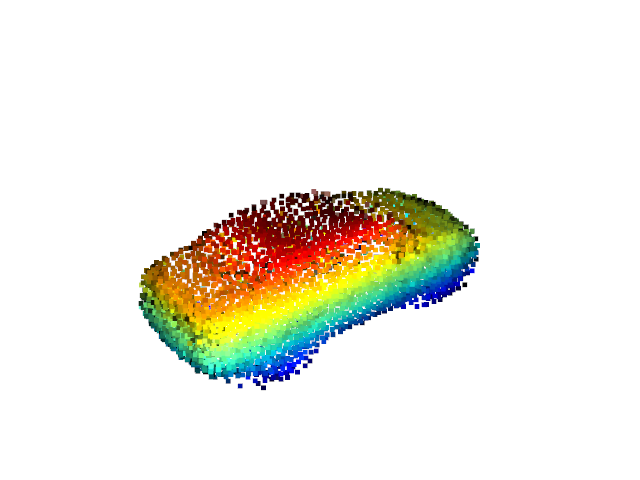}
    \end{subfigure}
    \begin{subfigure}[t]{0.16\textwidth}
        \includegraphics[width=\textwidth]{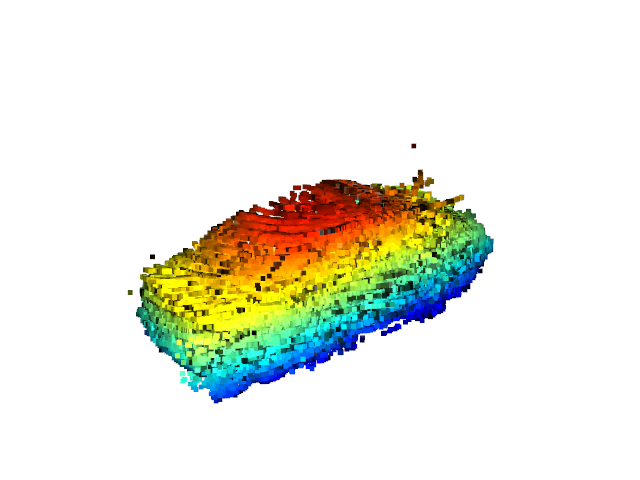}
    \end{subfigure}

    \begin{subfigure}[t]{0.16\textwidth}
        \includegraphics[width=\textwidth]{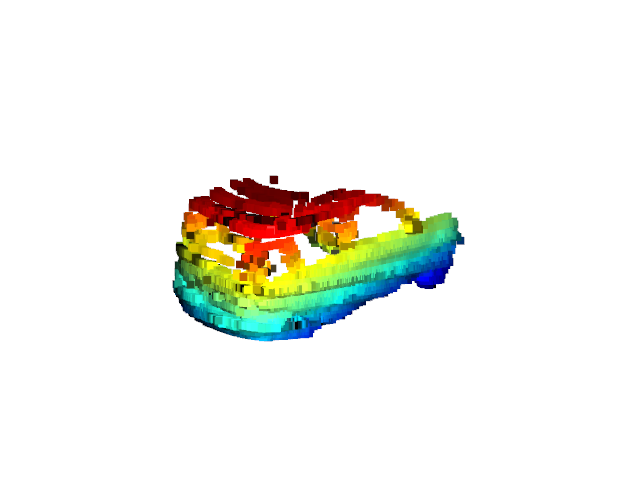}
    \end{subfigure}
    \begin{subfigure}[t]{0.16\textwidth}
        \includegraphics[width=\textwidth]{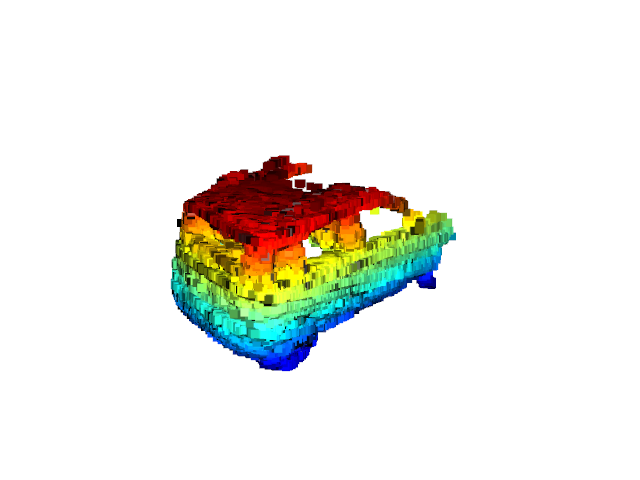}
    \end{subfigure}
    \begin{subfigure}[t]{0.16\textwidth}
        \includegraphics[width=\textwidth]{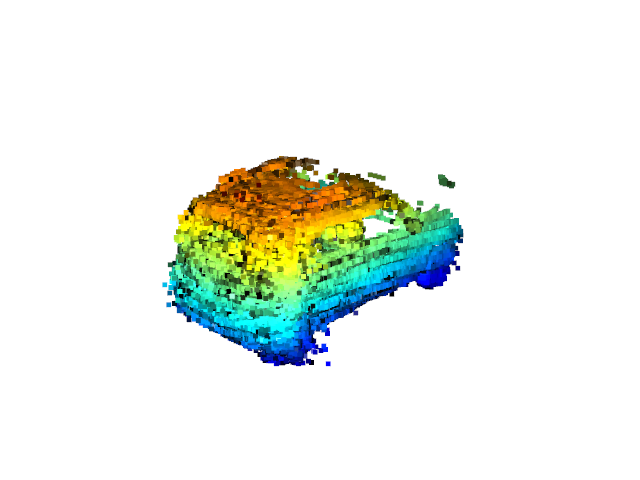}
    \end{subfigure}
    \begin{subfigure}[t]{0.16\textwidth}
        \includegraphics[width=\textwidth]{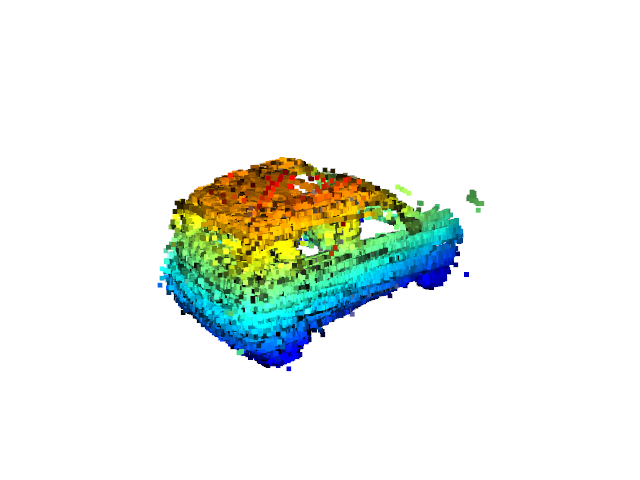}
    \end{subfigure}
    \begin{subfigure}[t]{0.16\textwidth}
        \includegraphics[width=\textwidth]{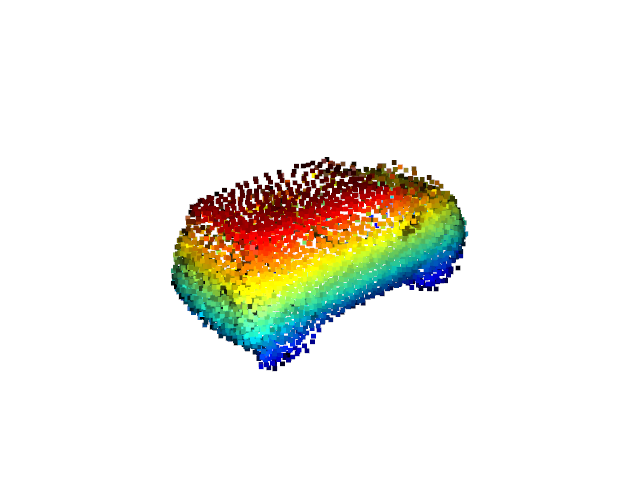}
    \end{subfigure}
    \begin{subfigure}[t]{0.16\textwidth}
        \includegraphics[width=\textwidth]{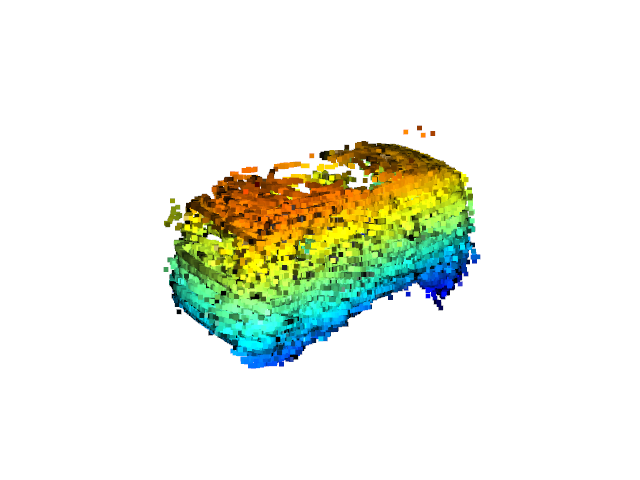}
    \end{subfigure}

    \caption{Qualitative results of our method compared against ground-truth and ICP on SemanticKITTI.
        All the point clouds are transformed to the ground-truth canonical frame and visualized at a fixed viewpoint.
        We denote our approach for 3D shape completion and point cloud registration by \emph{Ours(shape)} and \emph{Ours(registration)}.
    }
    \label{fig:skitti-supp}
\end{figure}

Fig~\ref{fig:tor4d-supp} and Fig~\ref{fig:skitti-supp} include more qualitative results on real LiDAR datasets.
Apart from shape completion, our weakly-supervised approach can be easily extended to point cloud registration.
As our method estimates the 6-DoF pose of the canonical shape, we can estimate the transformation from one partial point cloud to another, by first transforming the source point cloud to the canonical frame and then to the sensor coordinate system of the target point cloud.
We select the middle frame of a sequence as the target, and fuse all the partial observations in a sequence according to estimated transformations.
Fused point clouds are visualized in the last column (\emph{Ours(fusion)}) of Fig~\ref{fig:tor4d-supp}.
Although the predicted complete shape of our method lacks fine details, the estimated pose is accurate, and thus the fused point cloud is very close to the ground truth.
Our method outperforms ICP methods,
which implies that the knowledge of the complete shape eases the challenging problem of partial point cloud registration, especially for real, sparse point clouds.

\begin{figure}
    \centering
    \includegraphics[width=\textwidth]{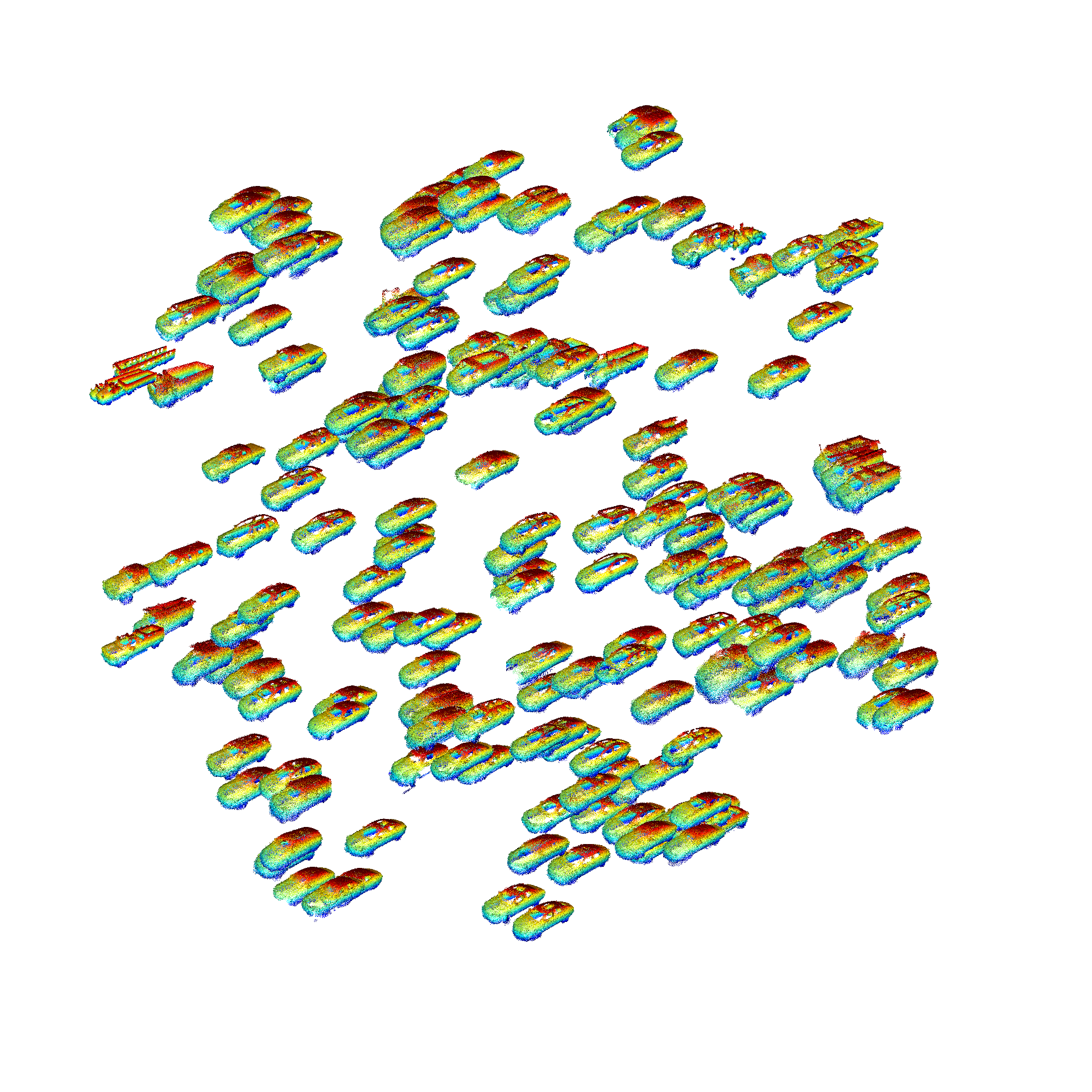}
    \caption{t-SNE visualization of the shape features learned from our 3D vehicle dataset. 200 samples from different instances are randomly chosen from the validation set. For each sample, we visualize its corresponding GT point cloud.}
    \label{fig:t-SNE}
\end{figure}

Moreover, we show t-SNE visualization of the shape features learned from our 3D vehicle dataset in Fig~\ref{fig:t-SNE}.
Close features correspond to instances with similar shapes, which indicates that the learned shape features are meaningful.


\section{Sensitivity to initialization}
\label{sec:sensitivity}
It is intuitive that the randomness of initialization and optimization will lead to very different results for not fully-supervised approaches.
Thus, we would like to investigate how sensitive our method as well as other not fully-supervised baselines are to initialization.
Table~\ref{tab:shapenet-sensitivity} shows the average and standard deviation of 3 trials on ShapeNet.
It is observed that our method shows a lower variance compared to DPC~\cite{insafutdinov2018unsupervised} in general.
In addtion, Table~\ref{tab:real-data-sensitivity} shows the average and standard deviation of 5 trials on real LiDAR datasets.
It is worthy of future work to study how to lower the variance.

\begin{table}[ht]
    \small
    \centering
    \setlength{\tabcolsep}{3pt}
    \renewcommand{\arraystretch}{1.1}
    \begin{tabular}{l|c|cc}
        \toprule
        Category & Method        & CD           & Acc($\leq 30^\circ$) \\
        \midrule
        \multirow{2}{*}{Airplane}
                 & DPC           & 7.20 (0.81)  & 76.11 (1.69)         \\
                 & DPC$^\dagger$ & 17.21 (3.59) & 34.83 (18.07)        \\
                 & Ours          & 1.95 (0.03)  & 90.87 (3.40)         \\
        \midrule
        \multirow{2}{*}{Car}
                 & DPC           & 3.64 (0.13)  & 83.33 (1.26)         \\
                 & DPC$^\dagger$ & 9.66 (4.31)  & 35.73 (40.47)        \\
                 & Ours          & 2.66 (0.05)  & 49.58 (0.58)         \\
        \midrule
        \multirow{2}{*}{Chair}
                 & DPC           & 6.24 (1.64)  & 57.13 (26.67)        \\
                 & DPC$^\dagger$ & 7.38 (0.05)  & 69.46 (0.91)         \\
                 & Ours          & 3.33 (0.002) & 95.20 (0.65)         \\
        \bottomrule
    \end{tabular}
    \caption{
        We report the chamfer distance and the pose accuracy of 3 trials on the test set of ShapeNet.
        The chamfer distance is multiplied by 100.
        The average with the standard deviation (in the parentheses) is reported.
    }
    \label{tab:shapenet-sensitivity}
\end{table}

\begin{table}[ht]
    \small
    \centering
    \setlength{\tabcolsep}{2.5pt}
    \renewcommand{\arraystretch}{1.1}
    \begin{tabular}{l|c|cccc}
        \toprule
        Dataset            & CD           & Acc($\leq 30^\circ$) & Rot $\Delta \theta$ & Trans $\Delta t$ \\
        \midrule
        3D vehicle dataset & 0.26 (0.009) & 76.54 (19.20)        & 4.21 (1.72)         & 0.16 (0.032)     \\
        SemanticKITTI      & 0.20 (0.09)  & 60.62 (19.17)        & 11.54 (6.26)     & 0.21 (0.032) \\
        \bottomrule
    \end{tabular}
    \caption{
        We report the chamfer distance, the pose accuracy, the median angle difference and the median translation MSE of 5 trials on the test set of real LiDAR datasets.
        The average with the standard deviation (in the parentheses) is reported.
    }
    \label{tab:real-data-sensitivity}
\end{table}

\section{Implementation details of DPC-LIDAR}
\label{sec:dpc-lidar}

\begin{figure}[ht]
    \centering
    \begin{subfigure}[t]{0.15\textwidth}
        \caption*{Input}
        \vspace{-0.5em}
        \includegraphics[width=\textwidth]{figures/tor4d_supp/example1_input.png}
    \end{subfigure}
    \begin{subfigure}[t]{0.15\textwidth}
        \caption*{GT}
        \vspace{-0.5em}
        \includegraphics[width=\textwidth]{figures/tor4d_supp/example1_gt.png}
    \end{subfigure}
    \begin{subfigure}[t]{0.15\textwidth}
        \caption*{DPC-LIDAR}
        \vspace{-0.5em}
        \includegraphics[width=\textwidth]{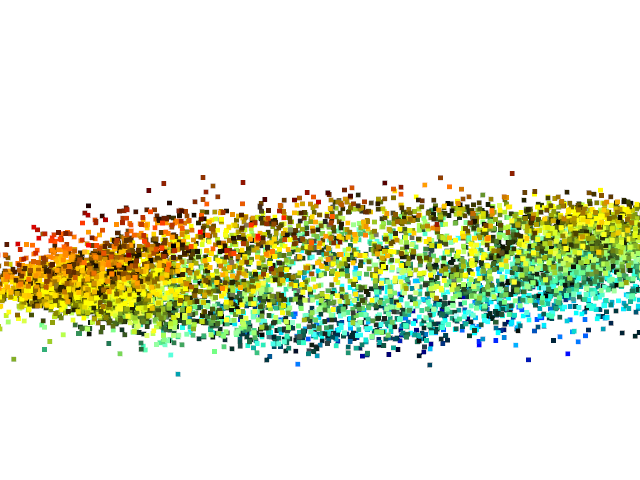}
    \end{subfigure}
    \begin{subfigure}[t]{0.15\textwidth}
        \caption*{Input}
        \vspace{-0.5em}
        \includegraphics[width=\textwidth]{figures/tor4d_supp/example2_input.png}
    \end{subfigure}
    \begin{subfigure}[t]{0.15\textwidth}
        \caption*{GT}
        \vspace{-0.5em}
        \includegraphics[width=\textwidth]{figures/tor4d_supp/example2_gt.png}
    \end{subfigure}
    \begin{subfigure}[t]{0.15\textwidth}
        \caption*{DPC-LIDAR}
        \vspace{-0.5em}
        \includegraphics[width=\textwidth]{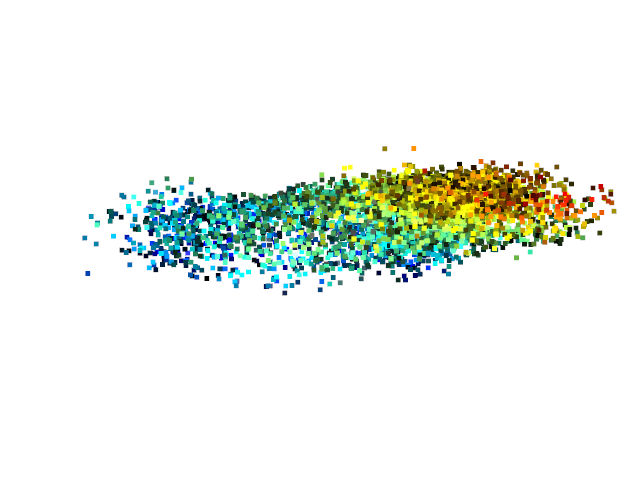}
    \end{subfigure}

    \caption{
        Qualitative results of DPC-LIDAR on the test set of our 3D vehicle dataset.
        All the point clouds are transformed to the ground-truth canonical frame and visualized at a fixed viewpoint.
    }
    \label{fig:dpc-lidar}
\end{figure}

In this section, we describe more details about the implementation of the baseline DPC-LIDAR.
First, We adapt DPC~\cite{insafutdinov2018unsupervised} to range images by replacing perspective transformation with polar transformation.
Different from synthetic data, real data is not normalized and the distance between the partial point cloud and the sensor varies significantly (e.g. 5-30 meters).
However, the camera distance is constant for the original DPC.
Other weakly-supervised approaches, like MVC~\cite{tulsiani2018multi}, also assume little or no translation in relative pose.
Thus, we then scale the canonical shape predicted by DPC in a unit cube to the real world dimensions.
The factor is selected as 6.0, as the average length of vehicles is about 5 meters.
In addition, a radial offset, which is the average of the maximum and the minimum radial distances of the partial point cloud, is provided.
The range image provided as input to DPC is generated directly from the input partial point cloud that we take as input for our approach.
The resolution is $128 \times 128$.
However, DPC-LIDAR performs poorly on real data, even with these modifications.
Fig~\ref{fig:dpc-lidar} showcases some examples of DPC-LIDAR on our 3D vehicle dataset.

\clearpage
\bibliographystyle{splncs04}
\bibliography{egbib}